\PassOptionsToPackage{table,xcdraw,dvipsnames}{xcolor}

\documentclass[11pt,letterpaper]{plum}

\usepackage[all]{hypcap}
\PassOptionsToPackage{round}{natbib}
\usepackage{amsmath,amsfonts,bm}

\def\eqref#1{equation~\ref{#1}}

\def\1{\bm{1}}

\DeclareMathAlphabet{\mathsfit}{\encodingdefault}{\sfdefault}{m}{sl}
\SetMathAlphabet{\mathsfit}{bold}{\encodingdefault}{\sfdefault}{bx}{n}

\usepackage{natbib}
\usepackage{pifont}
\usepackage{nicefrac}
\usepackage{arydshln}
\usepackage{wrapfig}
\usepackage{subcaption}
\usepackage{pgfplots}
\pgfplotsset{compat=1.18}
\usepgfplotslibrary{groupplots}
\tcbuselibrary{most,skins,theorems}
\usepackage{listings}
\usepackage{xspace}
\usepackage{soul}
\usepackage{cleveref}
\usepackage{etoc}
\usepackage[section]{placeins}

\definecolor{darkblue}{rgb}{0,0,0.5}
\definecolor{lightblue}{RGB}{220,235,250}
\definecolor{tabhead}{HTML}{DCE6F1}
\definecolor{tabband}{HTML}{F2F2F2}
\definecolor{cellgood}{HTML}{D8ECD2}
\definecolor{cellbad}{HTML}{F4D2D2}
\definecolor{cellbase}{HTML}{E8E8E8}
\definecolor{textgood}{HTML}{1B7A2D}
\definecolor{textbad}{HTML}{B22222}
\definecolor{textmute}{HTML}{6B6B6B}

\newcommand{\dn}[1]{\textcolor{textbad}{$-#1\,\downarrow$}}

\hypersetup{colorlinks=true,citecolor=darkblue,linkcolor=darkblue,urlcolor=darkblue}

\tcbset{
  takeawaysbox/.style={
    title=Takeaway,
    colback=lightblue!80,
    colframe=black,
    fonttitle=\bfseries\small,
    coltitle=white,
    colbacktitle=black,
    enhanced,
    attach boxed title to top left={xshift=2.5mm,yshift=-2.5mm},
    boxed title style={rounded corners, size=small, colframe=black, colback=black},
    width=\linewidth,
    arc=3.5mm
  }
}

\definecolor{lightorange}{HTML}{faa755}
\definecolor{annothl}{RGB}{255,232,150}
\definecolor{annotaccent}{RGB}{170,40,30}
\definecolor{diffaccent}{RGB}{30,90,170}
\sethlcolor{annothl}
\providecommand{\anchor}[1]{#1}

\tcbset{
  promptbox/.style={
    colback=lightorange!10,
    colframe=black,
    fonttitle=\bfseries\small,
    coltitle=white,
    colbacktitle=black,
    enhanced,
    breakable,
    attach boxed title to top left={xshift=2.5mm,yshift=-2.5mm},
    boxed title style={rounded corners, size=small, colframe=black, colback=black},
    width=\linewidth,
    arc=2mm
  },
  outputbox/.style={
    colback=lightblue,
    colframe=darkblue,
    fonttitle=\bfseries\small,
    coltitle=white,
    colbacktitle=darkblue,
    enhanced,
    breakable,
    attach boxed title to top left={xshift=2.5mm,yshift=-2.5mm},
    boxed title style={rounded corners, size=small, colframe=darkblue, colback=darkblue},
    width=\linewidth,
    arc=2mm
  },
  memorybox/.style={
    colback=gray!4,
    colframe=gray!55,
    fonttitle=\bfseries\small,
    coltitle=white,
    colbacktitle=gray!55,
    enhanced,
    breakable,
    attach boxed title to top left={xshift=2.5mm,yshift=-2.5mm},
    boxed title style={rounded corners, size=small, colframe=gray!55, colback=gray!55},
    width=\linewidth,
    arc=2mm,
    fontupper=\small
  },
  annotbox/.style={
    colback=annotaccent!5,
    colframe=annotaccent,
    fonttitle=\bfseries\small,
    coltitle=white,
    colbacktitle=annotaccent,
    enhanced,
    breakable,
    attach boxed title to top left={xshift=2.5mm,yshift=-2.5mm},
    boxed title style={rounded corners, size=small, colframe=annotaccent, colback=annotaccent},
    width=\linewidth,
    arc=2mm,
    fontupper=\small
  },
  diffbox/.style={
    colback=diffaccent!5,
    colframe=diffaccent,
    fonttitle=\bfseries\small,
    coltitle=white,
    colbacktitle=diffaccent,
    enhanced,
    breakable,
    attach boxed title to top left={xshift=2.5mm,yshift=-2.5mm},
    boxed title style={rounded corners, size=small, colframe=diffaccent, colback=diffaccent},
    width=\linewidth,
    arc=2mm,
    fontupper=\small
  }
}

\newcommand{\annot}[2]{\hl{#2}\textsuperscript{\textcolor{annotaccent}{\bfseries\,(#1)}}}
\newenvironment{annotations}[1][Why this memory is harmful]{%
  \begin{tcolorbox}[annotbox, title={#1}]%
  \begingroup
    \setlength{\parindent}{0pt}%
    \setlength{\parskip}{3pt plus 1pt}%
}{%
  \par
  \endgroup
  \end{tcolorbox}%
}
\newcommand{\annotitem}[3]{%
  \par
  \noindent\hangindent=2.4em\hangafter=1
  \makebox[2.4em][l]{\textcolor{annotaccent}{\bfseries(#1)}}\textbf{#2.}~#3%
}

\lstdefinestyle{solverpy}{
  basicstyle=\ttfamily\scriptsize,
  language=Python,
  breaklines=true,
  columns=fullflexible,
  keepspaces=true,
  showstringspaces=false,
  aboveskip=0pt, belowskip=0pt,
  commentstyle=\color{gray!60!black}\itshape,
  keywordstyle=\color{blue!60!black}\bfseries,
  stringstyle=\color{green!40!black}
}

\title{Useful Memories Become Faulty\\ When Continuously Updated by LLMs}

\author[1]{Dylan Zhang}
\affil[1]{University of Illinois Urbana-Champaign}
\affil[2]{IIIS, Tsinghua University, Work done at UIUC}
\author[2]{Yanshan Lin}
\author[2]{Zhengkun Wu}
\author[1]{Yihang Sun}
\author[1]{Bingxuan Li}
\makeatletter
\let\ABorig@authnote\AB@authnote
\renewcommand\AB@authnote[1]{}
\author[1]{Dianqi Li}
\let\AB@authnote\ABorig@authnote
\makeatother
\author[1]{Hao Peng}
\correspondingauthor{Dylan Zhang; \href{mailto:shizhuo2@illinois.edu}{shizhuo2@illinois.edu}. Project \href{https://github.com/DylanZSZ/Memory-Collapse-Eval}{Github} Here.}

\begin{abstract}
Learning from past experience benefits from two complementary forms of memory: \emph{episodic} traces---raw trajectories of what happened---and \emph{consolidated} abstractions distilled across many episodes into reusable, schema-like lessons. Recent agentic-memory systems pursue the consolidated form: an LLM rewrites past trajectories into a textual memory bank that it continuously updates with new interactions, promising self-improving agents without parameter updates. Yet we find that such consolidated memories produced by today's LLMs are often faulty even when derived from useful experiences. As consolidation proceeds, memory utility first rises, then degrades, and can fall below the no-memory baseline. More surprisingly, even when consolidating from ground-truth solutions, GPT-5.4 fails on 54\% of a set of ARC-AGI problems it had previously solved without memory. We trace the regression to the consolidation step rather than the underlying experience: the same trajectories yield qualitatively different memories under different update schedules, and an episodic-only control that simply retains those trajectories remains competitive with the consolidators we test. In a controlled ARC-AGI Stream environment that exposes \textsc{Retain}, \textsc{Delete}, and \textsc{Consolidate} actions, agents preserve raw episodes by default and double the accuracy of their forced-consolidation counterparts; disabling consolidation entirely (episodic management only) matches this auto regime. Practically, robust agent memory should treat raw episodes as first-class evidence and gate consolidation explicitly rather than firing it after every interaction. Looking forward, reliable agentic memory will require LLMs that can consolidate without overwriting the evidence they depend on.
\end{abstract}

\begin{document}

\maketitle
\makeatletter
\@thanks
\let\@thanks\@empty
\makeatother

\section{Introduction}
\label{sec:intro}

Memory consolidation is the cognitive process by which individual experiences are stabilized, reorganized, and integrated into long-term memory over time. Rather than preserving every episode as a raw record, consolidation helps extract recurring patterns, discard incidental details, and connect new information with existing knowledge. A central hallmark of intelligence is the ability to turn repeated experience into reusable knowledge. Humans do not treat every new situation as entirely unfamiliar; instead, they gradually summarize common structure from past episodes and organize it into abstractions through memory consolidation~\citep{mcclelland1995cls, squire2015memory, dudai2015consolidation, moscovitch2016episodic, schapiro2023gocls, bartlett1932}. These consolidated abstractions allow humans to transfer lessons learned in one context to novel but related situations, enabling efficient reasoning, faster adaptation, and more robust generalization beyond the exact events they have previously encountered.

Recent agentic-memory systems pursue a similar idea for LLM agents. After solving a problem, an agent can distill its trajectory into a textual abstraction of the skill it has learned, store this abstraction in persistent memory, and refine it as more related problems are encountered~\citep{reflexion2023, expel2024, zhang2025ace, sumers2024cognitive, park2023generative, clin2023, voyager2023, wang2024awm, li2026pearl, ouyang2025reasoningbank}. This paradigm is especially appealing because it suggests a practical path toward self-evolving agents: Experience can be accumulated, compressed, and reused as compact textual knowledge, without requiring any parameter updates.

\begin{wrapfigure}{r}{0.7\columnwidth}
\begin{minipage}{\linewidth}
\centering
\vspace{-0.6em}
\begin{subfigure}[b]{0.43\linewidth}
\centering
\includegraphics[width=\linewidth]{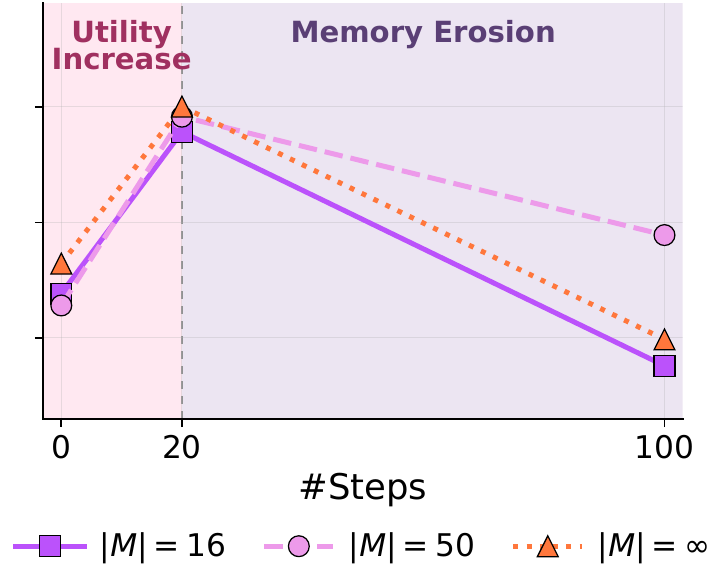}
\caption{ScienceWorld with CLIN}
\label{fig:science-world-bsz4}
\end{subfigure}
\hfill
\begin{subfigure}[b]{0.55\linewidth}
\centering
\includegraphics[width=\linewidth]{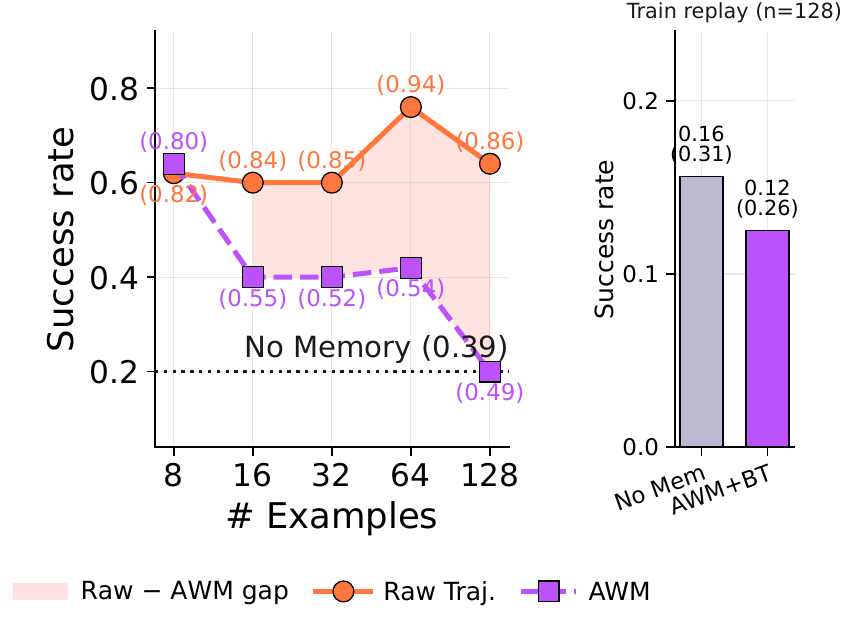}
\caption{Webshop with AWM.}
\label{fig:awm_webshop_scaling}
\end{subfigure}
\caption{Memory utility degrades as consolidation continues. (\subref{fig:science-world-bsz4}) ScienceWorld utility rises early then declines across memory sizes. (\subref{fig:awm_webshop_scaling}) On WebShop, AWM-distilled memory degrades with scale while raw trajectories remain robust. The y-axis is success rate (number solved out of 50 held-out episodes); the parenthesized value next to each marker is the average score, a separate WebShop metric.}
\label{fig:erosion_overview}
\end{minipage}
\vspace{-0.1in}
\end{wrapfigure}

\begin{wrapfigure}{l}{0.44\columnwidth}
\begin{minipage}{\linewidth}
\centering
\vspace{-0.6em}
\includegraphics[width=\linewidth]{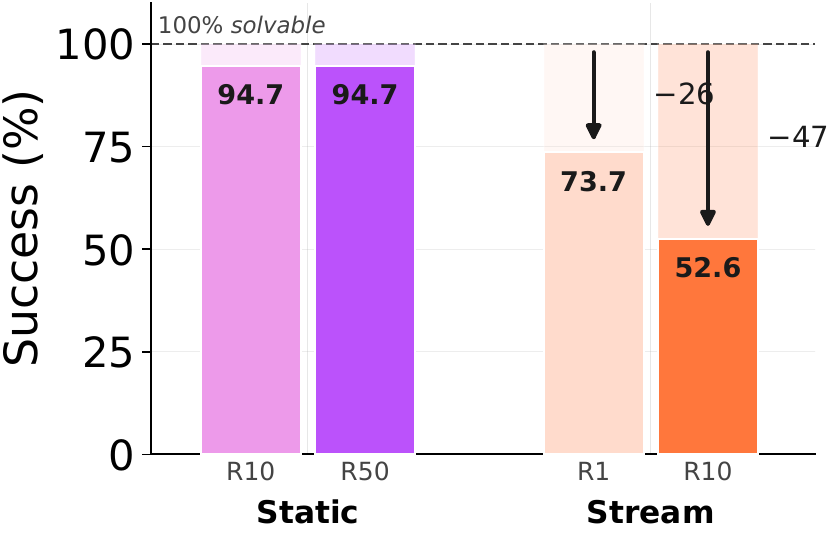}
\caption{{\small \textbf{Iterated stream consolidation collapses on solvable tasks.}
GPT-5.4 accuracy on a 19-problem ARC-AGI slice it solves at $100\%$
without memory (dashed ceiling). Since ground-truth solutions are
available at every consolidation step, all input trajectories are useful
by construction. We compare \textit{Static}, which consolidates the full
pool in one update, with \textit{Stream}, which updates memory after each
problem. Static remains at the ceiling after $R=10$ and $R=50$ refresh
rounds, whereas Stream falls to $52.6\%$ by Round~10 on the same
previously solved problems.}}
\label{fig:arc_19type_54}
\vspace{-0.1in}
\end{minipage}
\end{wrapfigure}

In this work, we systematically examine what happens when agents repeatedly compress past trajectories into memory and reuse that memory over long horizons. To make the experiences themselves informative by construction, we study settings in which the input trajectories are demonstrably useful: either drawn from problems the agent has already solved without memory, or supplied with ground-truth solutions. Even under these favorable conditions, we find that \textbf{current LLMs are not reliable memory consolidators}. Each consolidation step is a lossy rewrite of the memory store: useful details are dropped, spurious rules are introduced, and once-helpful abstractions drift away from the underlying task structure. Across ALFWorld, ScienceWorld, WebShop~\citep{yao2022webshop}, AppWorld~\citep{trivedi2024appworld}, and a controlled testbed we introduce to make consolidation failures more visible, agents that incrementally abstract their accumulating trajectories into a textual memory bank can briefly improve or plateau before degrading---in some cases falling below the same agents without memory at all (Fig.~\ref{fig:science-world-bsz4}). The clearest case isolates the consolidation step from any input-side excuse: GPT-5.4 first solves a set of ARC-AGI problems at $100\%$ accuracy with no memory; after consolidating from ground-truth solutions to those very problems, it then fails on $46\%$ of them (Fig.~\ref{fig:arc_19type_54}). The failure can therefore be attributed to \textbf{faulty memory} rather than to noisy trajectories.

Consolidation is fragile in a second sense: the same set of trajectories can yield qualitatively different memories depending on the order and grouping of updates. Consolidating the whole trajectory pool in one pass (\textbf{Static-All}) versus streaming it batch-by-batch (\textbf{Stream}) produces different end states (Fig.~\ref{fig:whole-batch-vs-stream}); updates on one task overwrite memory of another (Figs.~\ref{fig:arc_old_task_forgetting} and~\ref{fig:cross_task_erosion}); and a stream of repeated near-duplicates causes the memory to overfit to seen instances and generalize poorly within the same task (Fig.~\ref{fig:arc_overfit}). Meanwhile, an episodic-only control that consumes the same trajectories without abstracting them---appending raw rollouts to context as in-context demonstrations---is already competitive with the lesson-style consolidators we test (Table~\ref{tab:strong_baseline_appworld}). Because the trajectory pool is held fixed across these comparisons, the variance across schedules and the episodic-only control together point to the consolidation step itself, not the underlying experience, as the source of the failure.

We further identify three mechanisms behind faulty memory (\S\ref{sec:cogsci}). First, agents misgroup experiences before abstracting them, pooling episodes that do not share underlying structure. Second, even when grouping is correct, abstraction can strip the applicability conditions of a lesson, so that overgeneralized entries interfere with neighboring tasks. Third, when the input stream is narrow, abstraction overfits to seen instances. Together, these failure modes weaken the boundary between what should be generalized, what should remain task-specific, and what should be preserved as raw episodic evidence~\citep{schapiro2023gocls, bartlett1932, brainerd2002fuzzy}.\\

These findings suggest a limitation in current LLM agents: while they can often benefit from individual past episodes, they do not reliably turn those episodes into stable and reusable abstractions over long horizons. Many recent agentic-memory systems---CLIN~\citep{clin2023}, Agent Workflow Memory~\citep{wang2024awm}, Dynamic Cheatsheet~\citep{suzgun2025dynamiccheatsheet}, and ACE~\citep{zhang2025ace}---adopt update-after-every-interaction designs that treat each consolidation step as at worst neutral; our results contradict this assumption. For today's models, retaining episodes is the more robust default: when ARC-AGI Stream agents are given retain, delete, and consolidate actions, they preserve episodes by default and outperform their forced-consolidation counterparts. Yet episodic accumulation cannot be the long-term answer: as deployment continues, raw histories grow unboundedly and resist compositional reuse---compression and transfer ultimately require abstraction. We see faulty memory as a central open problem in agentic memory, and our results motivate architectures that keep the episodic and abstraction-forming roles distinct rather than collapsing both into a single rewrite loop.

\section{Background and Related Work}
\label{sec:related}
\definecolor{envShadeA}{HTML}{F6F0F0}
\definecolor{envShadeB}{HTML}{F2EAE0}

\begin{wraptable}{r}{0.5\linewidth}
\centering
\scriptsize
\renewcommand{\arraystretch}{1.20}
\setlength{\tabcolsep}{4pt}
\setlength{\dashlinedash}{1.4pt}
\setlength{\dashlinegap}{1.4pt}
\arrayrulecolor{black!55}
\vspace{-0.6em}

\label{tab:settings_main}
\begin{tabular}{@{}>{\raggedright\arraybackslash}p{0.76\linewidth} >{\raggedright\arraybackslash}p{0.20\linewidth}@{}}
\toprule
\textbf{Setting} & \textbf{Figure(s)} \\
\midrule
\rowcolor{envShadeA}\textbf{\textsc{ALFWorld}} & \\
\rowcolor{envShadeB}GT expert trajectories ($128$ per type, $6$ types). & Figs.~\ref{fig:whole-batch-vs-stream}, \ref{fig:alfworld_drop} \\
\hdashline
\rowcolor{envShadeA}\textbf{\textsc{ScienceWorld}} w. GPT5.4-mini & \\
\rowcolor{envShadeB}Offline task stream, batch size $4$. & Fig.~\ref{fig:science-world-bsz4} \\
\rowcolor{envShadeA}Self-solve task-switch sequence. & Fig.~\ref{fig:science_world_task_switch} \\
\hdashline
\rowcolor{envShadeB}\textbf{\textsc{WebShop}} w. GPT-5.4mini & \\
\rowcolor{envShadeA}AWM , $50$-episode held-out eval. & Fig.~\ref{fig:awm_webshop_scaling} \\
\hdashline
\rowcolor{envShadeB}\textbf{\textsc{AppWorld}} & \\
\rowcolor{envShadeA}AWM with trajectories from AgentGym. & \\
\hdashline
\rowcolor{envShadeB}\textbf{\textsc{ARC-AGI}} & \\
\rowcolor{envShadeA} Subsets For Individual Experiments. & Figs.~\ref{fig:arc_overfit}, \ref{fig:arc_old_task_forgetting}, \ref{fig:repeated_consolidation_forget}, \ref{fig:arc_19type_54} \\
\rowcolor{envShadeB}Stream: offline GT-solution; eval on $50$ held-out tasks (same family, unseen). & Figs.~\ref{fig:force_abstraction}, \ref{fig:bsz8_diagnostics}; Tab.~\ref{tab:arc-family-obs} \\
\rowcolor{envShadeA}Stream: self-solve, $400$ steps. & Fig.~\ref{fig:arc_train_curves} \\
\bottomrule
\end{tabular}
\arrayrulecolor{black}
\caption{Settings used across the paper. The right column points to the figure(s) and table(s) each setting produced.}
\vspace{-0.6em}
\end{wraptable}

\paragraph{Memory in LLM agents.}
LLM agents augment fixed-weight models with editable textual memory that they themselves maintain. Across architectures, this memory takes many forms: hierarchical context stores \citep{packer2023memgpt, letta2024}, reflective notes \citep{reflexion2023}, episodic streams \citep{park2023generative}, distilled lessons \citep{expel2024}, skill libraries \citep{voyager2023}, causal abstractions \citep{clin2023}, dynamic memory stores \citep{chhikara2025mem0, xu2025amem, langmem2024}, evolving playbooks \citep{zhang2025ace}, dynamic cheatsheets \citep{suzgun2025dynamiccheatsheet}, and reasoning memory banks \citep{ouyang2025reasoningbank}.
A popular paradigm is that LLMs abstract useful procedural information from experience into text. This work mainly concerns that category.

\paragraph{Schema formation and metacognitive control.}
Cognitive theories of memory describe two complementary processes: a fast episodic system that stores individual experiences, and a slow process that extracts compact \emph{schemas}---abstractions over many episodes that integrate with prior knowledge to support future generalization \citep{mcclelland1995cls, squire2015memory, dudai2015consolidation, moscovitch2016episodic}. Schema formation is beneficial when new evidence fits existing structure \citep{tse2007schemas}, but harmful when it is non-selective or poorly controlled: nonselective consolidation causes interference and loss of specificity \citep{mcclelland1995cls, schapiro2023gocls}. Whether consolidation helps therefore depends on \emph{metacognitive control}---deciding which experiences belong together, how abstractly to rewrite them, and when to preserve distinctions rather than collapse them.
Human metacognition is itself imperfect \citep{flavell1979metacognition, nelson1990metamemory, koriat1997monitoring}, and whether LLMs reliably monitor their self-generated abstractions is an open question---in adjacent settings, chain-of-thought explanations have been shown to diverge from the underlying computation \citep{turpin2023cot}. Agent-memory systems implement an explicit analogue, abstracting ``lessons'' that are reused on later tasks while putting the same model in charge of both generating the memory and monitoring its own abstraction; the failure modes we observe (\S\ref{sec:cogsci}) are consistent with this control loop being unreliable in practice.

\paragraph{Relation to catastrophic forgetting.}
The degradation we study is related to but distinct from catastrophic forgetting~\citep{mccloskey1989catastrophic, french1999catastrophic}---the loss of earlier-task performance when a model is trained sequentially on new tasks. Our failure is not induced by a change of task distribution: it occurs with the task set held fixed (repeated consolidation renders ${\sim}46\%$ of previously solved problems unsolvable) and it manifests as \emph{distortion}---merged task families, over-generalized rules, confident-but-incorrect procedures---rather than the fading of old knowledge. Consolidating the same trajectories in one pass also beats streaming them, so the damage comes from repeated rewriting rather than from the evidence. It therefore calls for different remedies (gated, delayed abstraction with preserved episodes) than the rehearsal or drift-regularization used against catastrophic forgetting; we expand the comparison in \Cref{app:cf}.

\section{Experiment Set-up}
\label{sec:setup}
\begin{wrapfigure}{r}{0.6\columnwidth}
\begin{minipage}{\linewidth}
\centering
\vspace{-0.6em}
\includegraphics[width=\linewidth]{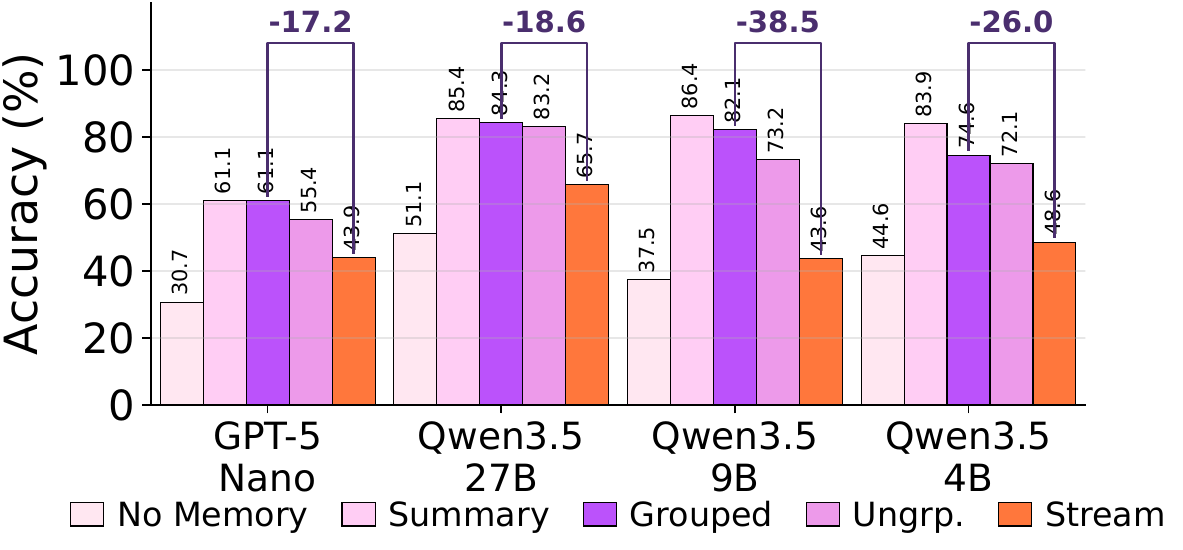}
\caption{\textbf{Streaming consolidation collapses.} \textit{Stream} loses
$17$--$38$ pts vs.\ whole-batch \textit{Pool}.}
\label{fig:whole-batch-vs-stream}
\vspace{-0.6em}
\end{minipage}
\end{wrapfigure}

We study how models abstract memory from experience across clean and widely-used agent benchmarks---ALFWorld, ScienceWorld, WebShop, AppWorld~\citep{trivedi2024appworld}, and Mind2Web~\citep{deng2023mind2web}---and a synthetic stream we built, \emph{ARC-AGI Stream}, to introspect the memory-management process under known task families and ground-truth solutions.

We adopt representative agentic-memory methods that distill lessons from past trajectories---CLIN~\citep{clin2023}, Agent Workflow Memory~\citep{wang2024awm}, Dynamic Cheatsheet~\citep{suzgun2025dynamiccheatsheet}, and ACE~\citep{zhang2025ace}.

\paragraph{Memory construction conditions.}
Three regimes differ only in how trajectories are presented to the memory manager: \textbf{Static-All} abstracts memory from the entire pool in one-shot; \textbf{Static-Group} does so for each task type (for ALFWorld and ScienceWorld) and concatenate: this simulates offline memory construction and is used for control; \textbf{Stream} updates in batches, simulating an agent that evolves continually. For Stream we change batch size and within-batch task composition (homogeneous vs.\ heterogeneous) and consider both unbounded and size-capped memory banks. At evaluation we provide all memory entries in context and average two repeats per question unless noted otherwise.

\paragraph{ARC-AGI Stream.}
We introduce \emph{ARC-AGI Stream}, a controlled testbed for introspecting the abstraction process itself rather than only its downstream accuracy. Building on ARCGEN~\citep{moffitt2025arcgen}, the testbed combines three properties that, jointly, are absent from existing agent benchmarks: \emph{(i)} a fully specified latent task taxonomy---following ConceptARC~\citep{CHECK:conceptarc}, each concept is a distinct problem type, so memory-management actions (Fig.~\ref{fig:misclassification_bsz8}) and results (Fig.~\ref{fig:arc_buffer_composition}) can be traced family-by-family. \emph{(ii)} programmatic ground-truth and \emph{(iii)} an exposed, structured vocabulary of memory operations whose use we can audit per step. These controls make abstraction errors attributable: if a model receives correct episodes from known families but later merges families, the failure lies in memory abstraction rather than others. Together they enable the family-coverage, misclassification, and buffer-composition analyses in \S\ref{sec:cogsci} that would not be measurable on ALFWorld/ScienceWorld/WebShop. Following the Complementary Learning Systems framing~\citep{mcclelland1995cls, schapiro2023gocls, tse2007schemas}, we maintain two stores: an \textbf{Episodic} buffer of problem --solution pairs and an \textbf{Abstract} store of consolidated lessons; the control conditions below are designed to test this gating prediction directly. At each step the model may \textsc{Retain} a raw episode, \textsc{Delete} an entry, or \textsc{Consolidate} buffered episodes into the Abstract store;
the solver can be conditioned on either store or both stores. We compare three control loops: \textbf{Force} requires consolidation each round and no Episodic entries persist between rounds; \textbf{Auto} lets the model choose whether to retain raw or abstract; \textbf{Episodic Management Only} allows only retention and deletion, disabling abstraction entirely. Two trajectory regimes complete the design: \textbf{GT}, where ground-truth solutions are streamed to simulate training on demonstrations, and \textbf{Running}, where the agent learns from its own attempts as it solves new problems. Prompt templates are documented in \Cref{app:prompts}; the family and skill taxonomy used by the generator is in \Cref{app:taxonomy}.

\section{Useful Memories Become Faulty When Continuously Updated by LLMs}
\label{sec:phenomenon}

\subsection{Stream Updates Reveal The Cumulative Effect Of Imperfect Abstraction}
\paragraph{
Abstracted memory utility is non-monotonic in updates.}
Starting from an empty memory store,
abstracted-memory utility rises early and then falls below the no-memory baseline as updates accumulate.
Figure~\ref{fig:science-world-bsz4} shows this curve on ScienceWorld with batches of 4 trajectories: the score peaks near step 20 and declines through step 100 for every memory size and sometimes below no-memory baseline. The same shape recurs on WebShop, where AWM goes from $0.64$ at $8$ examples to $0.20$ at $128$ while no-memory sits at $0.20$ (Fig.~\ref{fig:awm_webshop_scaling})---scaling the memory eventually erases its own benefit.

\paragraph{A strong memory is not a fixed point.}

\begin{wrapfigure}{l}{0.6\columnwidth}
\begin{minipage}{\linewidth}
\centering
\vspace{-0.6em}
\includegraphics[width=\linewidth]{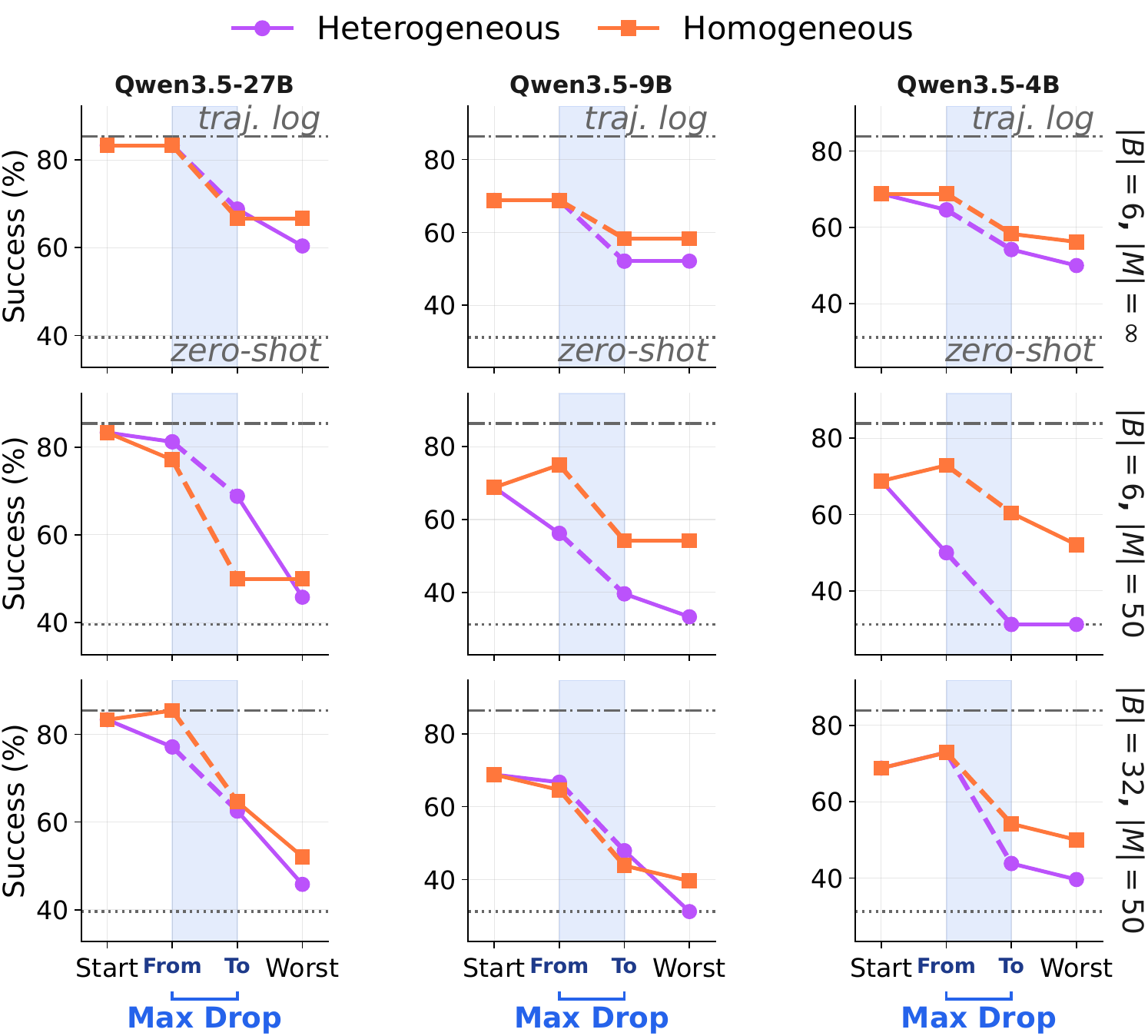}
\caption{Heterogeneous batches accelerate erosion.}
\label{fig:alfworld_drop}
\vspace{-0.1in}
\end{minipage}
\end{wrapfigure}

A high-quality starting memory does not stabilize under continued consolidation either. We first consolidate ALFWorld trajectories with GPT-5.4---the strongest model we test---following the \textbf{Static-Group} setting from \S\ref{sec:setup}, then continue updating that store in batches with GPT-5-nano on the same trajectory pool.
Evaluated under three solvers (Qwen3.5-\{27B, 9B, 4B\}) to avoid idiosyncraticy, utility declines across all three solvers and can drop drastically between consecutive evaluation steps (Fig.~\ref{fig:alfworld_drop}).

\paragraph{The agent regresses on tasks it has already solved.}
The cleanest, most striking instance is in-distribution regression on a 19-problem slice of the ARC-AGI Stream that GPT-5.4 solves at $100\%$ accuracy without memory. Streaming those same problems through the consolidation loop, with ground-truth solutions available at every step, brings GPT-5.4 down to $54\%$ on the very problems it had previously solved (Fig.~\ref{fig:arc_19type_54}). Repeating the experiment ten times reproduces the drop, and on the broader set of $33$ tasks the model solves without memory, consolidated memory leaves a third of them unsolved by epoch~1 and $40\%$ by epoch~10 (Table~\ref{tab:robustness-ci}).
\begin{table}[t]
\centering
\small
\setlength{\tabcolsep}{8pt}
\renewcommand{\arraystretch}{1.15}
\begin{tabular}{@{}lcc@{}}
\toprule
\textbf{GPT-5.4, ARC-AGI Stream (10 runs)} & \textbf{pass@1 (mean\,$\pm$\,std)} & \textbf{pass@10} \\
\midrule
\rowcolor{cellbase} No memory (baseline) & $58.5 \pm 5.4$ & $100$\% \,(33/33) \\
Consolidated memory, epoch 1  & $55.1 \pm 4.9$ & $66.7$\% \,(22/33) \\
Consolidated memory, epoch 10 & \cellcolor{cellbad}$38.5 \pm 5.3$ & \cellcolor{cellbad}$60.6$\% \,(20/33) \\
\bottomrule
\end{tabular}
\caption{\textbf{The regression is not sampling noise, and it shows up as lost coverage.}
Over ten end-to-end runs on the $33$ tasks solvable without memory, consolidated memory leaves a third of them unsolved by epoch~1 and $40\%$ by epoch~10 (pass@10), while mean accuracy falls from $58.5$ to $38.5$. The headline $100\%\!\to\!54\%$ in-distribution regression thus survives repetition and an expanded task set.}
\label{tab:robustness-ci}
\end{table}

\paragraph{The consolidation procedure itself shapes the rate of decline.}

The decline is not driven by content alone; how trajectories are presented to the consolidator changes the resulting memory in content-independent ways.
1. \textbf{Grouping by task family helps.} Holding update strategy fixed, consolidating each task family separately (\textbf{Static-Group}) outperforms consolidating across families jointly (\textbf{Static-All} - marked as Ungrp.); pooling heterogeneous tasks into a single abstraction step invites confusion between unrelated procedures (Fig.~\ref{fig:whole-batch-vs-stream}).
2. \textbf{One-shot consolidation beats streaming.} Holding the trajectory pool fixed, \textbf{Stream}ed incremental updates underperform a single whole-pool pass (Fig.~\ref{fig:whole-batch-vs-stream}).
    This suggests that early abstractions can anchor later rewrites and mistakes are reused as context for future consolidation, so small errors in segmentation or abstraction compound into progressively distorted memories. However, in continual learning loops, memory gets repeatedly updated as agents collect more experiences, instead of seeing the entire pool of past interactions, precisely falling to the lossy update regime.

3. \textbf{Heterogeneous batches accelerate the decline within streaming.} Even when updates must be streamed, batches that mix tasks produce worse memory than batches drawn from one task at a time (Fig.~\ref{fig:alfworld_drop}): the model merges incompatible lessons inside a single update.
\begin{wrapfigure}{r}{0.4\linewidth}
\centering
\vspace{-1em}
\includegraphics[width=\linewidth]{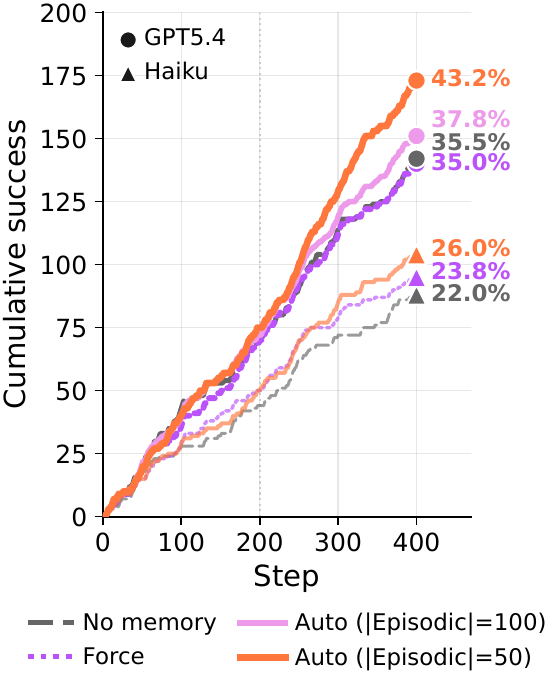}
\caption{{\small Cumulative success on ARC-AGI Stream across $400$ training steps.
Auto with a raw-episode buffer matches or beats both \emph{Force} (forced abstraction
of every step) and the \emph{No memory} baseline on both backbones.}}
\label{fig:arc_train_curves}
\vspace{-2cm}
\end{wrapfigure}

\paragraph{Implications for memory design.}
Each abstraction pass is a lossy rewrite that compounds over updates. The pattern makes two predictions for how memory should be organized, which \S\ref{sec:lessons-underperform} tests directly. (i)~The raw trajectory logs that the consolidator distills from should remain a competitive baseline, since they retain what each rewrite discards. (ii)~Allowing an agent to keep raw episodes alongside (or instead of) abstracting them should outperform a regime that forces every trajectory through consolidation.

\subsection{An episodic-only store should be a diagnostic baseline}
\label{sec:lessons-strong-baseline}
\begin{table}[t]
\centering
\small
\setlength{\tabcolsep}{7pt}
\renewcommand{\arraystretch}{1.2}
\begin{tabular}{@{}lcccc@{}}
\toprule
\textbf{Backbone} & \textbf{No memory} & \textbf{Peak} & \textbf{Step 400} & \textbf{Peak\,$\to$\,400} \\
\midrule
GPT-5.5          & $78.0$ & \cellcolor{cellgood}$80.0$ & \cellcolor{cellbad}$68.0$ & \dn{12.0} \\
Claude Sonnet 5  & $40.0$ & \cellcolor{cellgood}$58.0$ & $48.0$                     & \dn{10.0} \\
GLM 5.2          & $22.0$ & \cellcolor{cellgood}$44.0$ & \cellcolor{cellbad}$22.0$ & \dn{22.0} \\
DeepSeek V4 Pro  & $28.0$ & \cellcolor{cellgood}$40.0$ & \cellcolor{cellbad}$18.0$ & \dn{22.0} \\
\bottomrule
\end{tabular}
\caption{\textbf{The improve-then-degrade pattern holds for frontier agents that manage their own memory.}
Accuracy (\%) on ARC-AGI Stream when a general-purpose coding agent (OpenCode), at each backbone's maximum reasoning budget, is given full control to inspect, revise, and delete memory. Every backbone rises above no-memory and \colorbox{cellgood}{peaks}, then \colorbox{cellbad}{falls to or below its no-memory level} by step~400: GLM~5.2 returns to no-memory and DeepSeek~V4~Pro drops beneath it. Greater capability moves the peak but does not remove the decline.}
\label{tab:generality-opencode}
\end{table}

In WebShop, ALFWorld and AppWorld, an \emph{episodic-only} memory---append-only trajectory rollouts in context, with no cross-trajectory abstraction ---is competitive with lesson-style consolidator we test (Table~\ref{tab:strong_baseline_appworld}; WebShop in Table~\ref{tab:strong_baseline_webshop} in Appendix). The trajectory baseline stores the same experience that ACE, AWM, and Dynamic Cheatsheet are asked to compress, and all memory entries are placed in context at evaluation.

In most cases, abstracted memory cannot significantly outperform direct in-context learning from trajectory logs (or subsets of them). It means the evidence available before abstraction already contains task-relevant signal that the solver can exploit directly: observations, actions, intermediate failures, and environmental feedback remain tied to the concrete situation in which they occurred. Schema-like behavior can emerge
from in-context learning on preserved instances comparative to a separately stored abstraction \citep{hintzman1986schema,brooks1978nonanalytic,medin1978context,logan1988instance}. Therefore, we suggest that a memory method whose value depends on distillation should be tested against the unabstracted rollouts it distills.

\providecommand{\beatmark}{\textcolor{textbad}{\scriptsize\,$(\text{--})$}}
\begin{table}[t]
\begin{minipage}{\linewidth}
\centering
\scriptsize
\setlength{\tabcolsep}{2pt}
\renewcommand{\arraystretch}{1.20}
\resizebox{\textwidth}{!}{%
\begin{tabular}{@{}l c cccccc cccc@{}}
\toprule
& & \multicolumn{6}{c}{\textbf{Trajectory Logs (Summarized)}} & & & & \\
\cmidrule(lr){3-8}
\textbf{Backbone} & \textbf{NoMem}
& \textbf{All} & \textbf{FS=64} & \textbf{FS=32} & \textbf{FS=16}
& \textbf{Incr.\ Keep} & \textbf{One-Pass Keep}
& $\mathbf{ACE_{GT}}$ & \textbf{ACE} & \textbf{AWM} & \textbf{DC} \\
\midrule
\multicolumn{12}{@{}l}{\textit{AppWorld} (TGC\%) } \\
\rowcolor{cellbase}
Qwen3.5-27B
              & $66$
              & $73$
              & $66$
              & $65$
              & $68$
              & $69$
              & $69$
              & $\mathbf{76}$
              & $65$\beatmark
              & $68$\beatmark
              & $68$\beatmark \\
\rowcolor{tabband}
GPT-5-Mini
              & $14$
              & $\mathbf{23}$
              & $15$
              & $20$
              & $16$
              & $15$
              & $19$
              & $10$\beatmark
              & $13$\beatmark
              & $19$\beatmark
              & $8$\beatmark \\
GPT-5.4-Mini
              & $52$
              & $59$
              & $58$
              & $58$
              & $54$
              & $\mathbf{64}$
              & $57$
              & $57$\beatmark
              & $52$\beatmark
              & $52$\beatmark
              & $52$\beatmark \\
\rowcolor{tabband}
Claude-Haiku-4.5
              & $68$
              & $66$
              & $69$
              & $69$
              & $70$
              & $70$
              & $59$
              & $60$\beatmark
              & $\mathbf{74}$
              & $68$\beatmark
              & $64$\beatmark \\
\midrule
\multicolumn{12}{@{}l}{\textit{ALFWorld} (Task Success \%) } \\
\rowcolor{cellbase}
GPT-5.4-Mini
              & $54$
              & $90$
              & $\mathbf{92}$
              & $81$
              & $77$
              & $85$
              & $81$
              &  $85$\beatmark
              & $79$\beatmark
              & $65$\beatmark
              & $58$\beatmark \\
\rowcolor{tabband}
GPT-5-Mini
              & $52$
              & $\mathbf{81}$
              & $\mathbf{81}$
              & $77$
              & $67$
              & $75$
              & $65$
              & $56$\beatmark
              & $60$\beatmark
              & $48$\beatmark
              & $50$\beatmark \\
Claude-Haiku-4.5
              & $79$
              & $\mathbf{92}$
              & $85$
              & $77$
              & $73$
              & $79$
              & $90$
              & $83$\beatmark
              & $73$\beatmark
              & $81$\beatmark
              & $81$\beatmark \\
\bottomrule
\end{tabular}%
}
\vspace{0.1in}
\caption{We report consolidated memory's effect against the raw trajectories they abstract from. 
A red \textcolor{textbad}{$(\text{--})$} marks any non-trajectory-log approach that is beaten by at least one trajectory-log baseline in the same row.
Best per row in \textbf{bold}. WebShop results are reported separately in Table~\ref{tab:strong_baseline_webshop}.}
\label{tab:strong_baseline_appworld}
\end{minipage}
\end{table}

\subsection{The degradation generalizes across frameworks, backbones, and agents}
\label{sec:generality}
The phenomenon is not specific to the four consolidation methods or the backbones used above. We evaluate seven published memory frameworks---LangMem~\citep{langmem2024}, Mem0~\citep{chhikara2025mem0}, ProPlay, ACE~\citep{zhang2025ace}, AWM~\citep{wang2024awm}, Letta~\citep{letta2024}, and Dynamic Cheatsheet~\citep{suzgun2025dynamiccheatsheet}---with two backbones (GPT-5.4-mini and Claude Sonnet~5) across AppWorld, WebShop, and ScienceWorld, and every framework degrades from its peak in at least one setting (full table in \Cref{app:generality}). The effect also survives handing memory management to a frontier agent: using the OpenCode harness at each model's maximum reasoning budget, with full latitude to inspect, revise, and delete entries, all four backbones improve over no-memory, peak, and then decline---GLM~5.2 back to its no-memory level and DeepSeek~V4~Pro below it (Table~\ref{tab:generality-opencode}). The collapse is likewise not an artifact of context length (padding the prompt to ${\sim}790$K tokens changes accuracy by fewer than $4$ points; \Cref{app:contextrot}), nor is it repaired by prompt optimization (GEPA recovers only part of the loss, and does so by instructing the solver to distrust the memory; \Cref{app:gepa}).

\begin{tcolorbox}[takeawaysbox]
Repeated textual consolidation degrades useful memory across seven frameworks, five backbones, an autonomous coding agent, contexts up to ${\sim}790$K tokens, and prompt optimization. Capability and sophistication move where the peak sits; they do not remove the decline.
\end{tcolorbox}

\section{Mitigating Faulty Abstraction}
\label{sec:lessons-underperform}

The findings in \S~\ref{sec:phenomenon} imply experiences contain information the LLMs can consume for better task performance, but the challenge is that those information fades with repeated LLM updates instead of being refined. 

Therefore, the memory architecture should be allowed to preserve episodes instead of forcing every new experience through consolidation. In this section, we present a controlled study to test whether granting models autonomy of retention versus abstraction, and allowing parallel stores of abstract and episodic memory would be a mitigation to the effect. 

ARC-AGI Stream lets us instantiate the two-process design explicitly. The agent maintains both an \textbf{Episodic} buffer and an \textbf{Abstract} store, and at each step may \textsc{Retain}, \textsc{Delete}, or \textsc{Consolidate} (\textbf{Auto}; cf.\ \S~\ref{sec:setup}). We compare this to \textbf{Force}, where consolidation is required at every round and no episodic entry persists between rounds. Three observations follow.

First, the episodic store carries most of the gain.
Removing episodic evidence and using only abstract lessons collapses accuracy back to the no-memory baseline (Fig.~\ref{fig:arc_management_strategies}), indicating that the useful information was retained in raw episodes rather than distilled entries.

Second, \emph{Episodic Management Only}---retention and deletion of raw episodes with abstraction disabled---matches or exceeds the full Auto mode (Fig.~\ref{fig:arc_management_strategies}). In this regime, the agent's gains do not require an abstract store at all.
\begin{wrapfigure}{l}{0.45\linewidth}
\begin{minipage}{\linewidth}
\centering
\vspace{-0.6em}
\includegraphics[width=\linewidth]{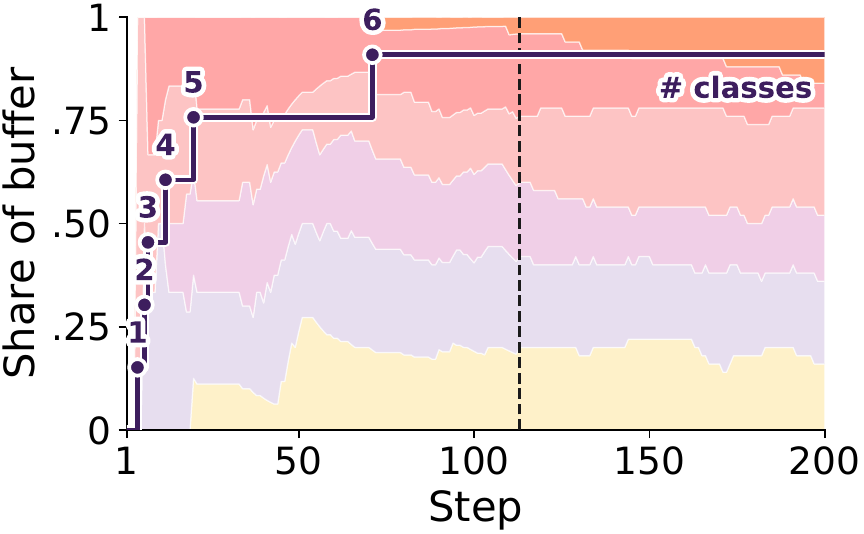}
\caption{Buffer composition over time on the ARC-AGI Stream
}
\label{fig:arc_buffer_composition}
\vspace{-0.6em}
\end{minipage}
\end{wrapfigure}

Third, forced abstraction underperforms retained-episode policies. Across $400$ training steps, Auto---which keeps episodes by default and uses abstraction sparingly---generally outperforms Force on both backbones we test (Fig.~\ref{fig:arc_train_curves}). Whatever Force gains from compression, it can lose more by overwriting evidence.

The agent's own management decisions point in the same direction. Given the option, it saturates the episodic buffer quickly at both budget levels while keeping the abstract store sparse (Fig.~\ref{fig:buffer_size_arcagi_gt}). The Auto-mode result therefore reflects an episodic-first policy that the agent chooses when the architecture permits it.

\paragraph{Principle.}
Episodic and schema-forming roles should not be collapsed into a single rewrite loop. Complementary Learning Systems make this prescription explicitly: a fast episodic store and a slow schema-forming store are kept architecturally distinct so that fast learning does not overwrite the slow one, and consolidation is gated by schema fit rather than triggered on every event \citep{mcclelland1995cls,kumaran2016cls,tse2007schemas,schapiro2023gocls}. The Force regime collapses this separation into a mandatory rewrite loop, recreating the interference conditions the dual-system design was meant to prevent \citep{mccloskey1989catastrophic,french1999catastrophic}. An episodic store should therefore retain the evidence; abstraction, when it happens, should be opt-in and gated by the agent rather than forced on every trajectory.

In our experiments, both policies that gate abstraction (Auto) and policies that disable it entirely (Episodic Management Only) match or exceed forced abstraction (Force), pointing to \emph{mandatory rewriting at every step}---rather than abstraction itself---as the decisive failure mode in this regime \citep{nelson1990metamemory,koriat1997monitoring}.

\section{Failure Modes}
\label{sec:cogsci}

We introspect three abstraction failures both quantitatively and qualitatively: misgrouping before abstraction, interference from overgeneralized lessons, and overfit to narrow streams of examples. We present full case study results in \Cref{app:wrong-examples}.

\subsection{Correct Segmentation Takes Time, Forced Abstraction Breaks It}
\label{sec:chunking-failure}

Correct segmentation---identifying which experiences belong together before abstraction---is a precondition: an agent cannot abstract a lesson from a set of episodes it has failed to group correctly~\citep{Zacks2007EventPA}. 

\begin{figure}[t]
\centering
\begin{subfigure}[t]{0.23\linewidth}
\centering
\includegraphics[width=\linewidth]{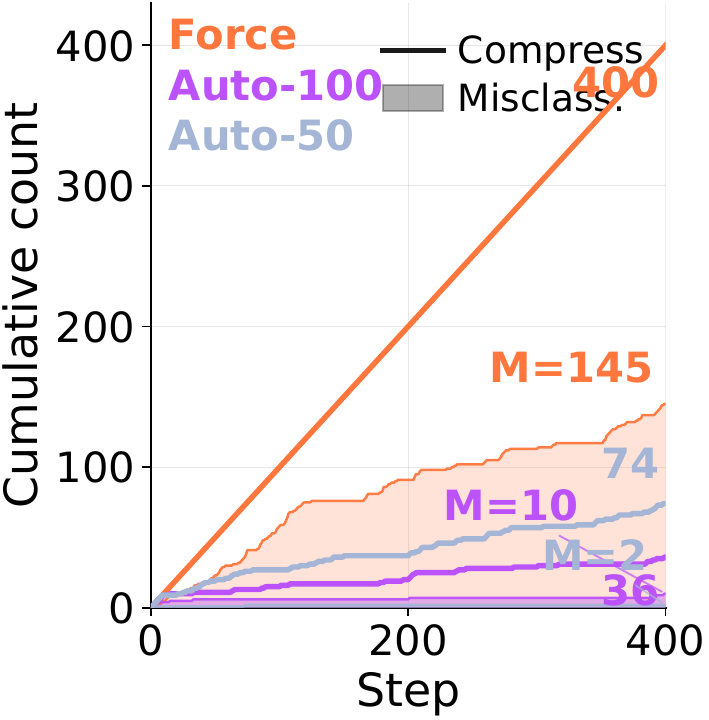}
\caption{Memory diagnostics at bsz=8.}
\label{fig:misclassification_bsz8}
\end{subfigure}
\hfill
\begin{subfigure}[t]{0.34\linewidth}
\centering
\includegraphics[width=\linewidth]{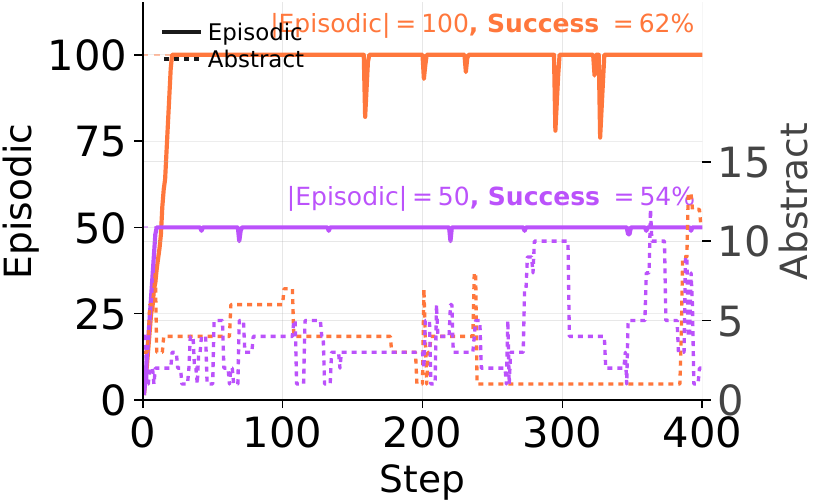}
\caption{Episodic store saturates for both 50 and 100.}
\label{fig:buffer_size_arcagi_gt}
\end{subfigure}
\hfill
\begin{subfigure}[t]{0.34\linewidth}
\centering
\includegraphics[width=\linewidth]{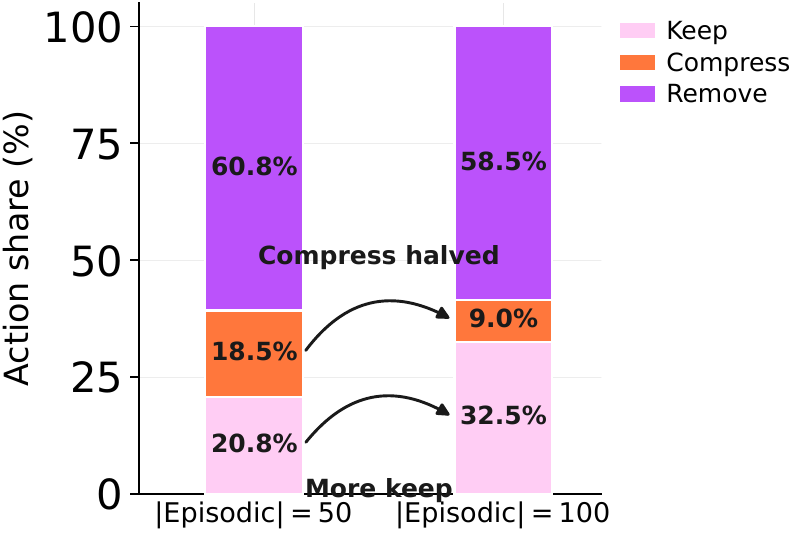}
\caption{Bigger buffer halves Compress, grows Keep.}
\label{fig:arc-meta-actions}
\end{subfigure}
\caption{\textbf{Auto-mode diagnostics on ARC-AGI \textbf{GT} Stream.} (\subref{fig:misclassification_bsz8}) Compression budget vs.\ misclassifications. (\subref{fig:buffer_size_arcagi_gt}) Episodic store saturation under different caps. (\subref{fig:arc-meta-actions}) Distribution of meta-level memory actions as the buffer grows.}
\label{fig:bsz8_diagnostics}
\end{figure}

The capacity for correct segmentation is in fact \textbf{present}. Given autonomy over its memory, the model eventually converges to a clean episodic store covering each of the 6 problem types(Fig.~\ref{fig:arc_buffer_composition}).
Yet the process takes time: only after 71 steps (568 examples) does the model admit all six problem types into the buffer.
In the Force regime, where the model must consolidate at every step (8 fresh examples), it makes mistakes frequently. Figure~\ref{fig:misclassification_bsz8} reports the \emph{misclassification count} — the number of memory-management actions that combine episodes from two or more distinct problem classes — and shows that, under forced consolidation, the model routinely merges across class boundaries. The bottleneck is not recognizing class structure but preserving it through
cross-episode abstraction: the forced act of abstraction overrides the segmentation the model is otherwise capable of.

\subsection{Interference}
\label{sec:catastrophic-interference}

\begin{figure}[t]
\centering
\begin{subfigure}[t]{0.32\linewidth}
\centering
\includegraphics[width=\linewidth]{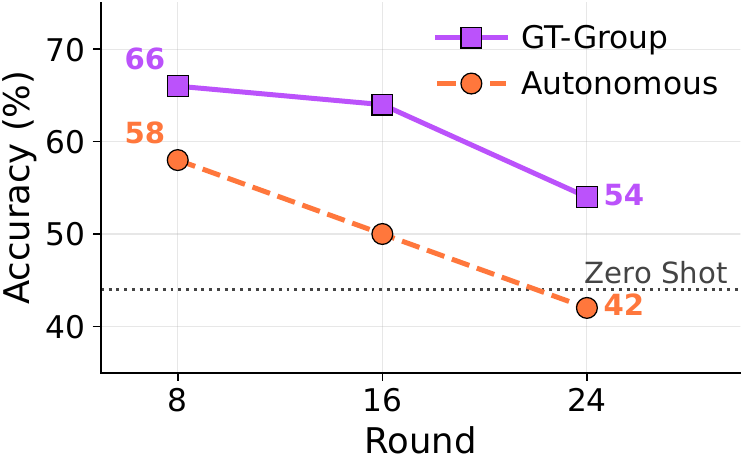}
\caption{{Forgetting under repeated abstraction.} \textit{Auto} falls below zero-shot.}
\label{fig:repeated_consolidation_forget}
\end{subfigure}
\hfill
\begin{subfigure}[t]{0.32\linewidth}
\centering
\includegraphics[width=\linewidth]{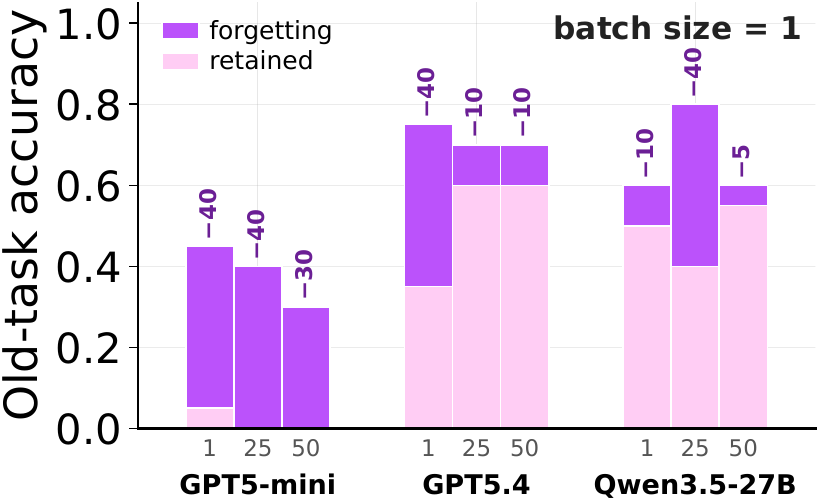}
\caption{Switch-induced forgetting in ARC-AGI.}
\label{fig:arc_old_task_forgetting}
\end{subfigure}
\hfill
\begin{subfigure}[t]{0.32\linewidth}
\centering
\includegraphics[width=\linewidth]{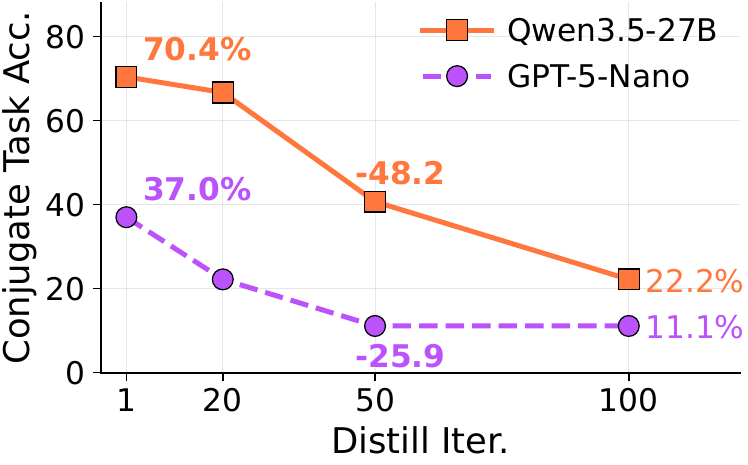}
\caption{More distillations on one task increasingly interfere with another type.}
\label{fig:cross_task_erosion}
\end{subfigure}
\caption{\textbf{Three views of consolidation-induced forgetting.} (\subref{fig:repeated_consolidation_forget}) Repeated consolidation on the same pool erodes utility round by round. (\subref{fig:arc_old_task_forgetting}) A single task switch is sufficient to overwrite a non-trivial fraction of what was previously learned. (\subref{fig:cross_task_erosion}) More distillations on one task increasingly interfere with another type.}
\label{fig:forgetting_combined}
\end{figure}

Each abstraction pass smooths and abstracts existing entries---a process analogous to the semanticization of episodic memories under repeated retelling~\citep{bartlett1932}; when the underlying chunks are imprecisely bounded, this abstraction strips the preconditions under which a lesson applies, leaving what reads as broadly relevant when it is not. On a related but different problem (Pick\&Place versus Pick-Clean-Place in ALFWorld), the agent sees these overgeneralized lessons and is misled rather than aided (Fig.~\ref{fig:cross_task_erosion}).
\begin{figure}[t]
\centering
\begin{subfigure}[t]{0.38\linewidth}
\centering
\includegraphics[width=\linewidth]{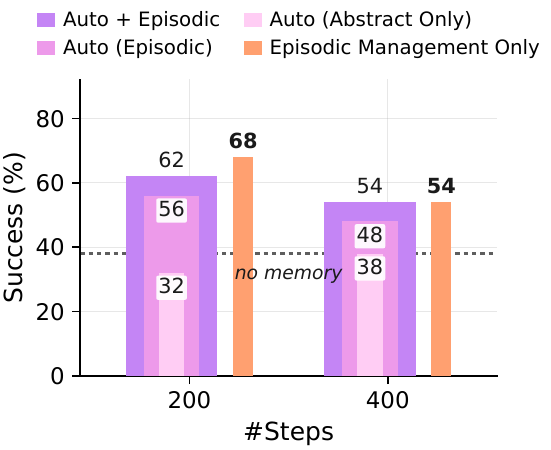}
\caption{In \textbf{Auto}, using the episodes it retained alone shows strong performance, and it synergizes with the abstract store. Meanwhile, managing episodic memory alone beats or matches the performance. }
\label{fig:arc_management_strategies}
\end{subfigure}
\hfill
\begin{subfigure}[t]{0.6\linewidth}
\centering
\includegraphics[width=\linewidth]{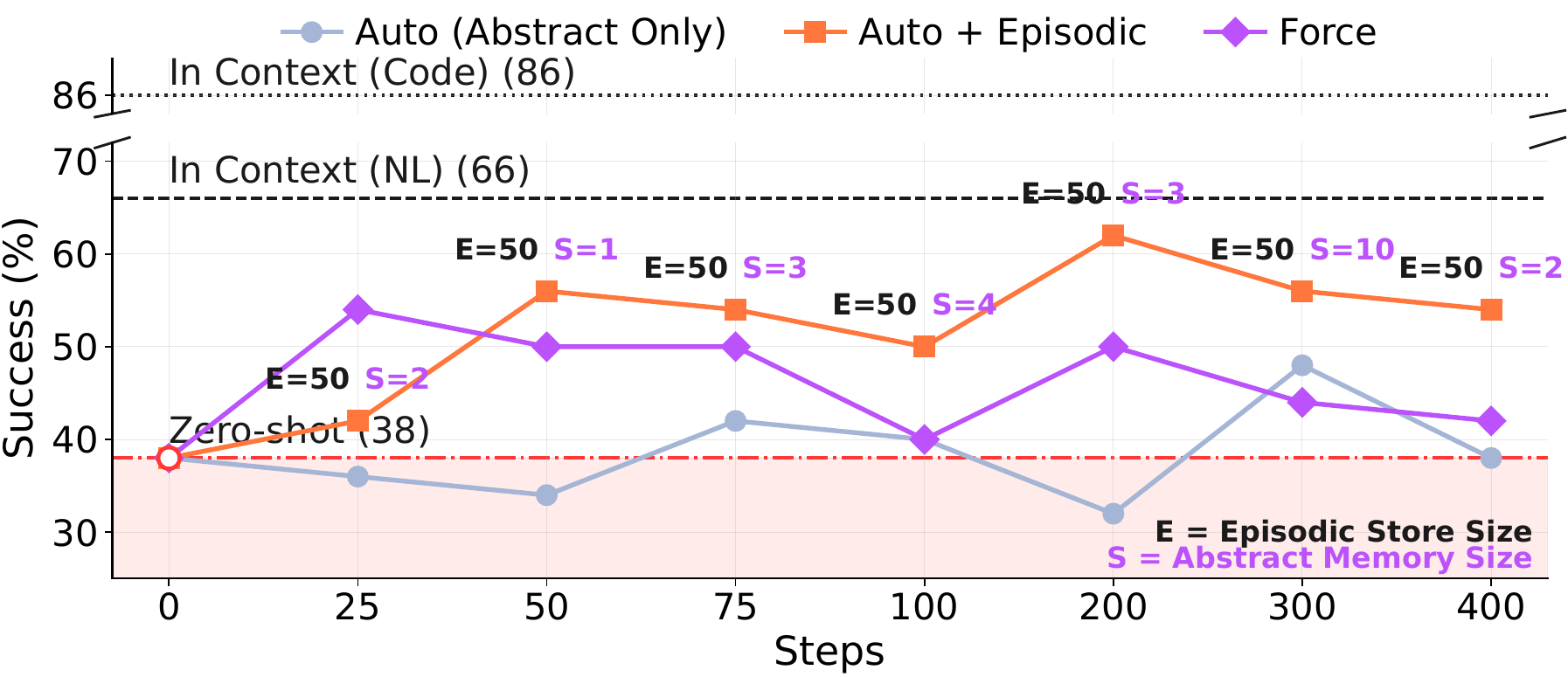}
\caption{\textbf{Training-time success across memory configurations.} Two in-context references --- raw code (\textit{In Context (Code)}, $86\%$) and natural language (\textit{In Context (NL)}, $66\%$) are in-context learning results on all problems of the same family.}
\label{fig:gt_400}
\end{subfigure}
\caption{\textbf{Forced abstraction underperforms retained episodes.} (\subref{fig:arc_management_strategies}) Across
memory-source ablations, episodic management matches or beats every consolidating mode. (\subref{fig:gt_400}) On the GT-400 stream, agents solve more problems with retained raw episodes than with forced abstractions.}
\label{fig:force_abstraction}
\end{figure}

\begin{wrapfigure}{r}{0.45\linewidth}
\begin{minipage}{\linewidth}
\centering
\vspace{-0.6em}
\includegraphics[width=\linewidth]{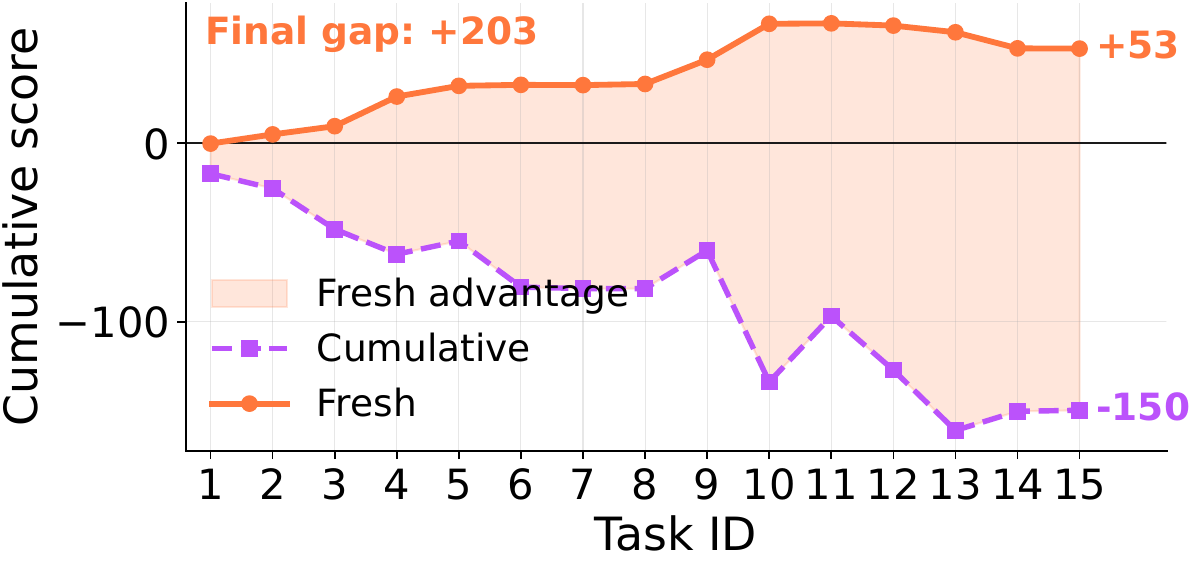}
\caption{\textbf{Per-task vs. accumulated consolidation on ScienceWorld.} Every additional task folded into a single store erodes the parts that supported the earlier ones.}
\label{fig:science_world_task_switch}
\vspace{-1cm}
\end{minipage}
\end{wrapfigure}

The same
interference dynamic shows up at the entire learning run level under task switching. On the 15-task ScienceWorld switch sequence, we compare two abstraction set-ups at each task $t$: \emph{Fresh}, which distills memories only on task $t$, against \emph{Cumulative}, which consolidates jointly over tasks $1\!\dots\!t$. Cumulative ends $+203$ points behind Fresh (Fig.~\ref{fig:science_world_task_switch}). We use LLM judge to label each memory entry and find Cumulative accumulates over-generalized memories at $\sim\!5\times$ Fresh's rate and garbage memories at $\sim\!20\times$, with both gaps widening monotonically across the 15 tasks (App.~\ref{app:judge-cumcount}, Fig.~\ref{fig:judge_cumcount}).

\subsection{Overfit}
\label{sec:overfit}

\begin{wrapfigure}{r}{0.42\linewidth}
\begin{minipage}{\linewidth}
\centering
\vspace{-0.6em}
\includegraphics[width=\linewidth]{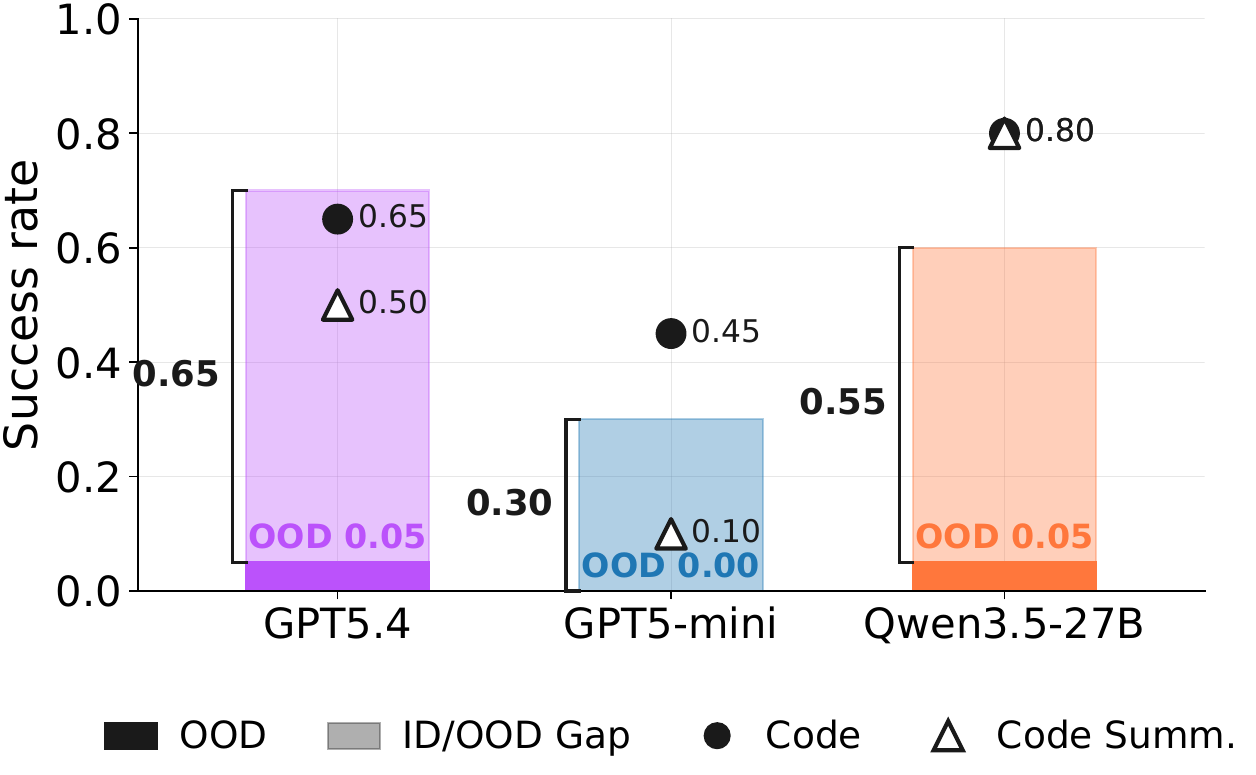}
\caption{Memory turns overspecified and hurt performance outside of the compressed questions.}
\label{fig:arc_overfit}
\vspace{-0.6em}
\end{minipage}
\end{wrapfigure}

A complementary failure mode emerges when the input distribution narrows rather than widens. In our ARC-AGI experiments, we repeatedly feed the model tasks drawn from a single strategy family across consolidation cycles. As the iteration count grows, the memory overfits to the specific instances it has seen: performance remains stable on exact repetitions of those problems but collapses on simple variations within the same strategy family (Fig.~\ref{fig:arc_overfit}; see \Cref{app:wrong-examples} for case studies).

The abstraction loop, given a narrow stream, abstracts toward the surface regularities of the observed instances rather than the underlying problem-solving strategy.

\subsection{What in the update does the damage?}
\label{sec:attribution}
Streamed consolidation could degrade for three separable reasons: \emph{initial compression error} (information lost the first time a batch is abstracted), \emph{cross-update conditioning} (earlier, imperfect abstractions in context biasing how the next batch is consolidated), and \emph{destructive rewriting} (editing entries in place erodes them cumulatively). We separate them by streaming the same batches in the same order while restricting only what the update step may see or do (Table~\ref{tab:attribution-ladder}). Forbidding in-place rewriting alone recovers most of the gap to raw episodes; additionally hiding the prior store, so earlier abstractions cannot condition later ones, recovers more still---yet none fully closes the gap, because information is already lost at the very first abstraction. All three mechanisms contribute and are separable: in-place rewriting is the most damaging single component, while pruning (delete without rewrite) actually helps. The full batch-size~$\times$~update-mode grid is in \Cref{app:attribution}.

\paragraph{Case Study}
\label{sec:case-study}
We ground the failures above in concrete memory artifacts across four environments: ARC-AGI strategy-memory pathologies (\Cref{app:wrong-examples}), ALFWorld memorycollapse (\Cref{sec:memory-case-studies}), WebShop AWM collapse-by-accretion (\Cref{sec:webshop-awm-collapse}), and ScienceWorld abnormal-memory categories (\Cref{sec:scienceworld-abnormal}). A \emph{memory surgery} ablation isolates their content-level bite: removing a single workflow (W8) from a WebShop consolidated memory raises pass rate (\Cref{sec:surgery-awm-w8}), since W8 biases the agent toward \texttt{click[Next >]} dead loops at the expense of \texttt{click[Buy Now]}.

\section{Conclusion}
\label{sec:conclusion}

Persistent memory is meant to let LLM agents move beyond static competence: experience is stored, compressed into reusable lessons, and carried forward. We identify an issue that may undermine this promise --- across agent benchmarks and ARC-AGI Stream, continuously updated textual memory can become less useful as experience accumulates; in the cleanest case, an agent becomes worse on the very problems its memory was built from.
\begin{table}[t]
\centering
\small
\setlength{\tabcolsep}{7pt}
\renewcommand{\arraystretch}{1.2}
\begin{tabular}{@{}llc@{}}
\toprule
\textbf{Streaming variant} & \textbf{What it removes vs.\ default} & \textbf{Accuracy} \\
\midrule
\rowcolor{cellbad} Default consolidation (in-place rewrite) & --- & $43.2$ \\
Append-only, prior store visible (A)      & in-place rewriting                 & $50.0$ \\
Add\,+\,delete, no in-place edit (C)      & in-place rewriting (keeps pruning) & $64.0$ \\
Append-only, prior store hidden (B)       & rewriting \emph{and} conditioning  & $70.0$ \\
\midrule
\rowcolor{cellgood} Raw trajectories (no consolidation) & all consolidation & $76.6$ \\
\bottomrule
\end{tabular}
\caption{\textbf{The closer to raw episodes, the better---and in-place rewriting is the most damaging single step.}
Accuracy on ARC-AGI Stream (GPT-5.4, batch size~1, forced updates, step~400) as we strip components of the streamed update. Forbidding in-place rewriting (A) recovers $+6.8$ over the default; also hiding the prior store so earlier abstractions cannot bias later ones (B) recovers $+26.8$; adding pruning to append-only (C) recovers $+20.8$. All remain below simply retaining the raw trajectories the consolidator distills from. Full grid in \Cref{app:attribution}.}
\label{tab:attribution-ladder}
\end{table}

The failure lies in the consolidation loop itself. Current LLMs do not reliably decide which episodes belong together, which distinctions should survive compression, or when an experience should remain episodic. They misgroup trajectories, overgeneralize lessons, and overfit to narrow streams. Because each update rewrites the products of earlier updates, small abstraction errors compound into faulty memory.

These findings suggest that raw episodes should be treated as first-class evidence, not disposable material to be compressed away. Abstraction should be selective, delayed, and grounded in recoverable trajectories. Until agents can control when and how to consolidate experience, continuously updated textual memory should be treated not as a reliable engine of self-improvement, but as a fragile mechanism that can make more experience produce worse memory.

\section*{Acknowledgment}

This work is in part supported by the NSF grant CHE-2505932, an Amazon AICE Award, gift funding from AI2, and a grant from Coefficient Giving.
This research used
both the DeltaAI advanced computing and data resource, which is supported by the National Science
Foundation (award OAC 2320345) and the State
of Illinois, and the Delta advanced computing and
data resource which is supported by the National
Science Foundation (award OAC 2005572) and the
State of Illinois. Delta and DeltaAI are joint efforts
of the University of Illinois Urbana-Champaign
and its National Center for Supercomputing Applications.

The first author (D.Z.) would like to thank Compute provided by Lambda Inc., and useful inputs from Pengcheng Jiang, Zidi Xiong, Zhenting Qi and Zhanyang Jin.

\small
\bibliography{references}

\newpage

\appendix

\providecommand{\beatmark}{\textcolor{textbad}{\scriptsize\,$(\text{--})$}}
\begin{table}[t]
\begin{minipage}{\linewidth}
\centering
\scriptsize
\setlength{\tabcolsep}{3pt}
\renewcommand{\arraystretch}{1.20}
\resizebox{\textwidth}{!}{%
\begin{tabular}{@{}l c cccccc cccc@{}}
\toprule
& & \multicolumn{6}{c}{\textbf{Trajectory Logs (Summarized)}} & & & & \\
\cmidrule(lr){3-8}
\textbf{Backbone} & \textbf{NoMem}
& \textbf{All} & \textbf{64} & \textbf{32} & \textbf{16}
& \textbf{Incr.\ Keep} & \textbf{One-Pass Keep}
& $\mathbf{ACE_{GT}}$ & \textbf{ACE} & \textbf{AWM} & \textbf{DC} \\
\midrule
\rowcolor{cellbase}
GPT-5.4-Mini
              & $18$
              & $47$
              & $44$
              & $49$
              & $\mathbf{51}$
              & $40$
              & $41$
              & \textcolor{textmute}{---}
              & $00^{*}$
              & $23$\beatmark
              & $03^{*}$ \\
\rowcolor{tabband}
GPT-5-Mini
              & $39$
              & $53$
              & $41$
              & $\mathbf{63}$
              & $49$
              & $49$
              & $50$
              & \textcolor{textmute}{---}
              & \textcolor{textmute}{---}
              & $59$\beatmark
              & \textcolor{textmute}{---} \\
Claude-Haiku-4.5
              & $19$
              & $59$
              & $\mathbf{63}$
              & $55$
              & $49$
              & $54$
              & $54$
              & \textcolor{textmute}{---}
              & \textcolor{textmute}{---}
              & $32$\beatmark
              & \textcolor{textmute}{---} \\
\bottomrule
\end{tabular}%
}
\caption{\textbf{Trajectory Log as a Strong Baseline (WebShop).}
Values are mean reward $\times 100$.
\textbf{NoMem} (no-memory baseline);
\textbf{Trajectory Logs (Summarized)} groups append-only summarized rollouts retrieved as in-context exemplars---\textbf{All}/\textbf{64}/\textbf{32}/\textbf{16} use the full pool or downsample to that many shots, while \textbf{Incr.\ Keep} and \textbf{One-Pass Keep} are LLM-curated retain/delete variants (streamed and one-pass, respectively);
$\mathbf{ACE_{GT}}$ (ACE with ground-truth labels), \textbf{ACE} (ACE without GT),
\textbf{AWM} (Agent Workflow Memory), \textbf{DC} (Dynamic Cheatsheet).
A red \textcolor{textbad}{$(\text{--})$} marks any non-trajectory-log approach that is beaten by at least one trajectory-log baseline in the same row.
Best per row in \textbf{bold}. Cells marked \textcolor{textmute}{---} are not measured for that backbone. The results with $^{*}$ are evaluted by model GPT-5-Mini. Then we observed that its performance were near to 0, so we did not do remaining ACE or DC experiments on webshop.} 
\label{tab:strong_baseline_webshop}
\end{minipage}
\end{table}

\begin{table}
\centering
\footnotesize
\setlength{\tabcolsep}{3pt}
\renewcommand{\arraystretch}{1.15}
\vspace{-0.6em}
\begin{tabular}{l c c c c}
\toprule
& \multicolumn{2}{c}{$|B|{=}1$} & \multicolumn{2}{c}{$|B|{=}8$} \\
\cmidrule(lr){2-3} \cmidrule(lr){4-5}
& Auto & Force & Auto & Force \\
\midrule
Avg.\ Covered & $1.33$ & $1.67$ & $2.17$ & \cellcolor[HTML]{FFDBCC}$\mathbf{5.00}$ \\
\rowcolor{tabband}
Avg.\ Fused   & $1.00$ & $1.04$ & $1.06$ & \cellcolor[HTML]{FFDBCC}$\mathbf{1.20}$ \\
Buffer Size & $3.33$ & --- & \cellcolor[HTML]{f4e7fb}$50.00$ & --- \\
\bottomrule
\end{tabular}
\caption{\textbf{Memory Coverage.}}
\label{tab:arc-family-obs}
\vspace{-0.6em}
\end{table}

\etocsetlocaltop.toc{part}
\etocsetnexttocdepth{subsection}
\localtableofcontents
\clearpage

\section{Family and skill taxonomy [ARC-AGI]}
\label{app:taxonomy}

This section defines the two terms ``family'' and ``skill'' and
lists the six families and seven skills used in the ARC-AGI experiments.

\subsection{Definitions}
\label{subsec:taxonomy-defs}

Our task generator decomposes an ARC-style task along two
independent axes:
\begin{itemize}[leftmargin=*,topsep=2pt,itemsep=2pt]
  \item \textbf{Family}: a rule that selects the subset of
        connected non-background objects in the input grid that
        participates in the task's transformation.
  \item \textbf{Skill}: a fixed transformation applied to each
        selected object.
\end{itemize}
The two axes are orthogonal: the same family can be paired with
different skills, and the same skill can be paired with different
families. Some families are adapted from concept groups in
ConceptARC; for example, \texttt{detect\_inside\_frame\_relation}
corresponds to the \textsc{InsideOutside} concept group.

\subsection{Family catalog}
\label{subsec:taxonomy-strategies}

The six families span five distinct selection criteria: two are based
on intrinsic per-object attributes (color, size extremum), one on a
global conditional cue (a marker cell that triggers a branch), one
on a cross-object relation (shape-frequency mode), one on a spatial
containment relation (objects strictly inside a frame), and one on a
cross-grid structural property (a sub-scene of a two-panel input).
The chosen skill is then applied uniformly to whichever objects the
family selects.

\begin{table}[!htbp]
  \centering
  \small
  \caption{The six families evaluated in this paper, grouped by what
    they select on. The rightmost column gives the rule by which each
    family picks the object subset that participates in the task's
    transformation. The first column reproduces the
    \texttt{strategy\_id} field used to identify each family in the
    released code and data.}
  \label{tab:strategies}
  \begin{tabular}{@{}l l p{0.42\linewidth}@{}}
    \toprule
    \textbf{strategy\_id} & \textbf{Selects by} & \textbf{Definition} \\
    \midrule
    \texttt{detect\_color\_property}          & intrinsic (color)         & objects of a target color (inferred from the examples)   \\
    \texttt{detect\_largest\_objects\_select} & intrinsic (size extremum) & the largest object(s) by cell count (ties all selected) \\
    \texttt{detect\_key\_marker\_rule}        & global conditional cue    &  all-or-none global selection gated by the upper-left cell's color \\
    \texttt{group\_by\_shape\_then\_select}   & cross-object relation     & objects in the shape-frequency mode group \\
    \texttt{detect\_inside\_frame\_relation}  & spatial containment       & objects strictly inside a hollow frame \\
    \texttt{compose\_horizontal}              & cross-grid structure      & the left or right sub-scene of a two-panel input \\
    \bottomrule
  \end{tabular}
\end{table}

\begin{figure}[!htbp]
  \centering
  \begin{subfigure}[t]{0.45\linewidth}
    \includegraphics[width=\linewidth]{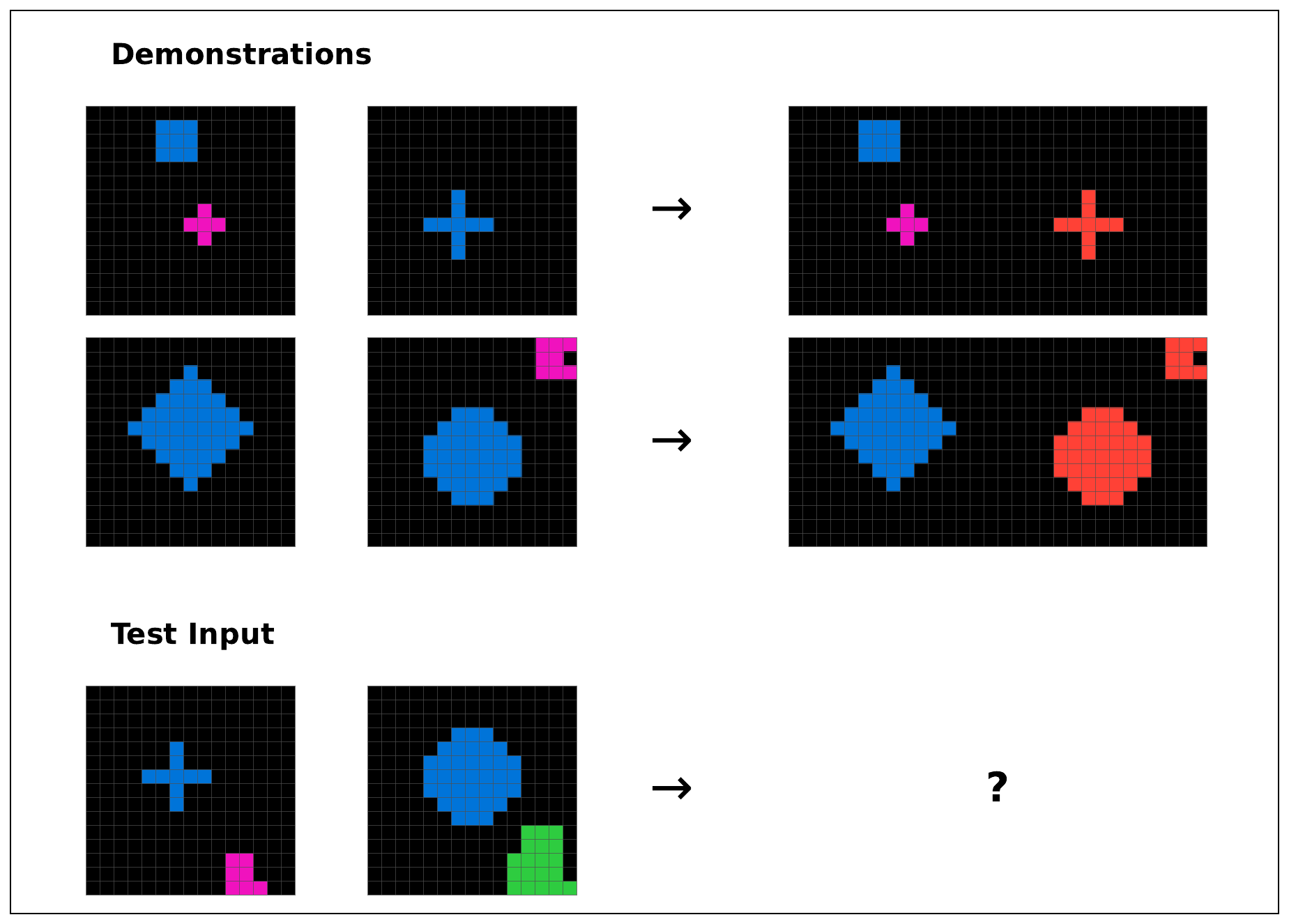}
    \caption{Compose Horizontal.}
    \label{fig:strategies_compose}
  \end{subfigure}\hfill
  \begin{subfigure}[t]{0.27\linewidth}
    \includegraphics[width=\linewidth]{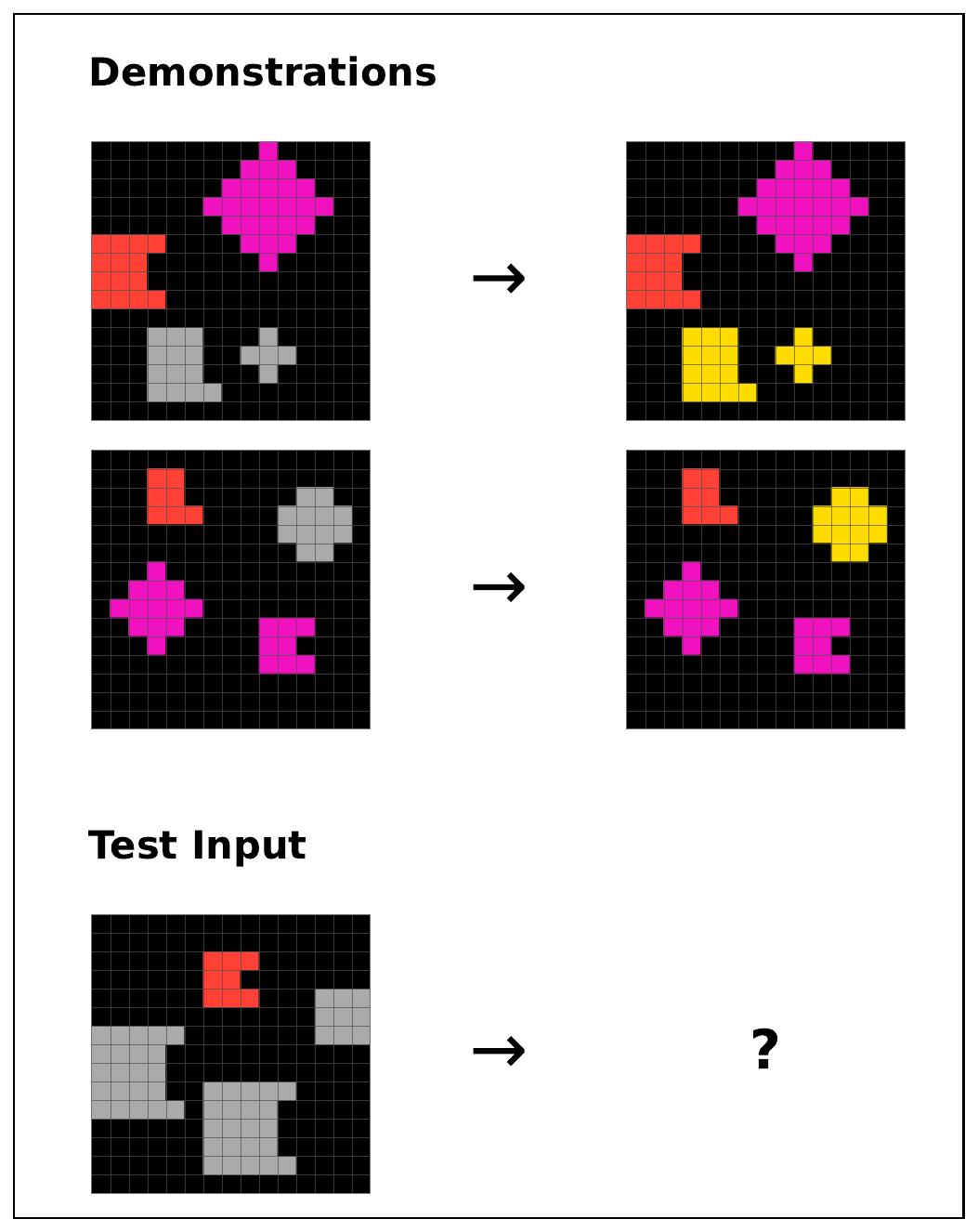}
    \caption{Color Property.}
    \label{fig:strategies_color}
  \end{subfigure}\hfill
  \begin{subfigure}[t]{0.27\linewidth}
    \includegraphics[width=\linewidth]{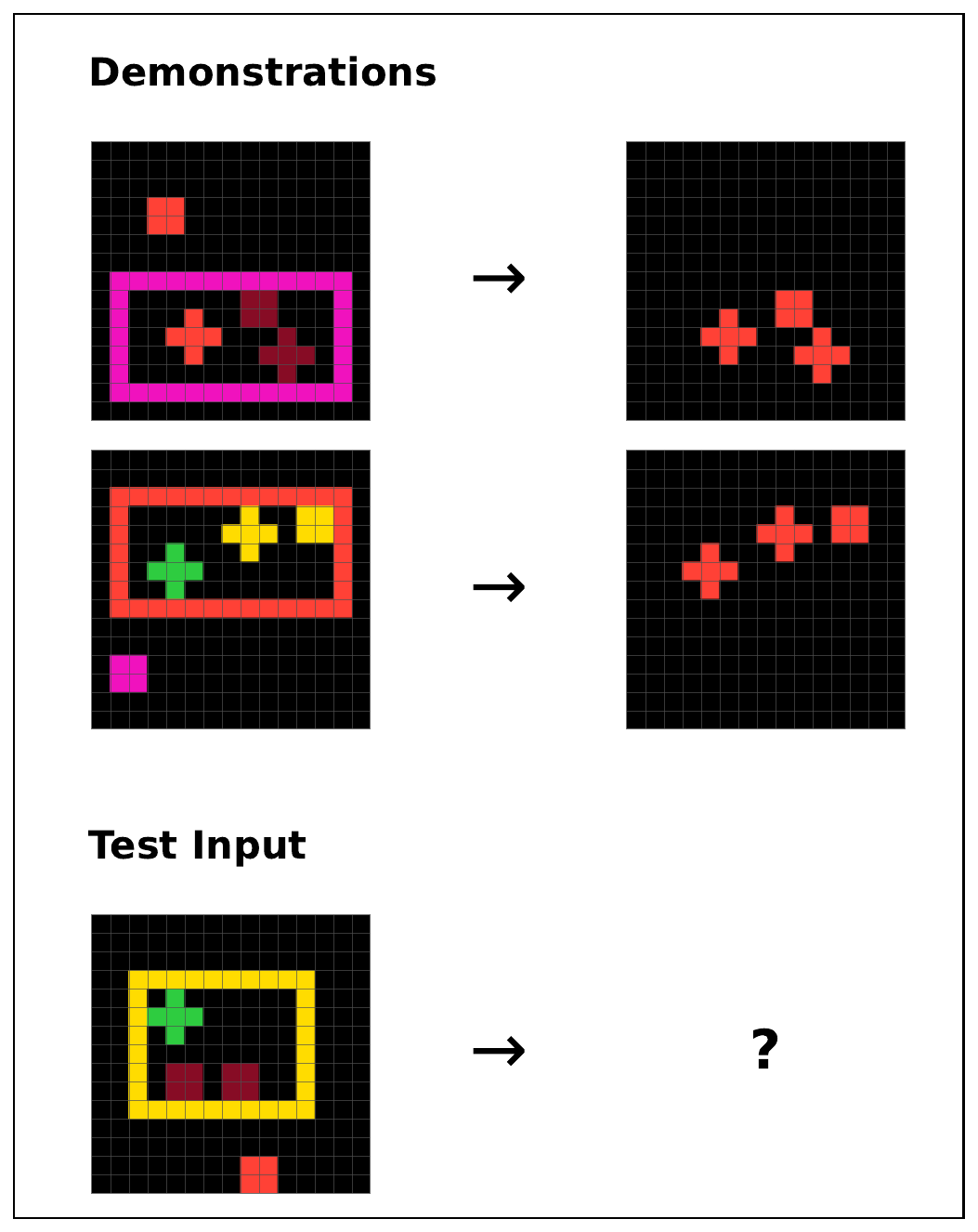}
    \caption{Inside Frame.}
    \label{fig:strategies_frame}
  \end{subfigure}

  \vspace{0.5em}

  \begin{subfigure}[t]{0.32\linewidth}
    \includegraphics[width=\linewidth]{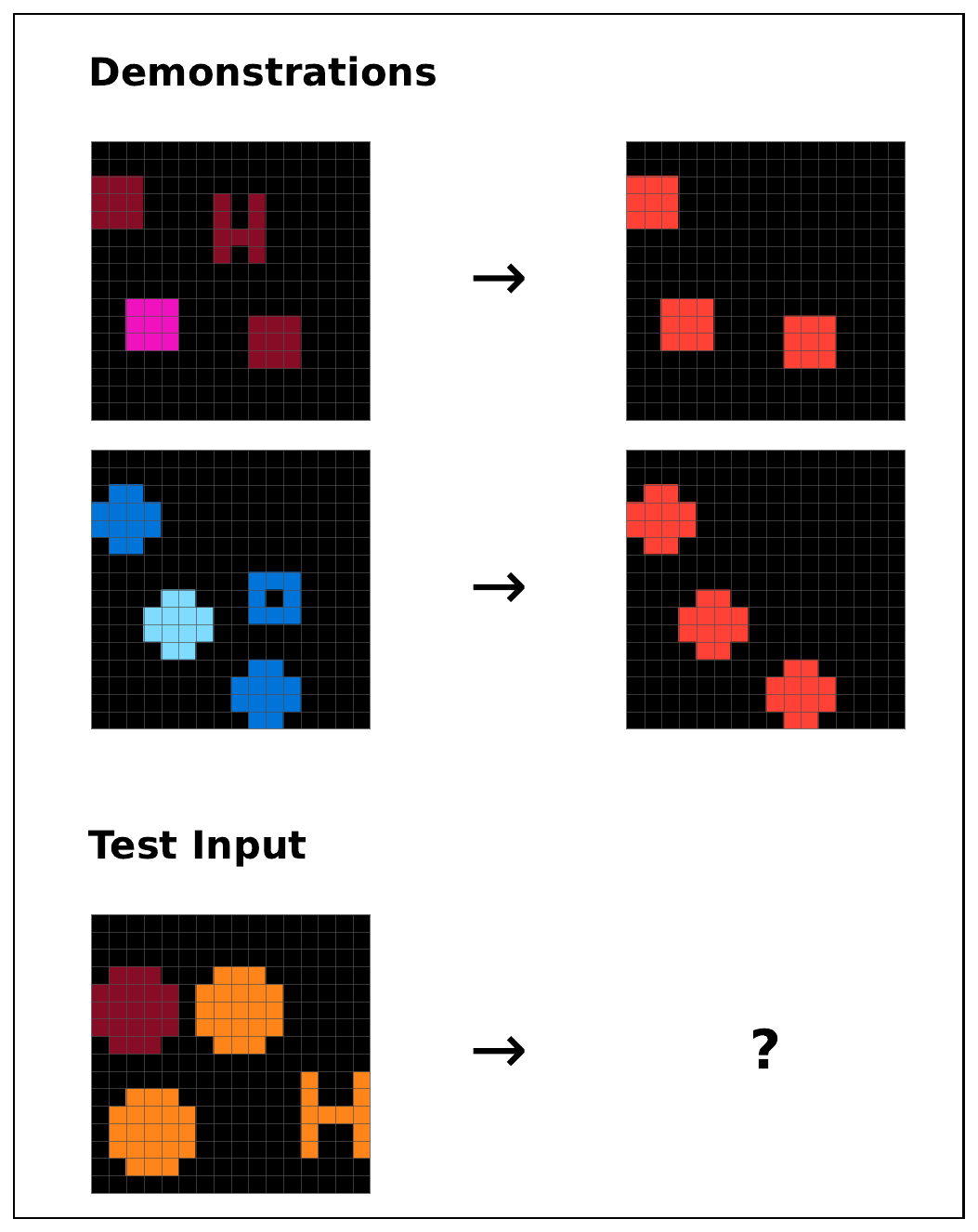}
    \caption{Group by Shape.}
    \label{fig:strategies_group}
  \end{subfigure}\hfill
  \begin{subfigure}[t]{0.32\linewidth}
    \includegraphics[width=\linewidth]{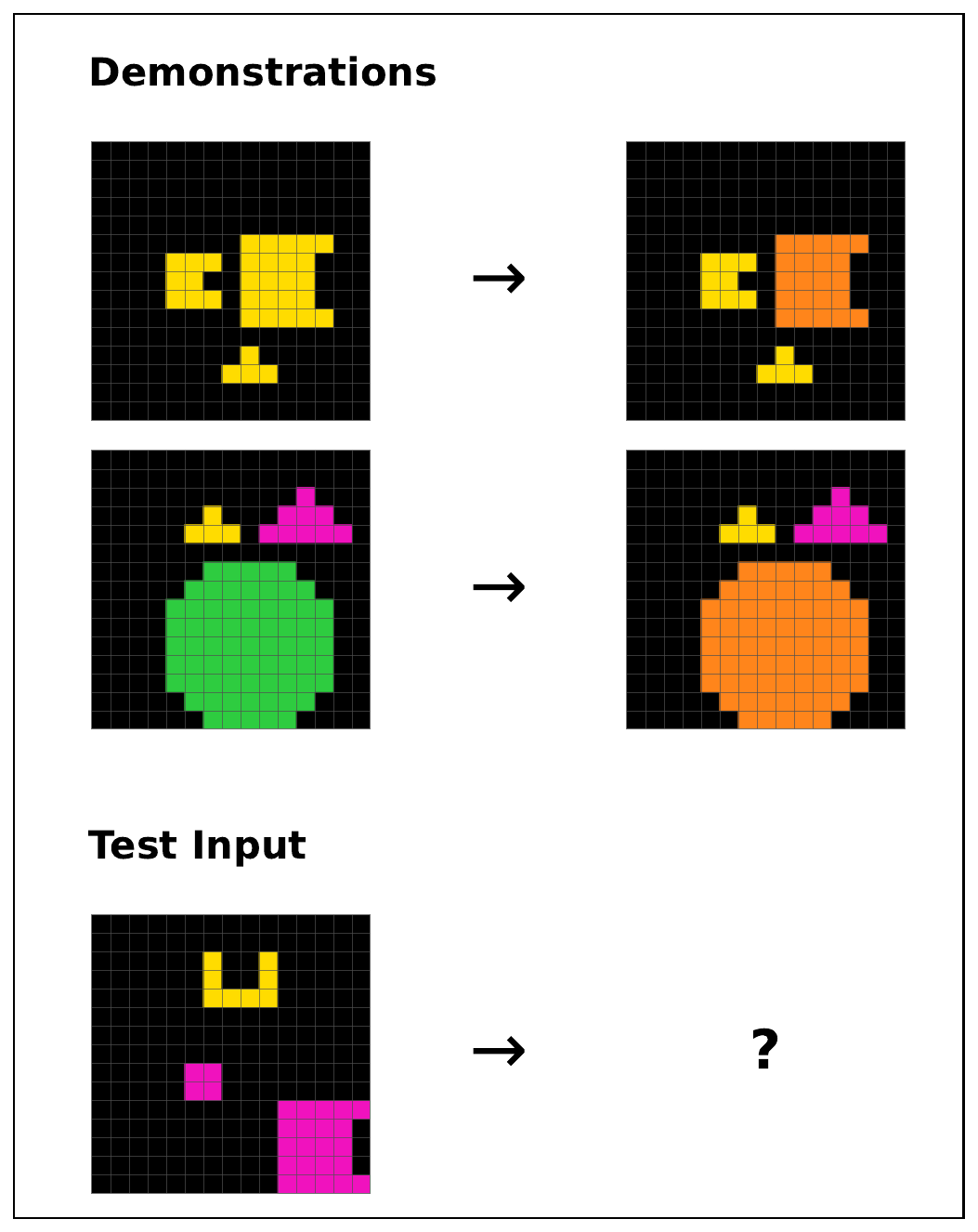}
    \caption{Largest Object.}
    \label{fig:strategies_largest}
  \end{subfigure}\hfill
  \begin{subfigure}[t]{0.32\linewidth}
    \includegraphics[width=\linewidth]{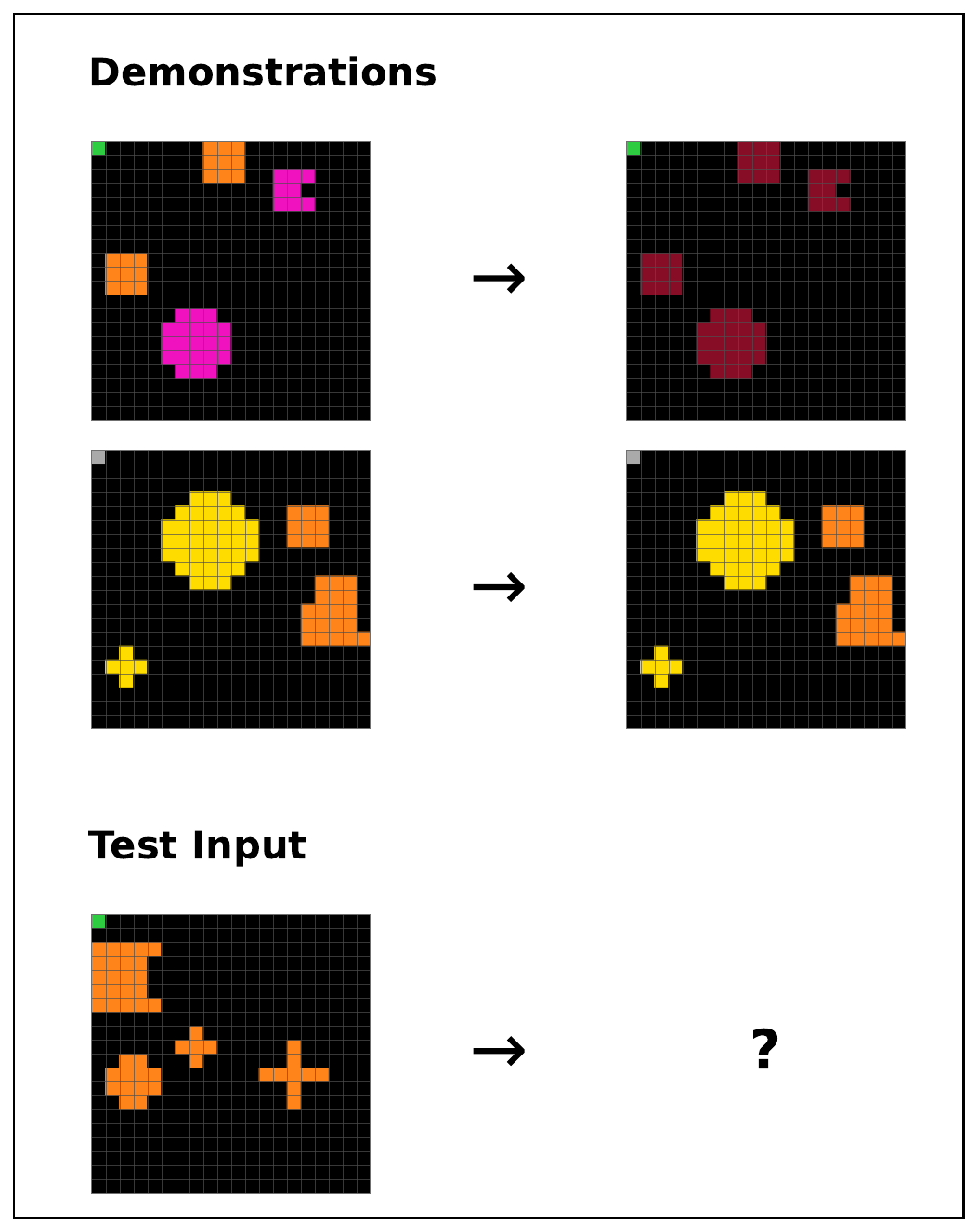}
    \caption{Key Marker.}
    \label{fig:strategies_marker}
  \end{subfigure}
  \caption{One task per family, all rendered with the skill axis
    fixed to \texttt{recolor}. Each panel shows two demonstration
    input--output pairs and a held-out test input. Recolored objects
    in the output identify which objects the family selects.
    Compose Horizontal takes two input grids per example and
    transforms the left sub-scene; the other five families operate
    on a single input. The Key Marker panel additionally illustrates
    conditional branching: when the upper-left marker color matches
    the trigger value, objects are recolored; otherwise the output
    equals the input. Cell colors follow the 10-color ARC palette;
    color~$0$ (black) denotes background and colors~$1$--$9$ denote
    object pixels.}
  \label{fig:strategies}
\end{figure}

\Cref{fig:strategies} shows one task per family under a fixed
skill of \texttt{recolor}: recolored objects in the output identify
the family's selection by color contrast.

\FloatBarrier

\subsection{Skill catalog}
\label{subsec:taxonomy-skills}

Each skill applies a fixed transformation to every object in the
family's selection.

\begin{table}[!htbp]
  \centering
  \small
  \caption{The seven skills evaluated in this paper. Each skill is
    applied uniformly to every object selected by the family.}
  \label{tab:skills}
  \begin{tabular}{@{}l p{0.65\linewidth}@{}}
    \toprule
    \textbf{skill\_id} & \textbf{Definition} \\
    \midrule
    \texttt{keep}             & Retain selected objects unchanged; erase non-selected objects to background. \\
    \texttt{border}           & Draw a one-cell border around each selected object. \\
    \texttt{recolor}          & Repaint each selected object in a single new color. \\
    \texttt{translate}        & Shift each selected object by a fixed offset. \\
    \texttt{flip\_horizontal} & Mirror each selected object within its bounding box. \\
    \texttt{mark\_center}     & Mark the centroid cell of each selected object with a marker color. \\
    \texttt{hollow}           & Erase the interior of each selected object, leaving only its outer border. \\
    \bottomrule
  \end{tabular}
\end{table}

\begin{figure}[!htbp]
  \centering
  \begin{subfigure}[t]{0.245\linewidth}
    \includegraphics[width=\linewidth]{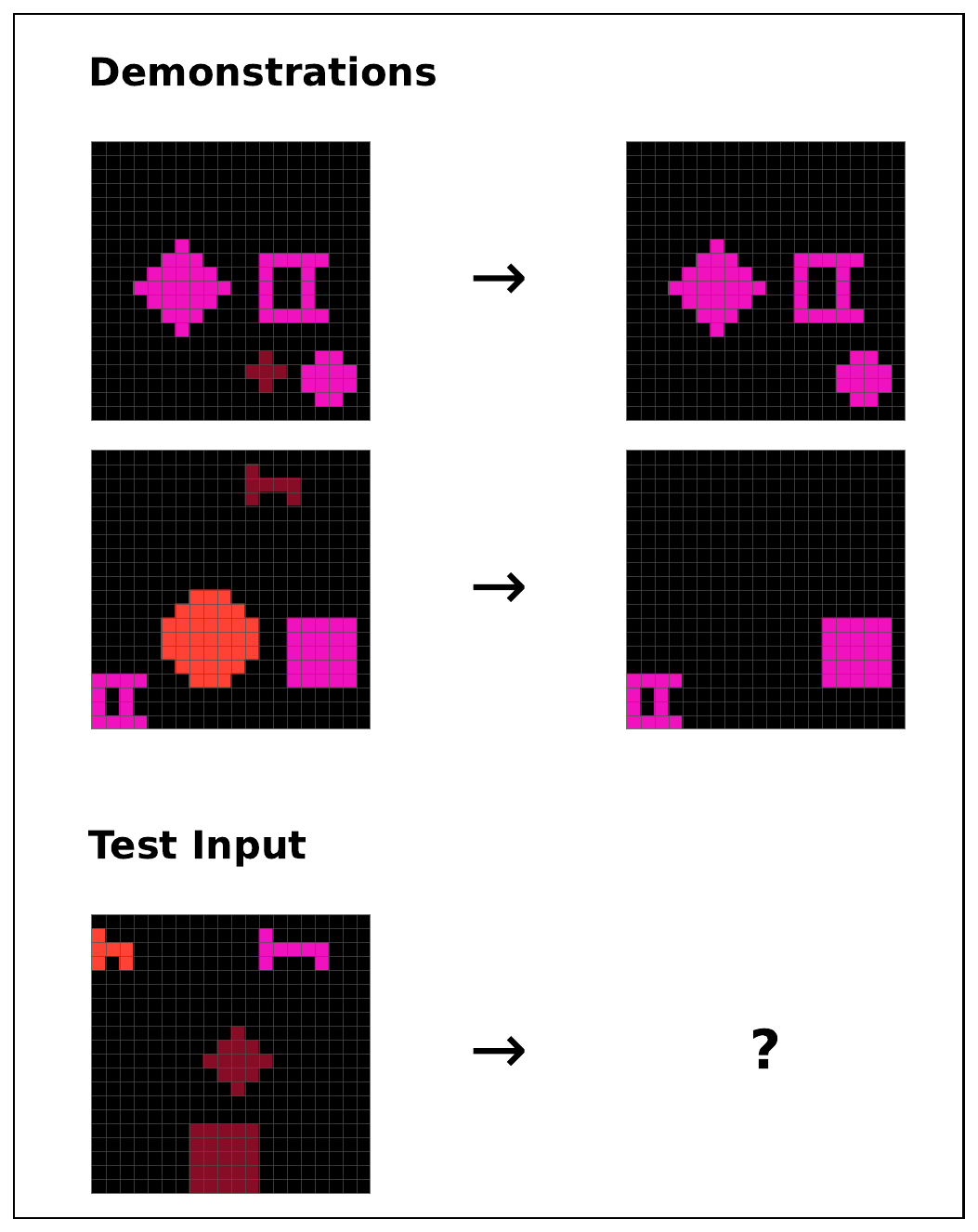}
    \caption{Keep.}
    \label{fig:skills_keep}
  \end{subfigure}\hfill
  \begin{subfigure}[t]{0.245\linewidth}
    \includegraphics[width=\linewidth]{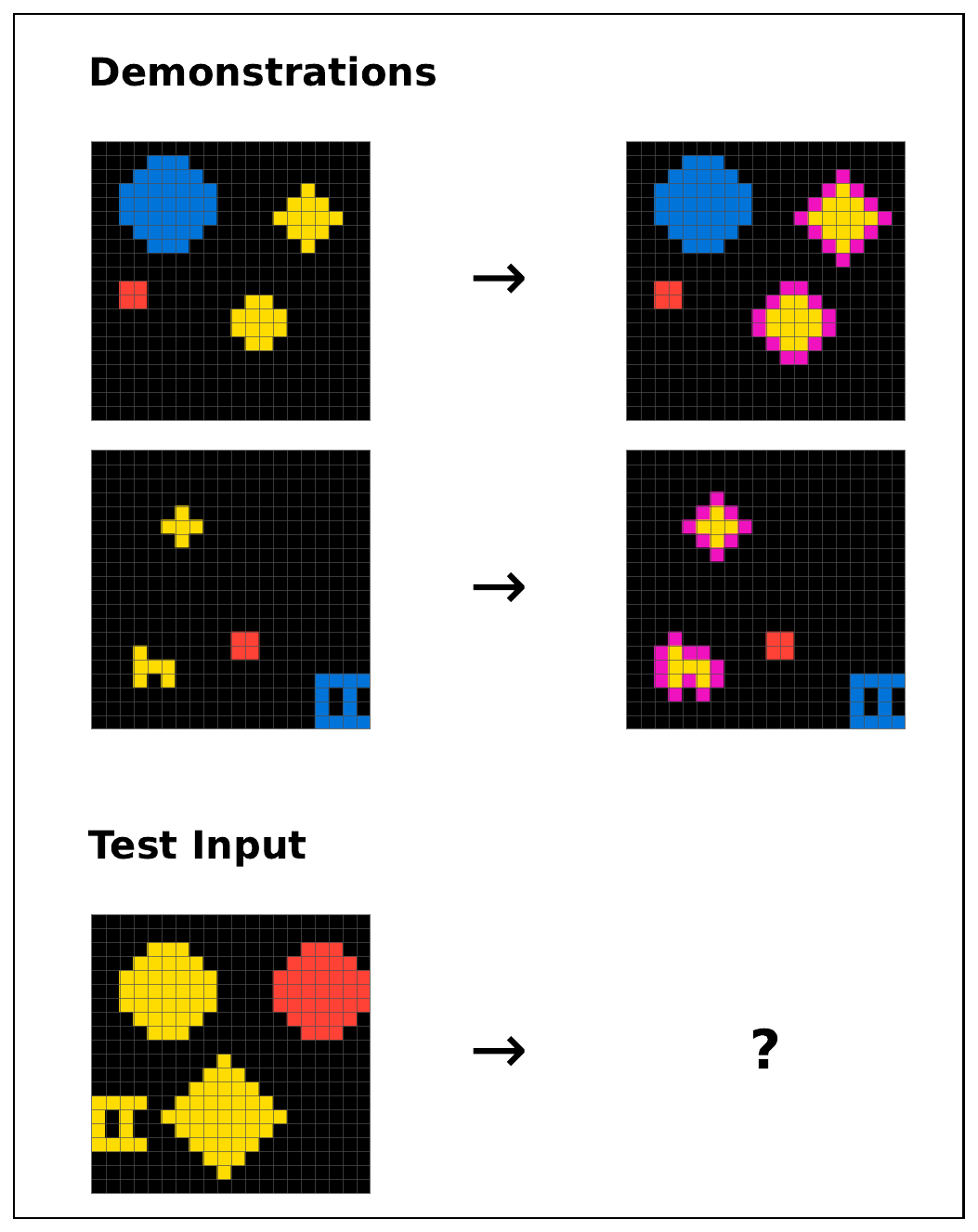}
    \caption{Border.}
    \label{fig:skills_border}
  \end{subfigure}\hfill
  \begin{subfigure}[t]{0.245\linewidth}
    \includegraphics[width=\linewidth]{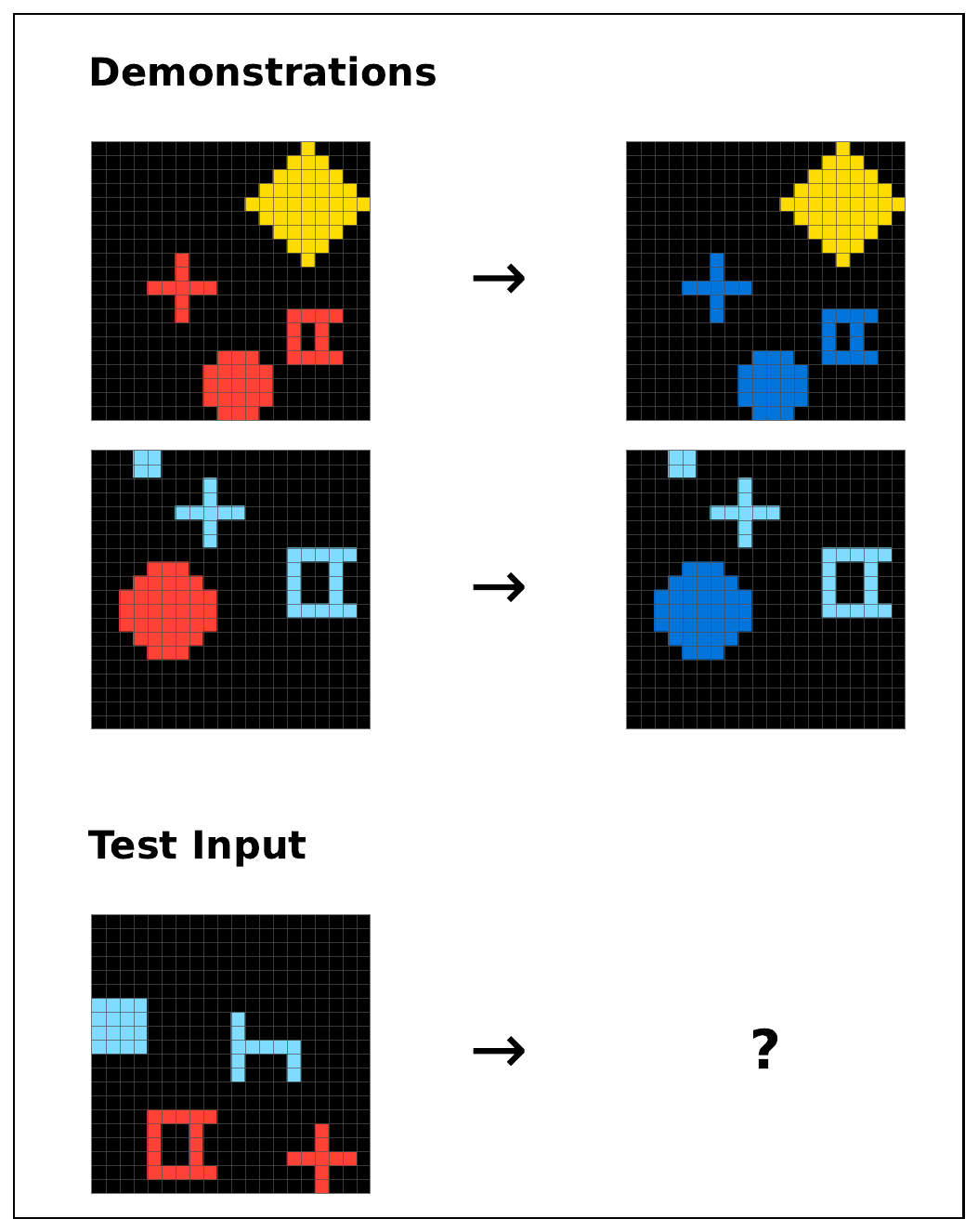}
    \caption{Recolor.}
    \label{fig:skills_recolor}
  \end{subfigure}\hfill
  \begin{subfigure}[t]{0.245\linewidth}
    \includegraphics[width=\linewidth]{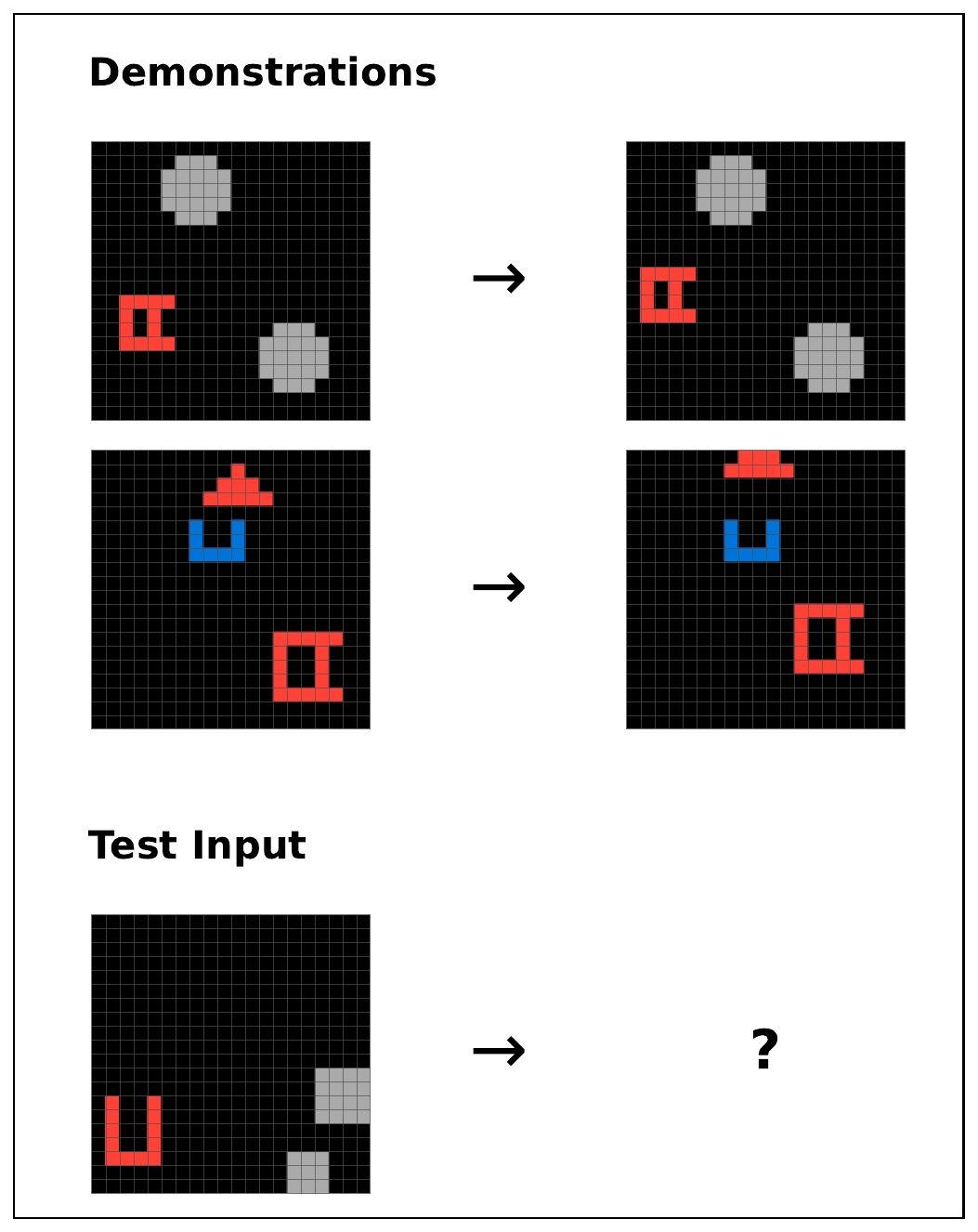}
    \caption{Translate.}
    \label{fig:skills_translate}
  \end{subfigure}

  \vspace{0.5em}

  \begin{subfigure}[t]{0.328\linewidth}
    \includegraphics[width=\linewidth]{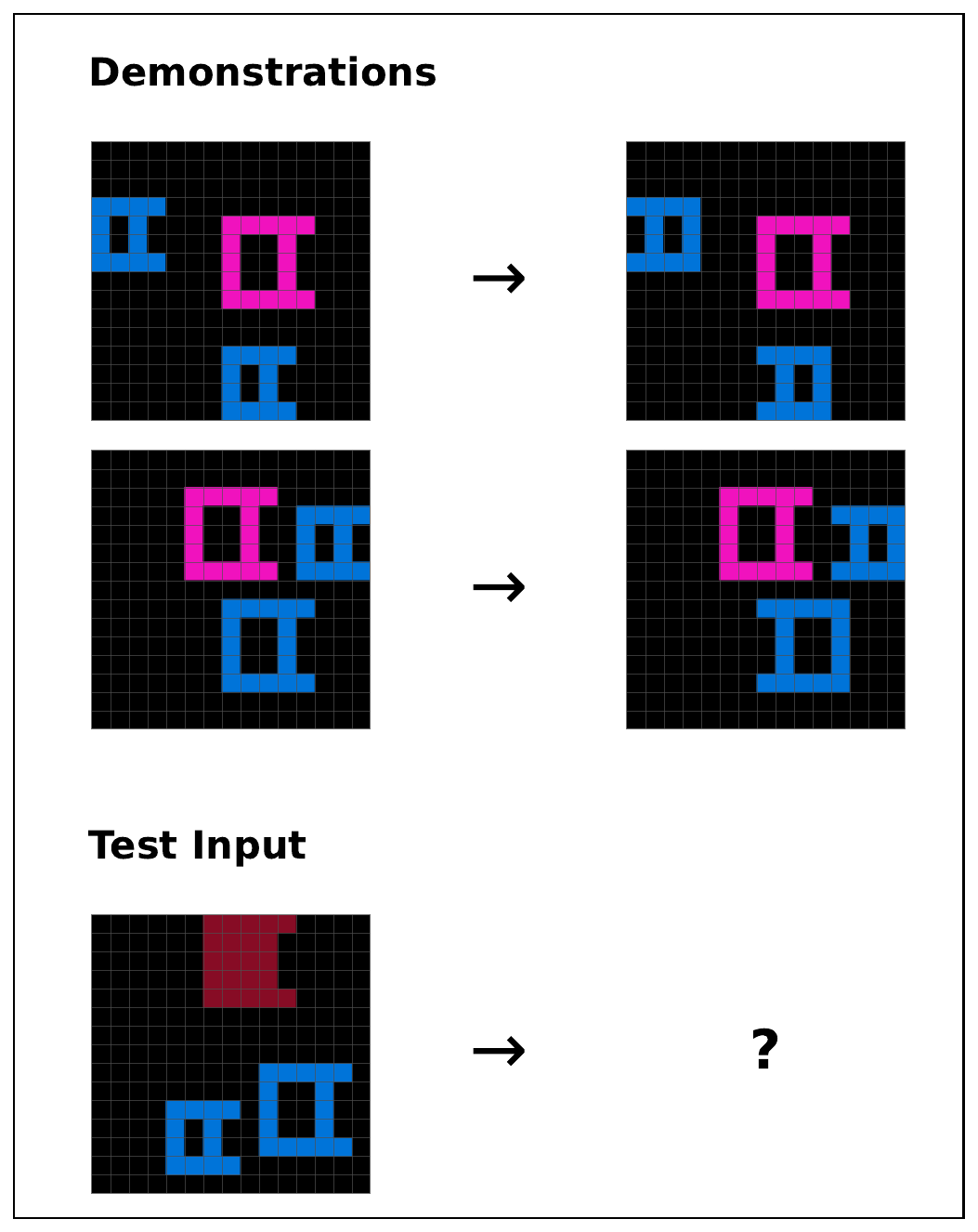}
    \caption{Flip Horizontal.}
    \label{fig:skills_flip}
  \end{subfigure}\hfill
  \begin{subfigure}[t]{0.328\linewidth}
    \includegraphics[width=\linewidth]{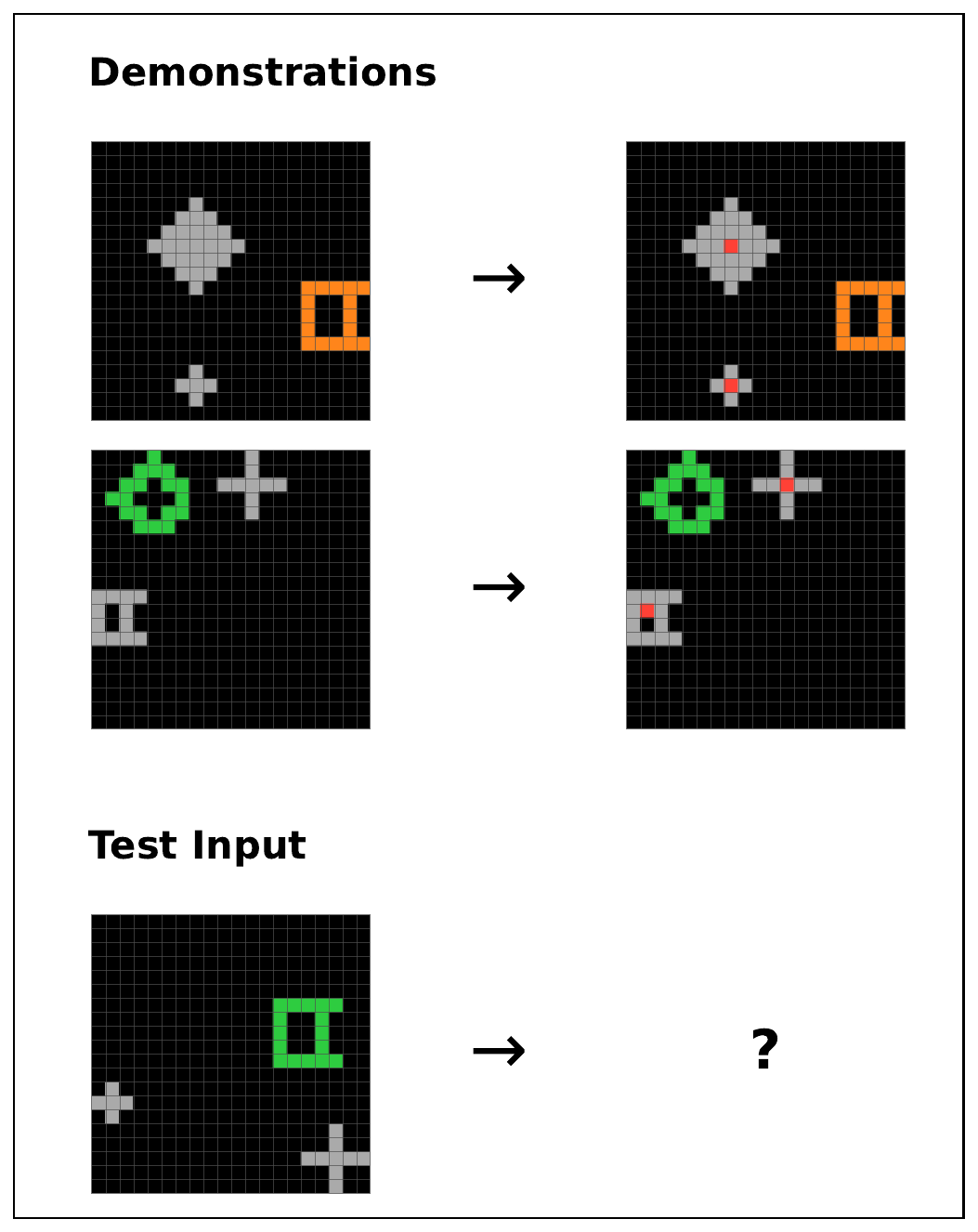}
    \caption{Mark Center.}
    \label{fig:skills_mark}
  \end{subfigure}\hfill
  \begin{subfigure}[t]{0.328\linewidth}
    \includegraphics[width=\linewidth]{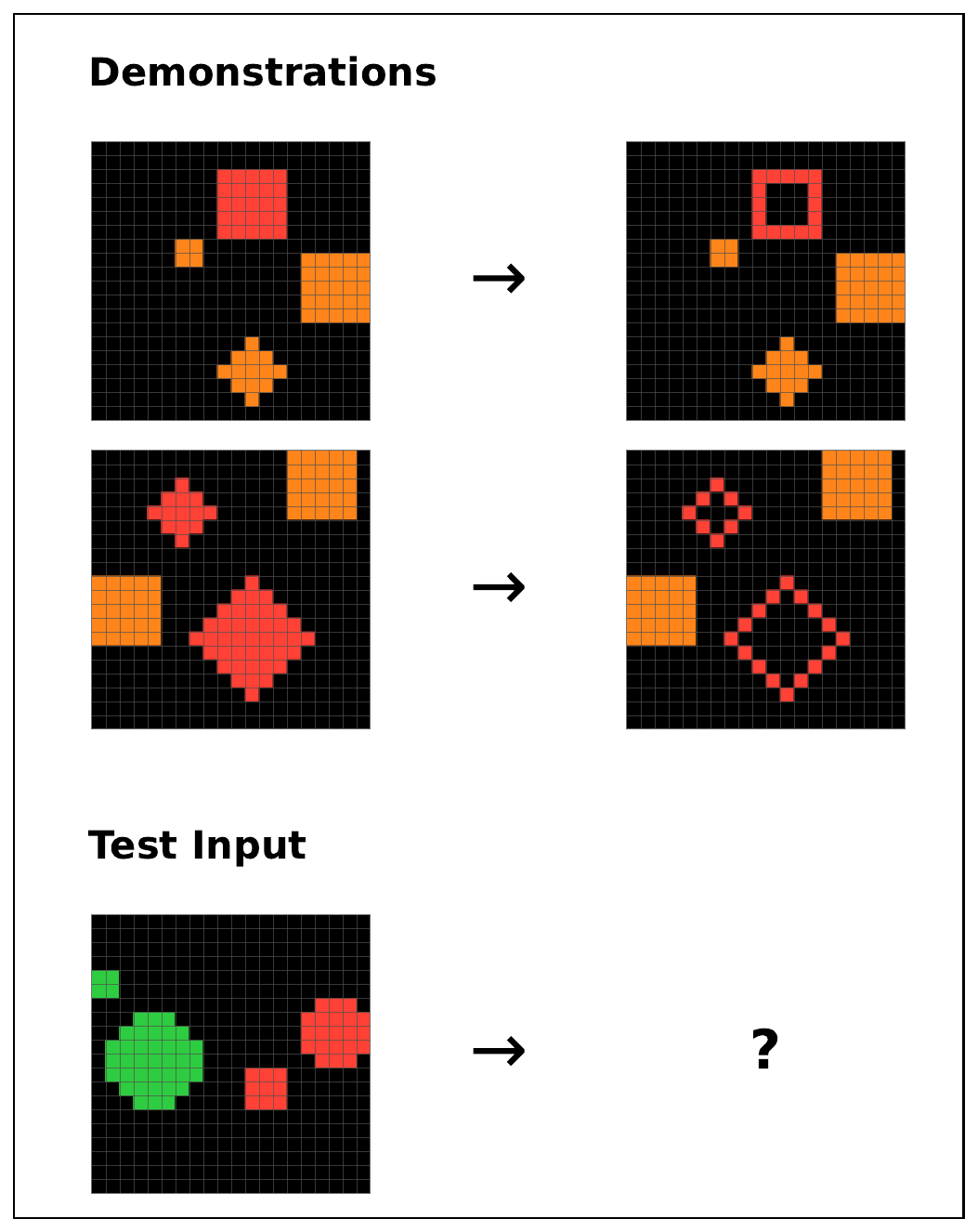}
    \caption{Hollow.}
    \label{fig:skills_hollow}
  \end{subfigure}
  \caption{One task per skill, all rendered with the family axis
    fixed to \texttt{detect\_color\_property}. Each panel shows two
    demonstration input--output pairs and a held-out test input. The
    family selects objects whose color matches a target property
    demonstrated by the examples; the skill then applies its
    transformation to those objects only.}
  \label{fig:skills}
\end{figure}

\Cref{fig:skills} shows the per-skill effect under a fixed
family of \texttt{detect\_color\_property}, where the selection is
visually simple (objects whose color matches a target predicate
demonstrated by the examples), isolating the skill's per-object
action.

\FloatBarrier

\section{Prompts [ARC-AGI]}
\label{app:prompts}

The experiments use three prompts: a solver prompt issued once per task, a consolidator decision prompt that selects an action on the recent history buffer after each passing step, and a consolidator extraction prompt that produces a replacement strategy memory from a selected subset of history. To make every field concrete we reproduce one full step from a real run in which the current solve passes; the history buffer carries two prior passing trajectories and strategy memory holds one entry distilled at an earlier step. Each prompt is shown together with the response it elicited. Where long content repeats across prompts (the eight extra IO examples for the current task; the 8 tool-buffer helper bodies; grids and solve code already shown above), we mark the elision in square brackets; everything else is byte-for-byte from the run logs.

\subsection{Solver prompt}
\label{prompt:solver}

The solver prompt is issued once per task at the start of the step. It bundles the 10 IO examples for the current task, the existing strategy memory, the tool-buffer helper signatures, and the recent history buffer into one call.

\paragraph{Input.}
\begin{tcblisting}{
  promptbox,
  title={Solver prompt (Examples 3--10 and helper bodies elided)},
  listing only,
  listing options={
    basicstyle=\ttfamily\scriptsize,
    breaklines=true,
    columns=fullflexible,
    keepspaces=true,
    showstringspaces=false
  }
}
You are an expert Python programmer.
**Current Task:**
Write a Python function `solve` that passes the following examples.
Use the signature `def solve(grid):`.
Example 1:
Input:
0 0 0 0 0 0 0 0 0 0 0 0 0 0 0 0 0 0 0 0
0 0 0 0 0 0 0 0 0 0 0 0 0 0 0 0 0 0 0 0
0 0 0 0 0 0 0 0 0 0 0 0 0 0 0 0 0 0 0 0
0 0 0 0 0 0 0 0 0 0 0 0 0 0 0 0 0 0 0 0
0 0 0 0 0 0 2 2 2 2 2 2 2 2 2 2 2 0 0 0
0 0 0 0 0 0 2 0 0 0 0 0 0 0 0 0 2 0 0 0
0 0 0 0 0 0 2 0 0 0 0 0 0 0 0 0 2 0 0 0
0 0 0 0 0 0 2 7 7 0 0 0 0 0 7 0 2 0 0 0
0 0 0 0 0 0 2 7 7 0 0 0 0 0 7 0 2 0 0 0
0 0 0 0 0 0 2 0 0 0 0 0 0 0 7 7 2 0 0 0
0 0 0 0 0 0 2 0 0 0 0 0 0 0 0 0 2 0 0 0
0 0 0 0 0 0 2 0 0 5 5 5 0 0 0 0 2 0 0 0
0 0 0 0 0 0 2 0 0 0 5 0 0 0 0 0 2 0 0 0
0 0 0 0 0 0 2 0 0 0 0 0 0 0 0 0 2 0 0 0
0 0 0 3 3 0 2 0 0 0 0 0 0 0 0 0 2 0 0 0
0 0 0 3 3 0 2 0 0 0 0 0 0 0 0 0 2 0 0 0
0 0 0 0 0 0 2 0 0 0 0 0 0 0 0 0 2 0 0 0
0 0 0 0 0 0 2 2 2 2 2 2 2 2 2 2 2 0 0 0
0 0 0 0 0 0 0 0 0 0 0 0 0 0 0 0 0 0 0 0
0 0 0 0 0 0 0 0 0 0 0 0 0 0 0 0 0 0 0 0
Output:
0 0 0 0 0 0 0 0 0 0 0 0 0 0 0 0 0 0 0 0
0 0 0 0 0 0 0 0 0 0 0 0 0 0 0 0 0 0 0 0
0 0 0 0 0 0 0 0 0 0 0 0 0 0 0 0 0 0 0 0
0 0 0 0 0 0 0 0 0 0 0 0 0 0 0 0 0 0 0 0
0 0 0 0 0 0 0 0 0 0 0 0 0 0 0 0 0 0 0 0
0 0 0 0 0 0 0 0 0 0 0 0 0 0 0 0 0 0 0 0
0 0 0 0 0 0 0 0 0 0 0 0 0 0 0 0 0 0 0 0
0 0 0 0 0 0 0 7 7 0 0 0 0 0 7 0 0 0 0 0
0 0 0 0 0 0 0 7 7 0 0 0 0 0 7 0 0 0 0 0
0 0 0 0 0 0 0 0 0 0 0 0 0 0 7 7 0 0 0 0
0 0 0 0 0 0 0 0 0 0 0 0 0 0 0 0 0 0 0 0
0 0 0 0 0 0 0 0 0 5 5 5 0 0 0 0 0 0 0 0
0 0 0 0 0 0 0 0 0 0 5 0 0 0 0 0 0 0 0 0
0 0 0 0 0 0 0 0 0 0 0 0 0 0 0 0 0 0 0 0
0 0 0 0 0 0 0 0 0 0 0 0 0 0 0 0 0 0 0 0
0 0 0 0 0 0 0 0 0 0 0 0 0 0 0 0 0 0 0 0
0 0 0 0 0 0 0 0 0 0 0 0 0 0 0 0 0 0 0 0
0 0 0 0 0 0 0 0 0 0 0 0 0 0 0 0 0 0 0 0
0 0 0 0 0 0 0 0 0 0 0 0 0 0 0 0 0 0 0 0
0 0 0 0 0 0 0 0 0 0 0 0 0 0 0 0 0 0 0 0

Example 2:
Input:
0 0 0 0 0 0 0 0 0 0 0 0 0 0 0 0 0 0 0 0
0 0 0 0 0 0 0 0 0 0 0 0 0 0 0 0 0 0 0 0
0 0 0 0 4 4 4 4 4 4 4 4 4 4 4 0 0 0 0 0
0 0 0 0 4 0 0 0 0 0 0 0 0 0 4 0 0 0 0 0
0 0 0 0 4 0 0 0 0 6 6 6 0 0 4 0 0 0 0 0
0 0 0 0 4 0 0 0 0 0 6 0 0 0 4 0 0 0 0 0
0 0 0 0 4 0 0 0 0 0 0 0 0 0 4 0 0 0 0 0
0 0 0 0 4 0 0 0 0 0 0 0 0 0 4 0 0 0 0 0
0 0 0 0 4 5 0 0 0 0 0 0 0 0 4 0 0 0 0 0
0 0 0 0 4 5 0 0 0 5 0 0 0 0 4 0 0 0 0 0
0 0 0 0 4 5 5 0 0 5 0 0 0 0 4 0 0 0 0 0
0 0 0 0 4 0 0 0 0 5 5 0 0 0 4 0 0 0 0 0
0 0 0 0 4 0 0 0 0 0 0 0 0 0 4 0 0 0 0 0
0 0 0 0 4 4 4 4 4 4 4 4 4 4 4 0 0 0 0 0
0 0 0 0 0 0 0 0 0 0 0 0 0 0 0 0 0 0 0 0
0 0 0 0 0 0 0 0 0 0 0 0 0 0 0 0 0 0 0 0
0 0 0 0 0 0 0 0 0 0 0 0 0 0 0 0 0 0 0 0
0 0 0 0 0 0 7 7 0 0 0 0 0 0 0 0 0 0 0 0
0 0 0 0 0 0 7 7 0 0 0 0 0 0 0 0 0 0 0 0
0 0 0 0 0 0 0 0 0 0 0 0 0 0 0 0 0 0 0 0
Output:
0 0 0 0 0 0 0 0 0 0 0 0 0 0 0 0 0 0 0 0
0 0 0 0 0 0 0 0 0 0 0 0 0 0 0 0 0 0 0 0
0 0 0 0 0 0 0 0 0 0 0 0 0 0 0 0 0 0 0 0
0 0 0 0 0 0 0 0 0 0 0 0 0 0 0 0 0 0 0 0
0 0 0 0 0 0 0 0 0 6 6 6 0 0 0 0 0 0 0 0
0 0 0 0 0 0 0 0 0 0 6 0 0 0 0 0 0 0 0 0
0 0 0 0 0 0 0 0 0 0 0 0 0 0 0 0 0 0 0 0
0 0 0 0 0 0 0 0 0 0 0 0 0 0 0 0 0 0 0 0
0 0 0 0 0 5 0 0 0 0 0 0 0 0 0 0 0 0 0 0
0 0 0 0 0 5 0 0 0 5 0 0 0 0 0 0 0 0 0 0
0 0 0 0 0 5 5 0 0 5 0 0 0 0 0 0 0 0 0 0
0 0 0 0 0 0 0 0 0 5 5 0 0 0 0 0 0 0 0 0
0 0 0 0 0 0 0 0 0 0 0 0 0 0 0 0 0 0 0 0
0 0 0 0 0 0 0 0 0 0 0 0 0 0 0 0 0 0 0 0
0 0 0 0 0 0 0 0 0 0 0 0 0 0 0 0 0 0 0 0
0 0 0 0 0 0 0 0 0 0 0 0 0 0 0 0 0 0 0 0
0 0 0 0 0 0 0 0 0 0 0 0 0 0 0 0 0 0 0 0
0 0 0 0 0 0 0 0 0 0 0 0 0 0 0 0 0 0 0 0
0 0 0 0 0 0 0 0 0 0 0 0 0 0 0 0 0 0 0 0
0 0 0 0 0 0 0 0 0 0 0 0 0 0 0 0 0 0 0 0

[Examples 3 through 10 elided -- eight more 20x20 input/output grid pairs from this task, same Input:/Output: format.]

**Memory** (all extracted patterns -- use as reference; apply what is relevant):

[1]
When to use: The task has two same-sized input grids and the output has the same height but double the width, arranged as a left-right concatenation. The left half reproduces the shape pattern from the first input but normalizes every non-background cell to one single fill color, while the right half copies the second input unchanged.

Strategy: (1) Verify that the two inputs have identical dimensions and that the output width should be the sum of their widths. (2) Build the left half from the first input by keeping background cells as background and changing every non-background cell, regardless of its original color, to the common fill color required by the task. This preserves the exact geometry and positions of all shapes from the first input but removes color distinctions. (3) Build the right half by copying the second input exactly without moving or recoloring anything. (4) Concatenate the transformed first grid and the unchanged second grid horizontally, row by row, to form the final output.
[Tool Buffer - Callable Helpers]
These helper functions are callable in the execution environment and should be reused by direct call when applicable.
Callable helper names: apply_border, apply_flip_horizontal, apply_hollow, apply_mark_center, apply_op_per_object, apply_recolor, apply_translate, extract_objects
**DO NOT redefine these helpers.** Do not use globals(), locals(), or dynamic lookup; call helpers by name only.

[Helper bodies elided here -- the 8 inlined function definitions are reproduced in full in the Tool helper definitions subsection that follows. The solver may call them but must not redefine them.]

[History Buffer - Previous Task Trajectories - Reference Only]
These are previous task-specific trajectories. They are for pattern reference only and are NOT callable.

[Task A]
Input:
  5 0 0 0 0 0 0 0 0 0 0 0 0 0 0
  0 8 8 8 8 0 0 0 0 0 0 0 0 0 0
  0 8 8 8 8 0 3 3 3 3 0 0 0 0 0
  0 8 8 8 8 0 3 0 3 0 0 0 0 0 0
  0 8 8 8 8 0 3 0 3 0 0 0 0 0 0
  0 0 0 0 0 0 3 3 3 3 0 0 0 0 0
  0 0 0 0 0 0 0 0 0 0 0 0 0 0 0
  0 0 0 0 0 0 0 0 0 0 0 0 0 0 0
  0 0 0 0 0 0 0 0 0 0 8 0 0 0 0
  0 0 0 0 0 0 0 0 0 0 8 0 0 0 0
  0 0 0 0 0 0 0 0 0 0 8 8 8 8 0
  0 0 0 0 0 0 8 0 0 0 8 0 0 8 0
  0 0 0 0 0 0 8 8 8 0 0 0 0 0 0
  0 0 0 0 0 0 8 0 8 0 0 0 0 0 0
  0 0 0 0 0 0 0 0 0 0 0 0 0 0 0
Output:
  5 0 0 0 0 0 0 0 0 0 0 0 0 0 0
  0 0 0 0 0 0 0 0 0 0 0 0 0 0 0
  0 0 0 0 0 0 0 0 0 0 0 0 0 0 0
  0 0 0 0 0 0 0 0 0 0 0 0 0 0 0
  0 0 0 0 0 0 0 0 0 0 0 0 0 0 0
  0 0 0 0 0 0 0 0 0 0 0 0 0 0 0
  0 0 0 0 0 0 0 0 0 0 0 0 0 0 0
  0 0 0 0 0 0 0 0 0 0 0 0 0 0 0
  0 0 0 0 0 0 0 0 0 0 0 0 0 0 0
  0 0 0 0 0 0 0 0 0 0 0 0 0 0 0
  0 0 0 0 0 0 0 0 0 0 0 0 0 0 0
  0 0 0 0 0 0 0 0 0 0 0 0 0 0 0
  0 0 0 0 0 0 0 0 0 0 0 0 0 0 0
  0 0 0 0 0 0 0 0 0 0 0 0 0 0 0
  0 0 0 0 0 0 0 0 0 0 0 0 0 0 0

Solution:

def solve_task_A(grid):
    h = len(grid)
    w = len(grid[0]) if h else 0

    # Control is the top-left cell.
    marker = grid[0][0] if h and w else 0

    # From the examples:
    # - marker 4 => keep grid unchanged
    # - marker 5 => clear everything except the marker cell
    if marker == 4:
        return [row[:] for row in grid]

    out = [[0] * w for _ in range(h)]
    if h and w:
        out[0][0] = marker
    return out

---

[Task B]
Input:
  0 0 0 0 0 0 0 0 0 0 0 0 0 0 0
  0 0 0 0 0 0 0 0 0 0 0 0 0 0 0
  0 0 0 0 0 0 0 0 0 0 0 0 0 0 0
  0 0 0 2 2 0 0 0 0 0 0 0 0 0 0
  0 0 0 2 2 0 0 0 0 0 0 0 0 0 0
  0 0 0 0 0 0 0 0 0 0 0 0 0 0 0
  0 0 0 0 0 0 0 0 0 0 0 0 0 0 0
  0 6 6 6 6 6 6 6 6 6 6 6 6 6 0
  0 6 0 0 0 0 0 0 9 9 0 0 0 6 0
  0 6 0 0 0 2 0 0 9 9 0 0 0 6 0
  0 6 0 0 2 2 2 0 0 0 9 0 0 6 0
  0 6 0 0 0 2 0 0 0 9 9 9 0 6 0
  0 6 0 0 0 0 0 0 0 0 9 0 0 6 0
  0 6 6 6 6 6 6 6 6 6 6 6 6 6 0
  0 0 0 0 0 0 0 0 0 0 0 0 0 0 0
Output:
  0 0 0 0 0 0 0 0 0 0 0 0 0 0 0
  0 0 0 0 0 0 0 0 0 0 0 0 0 0 0
  0 0 0 0 0 0 0 0 0 0 0 0 0 0 0
  0 0 0 0 0 0 0 0 0 0 0 0 0 0 0
  0 0 0 0 0 0 0 0 0 0 0 0 0 0 0
  0 0 0 0 0 0 0 0 0 0 0 0 0 0 0
  0 0 0 0 0 0 0 0 0 0 0 0 0 0 0
  0 0 0 0 0 0 0 0 0 0 0 0 0 0 0
  0 0 0 0 0 0 0 0 2 2 0 0 0 0 0
  0 0 0 0 0 2 0 0 2 2 0 0 0 0 0
  0 0 0 0 2 2 2 0 0 0 2 0 0 0 0
  0 0 0 0 0 2 0 0 0 2 2 2 0 0 0
  0 0 0 0 0 0 0 0 0 0 2 0 0 0 0
  0 0 0 0 0 0 0 0 0 0 0 0 0 0 0
  0 0 0 0 0 0 0 0 0 0 0 0 0 0 0

Solution:

def solve_task_B(grid):
    h = len(grid)
    w = len(grid[0]) if h else 0

    # Find the largest hollow rectangular frame:
    # a color whose bbox perimeter is fully occupied by that color.
    objs = extract_objects(grid)
    best = None

    for obj in objs:
        cells = set(obj["cells"])
        b = obj["bbox"]
        top, left, bottom, right = b["top"], b["left"], b["bottom"], b["right"]

        # Need a non-degenerate rectangle
        if bottom - top < 2 or right - left < 2:
            continue

        ok = True
        # Check full perimeter
        for c in range(left, right + 1):
            if (top, c) not in cells or (bottom, c) not in cells:
                ok = False
                break
        if ok:
            for r in range(top, bottom + 1):
                if (r, left) not in cells or (r, right) not in cells:
                    ok = False
                    break

        if ok:
            area = (bottom - top + 1) * (right - left + 1)
            if best is None or area > best[0]:
                best = (area, obj)

    out = [[0] * w for _ in range(h)]
    if best is None:
        return out

    frame = best[1]
    b = frame["bbox"]
    top, left, bottom, right = b["top"], b["left"], b["bottom"], b["right"]
    frame_cells = set(frame["cells"])

    # Keep only cells strictly inside the frame, excluding the frame itself,
    # and recolor everything kept to 2.
    for r in range(top + 1, bottom):
        for c in range(left + 1, right):
            if (r, c) not in frame_cells and grid[r][c] != 0:
                out[r][c] = 2

    return out
Reason briefly from the examples, then write the simplest correct `solve`.
Do not output long chain-of-thought or extra prose.
\end{tcblisting}

\paragraph{Response.} The model's reply, extracted as the first Markdown-fenced Python block and graded as passing on all 10 examples:

\begin{tcblisting}{
  outputbox,
  title={Solver response (graded passing)},
  listing only,
  listing options={
    basicstyle=\ttfamily\scriptsize,
    breaklines=true,
    columns=fullflexible,
    keepspaces=true,
    showstringspaces=false
  }
}
def solve(grid):
    h = len(grid)
    w = len(grid[0]) if h else 0

    # Find the largest rectangular hollow frame:
    # a single-color object whose entire bbox perimeter is filled by that color.
    def largest_frame():
        objs = extract_objects(grid)
        best = None  # (area, obj)
        for obj in objs:
            cells = set(obj["cells"])
            b = obj["bbox"]
            top, left, bottom, right = b["top"], b["left"], b["bottom"], b["right"]

            # Need a nontrivial hollow rectangle
            if bottom - top < 2 or right - left < 2:
                continue

            ok = True
            for c in range(left, right + 1):
                if (top, c) not in cells or (bottom, c) not in cells:
                    ok = False
                    break
            if ok:
                for r in range(top, bottom + 1):
                    if (r, left) not in cells or (r, right) not in cells:
                        ok = False
                        break

            if ok:
                area = (bottom - top + 1) * (right - left + 1)
                if best is None or area > best[0]:
                    best = (area, obj)
        return best[1] if best else None

    frame = largest_frame()
    out = [[0] * w for _ in range(h)]
    if frame is None:
        return out

    b = frame["bbox"]
    top, left, bottom, right = b["top"], b["left"], b["bottom"], b["right"]

    # Keep only non-frame cells strictly inside the frame; remove the frame and everything outside.
    for r in range(top + 1, bottom):
        for c in range(left + 1, right):
            if grid[r][c] != 0:
                out[r][c] = grid[r][c]

    return out
\end{tcblisting}

\subsection{Tool helper bodies}
\label{prompt:tool-helpers}

Tool helpers are defined per skill (\cref{app:taxonomy}); they are used
both to generate the ground-truth solver code for each task and to
support the LLM solver at inference, inlined into the Tool Buffer block
of \cref{prompt:solver} and \cref{prompt:selection}. The full set is
reproduced below.

\begin{tcolorbox}[promptbox, title={Tool Buffer (verbatim expansion of the elision in \cref*{prompt:solver} and \cref*{prompt:selection})}]
\begin{lstlisting}[style=solverpy]
[Tool Buffer - Callable Helpers]
These helper functions are callable in the execution environment and should be reused by direct call when applicable.
Callable helper names: apply_border, apply_flip_horizontal, apply_hollow, apply_mark_center, apply_op_per_object, apply_recolor, apply_translate, extract_objects
**DO NOT redefine these helpers.** Do not use globals(), locals(), or dynamic lookup; call helpers by name only.

def apply_border(
    grid: List[List[int]],
    obj: Dict[str, Any],
    border_color: int,
) -> List[List[int]]:
    """Paint ``border_color`` at each background 4-neighbour of ``obj``. Obj cells unchanged."""
    cells = obj.get("cells") or []
    if not grid or not grid[0] or not cells:
        return grid
    h, w = len(grid), len(grid[0])
    obj_set = set(cells)
    bc = int(border_color)
    for r, c in cells:
        for dr, dc in ((-1, 0), (1, 0), (0, -1), (0, 1)):
            nr, nc = r + dr, c + dc
            if 0 <= nr < h and 0 <= nc < w and grid[nr][nc] == 0 and (nr, nc) not in obj_set:
                grid[nr][nc] = bc
    return grid

def apply_flip_horizontal(grid: List[List[int]], obj: Dict[str, Any]) -> List[List[int]]:
    """Mirror ``obj`` left/right around its bbox center column."""
    cells = obj.get("cells") or []
    if not grid or not grid[0] or not cells:
        return grid
    h, w = len(grid), len(grid[0])
    color = int(obj.get("color", 0))
    bbox = obj["bbox"]
    cc = (bbox["left"] + bbox["right"]) / 2.0
    for r, c in cells:
        if 0 <= r < h and 0 <= c < w and grid[r][c] == color:
            grid[r][c] = 0
    for r, c in cells:
        nc = int(cc - (c - cc))
        if 0 <= r < h and 0 <= nc < w:
            grid[r][nc] = color
    return grid

def apply_hollow(
    grid: List[List[int]],
    obj: Dict[str, Any],
    fill_color: int = 0,
) -> List[List[int]]:
    """Keep ``obj`` cells whose 4-neighbour is bg or off-grid (border); set interior cells to ``fill_color``. Foreign-obj neighbours do NOT count as border."""
    cells = obj.get("cells") or []
    if not grid or not grid[0] or not cells:
        return grid
    h, w = len(grid), len(grid[0])
    obj_set = set(cells)
    boundary: set = set()
    for r, c in cells:
        for dr, dc in ((-1, 0), (1, 0), (0, -1), (0, 1)):
            nr, nc = r + dr, c + dc
            if not (0 <= nr < h and 0 <= nc < w):
                boundary.add((r, c))
                break
            if (nr, nc) not in obj_set and grid[nr][nc] == 0:
                boundary.add((r, c))
                break
    fc = int(fill_color)
    for r, c in cells:
        if (r, c) not in boundary:
            grid[r][c] = fc
    return grid

def apply_mark_center(
    grid: List[List[int]],
    obj: Dict[str, Any],
    mark_color: int = 0,
) -> List[List[int]]:
    """Write ``mark_color`` at ``obj``'s integer bbox center cell.
    """
    cells = obj.get("cells") or []
    if not grid or not grid[0] or not cells:
        return grid
    h, w = len(grid), len(grid[0])
    bbox = obj["bbox"]
    cr = (bbox["top"] + bbox["bottom"]) // 2
    cc = (bbox["left"] + bbox["right"]) // 2
    target = int(mark_color)
    if target <= 0:
        color = int(obj.get("color", 0))
        target = (color
        if target == color:
            target = ((color + 1)
    if 0 <= cr < h and 0 <= cc < w:
        grid[cr][cc] = target
    return grid

def apply_op_per_object(
    grid: List[List[int]],
    op_name: str,
    **params: Any,
) -> List[List[int]]:
    """Apply ``op_name`` to every connected component INDEPENDENTLY (on its own isolated patch), then OR-composite the transformed patches onto a fresh blank canvas. Returns the new canvas."""
    per_object_fns = {
        "recolor":         apply_recolor,
        "translate":       apply_translate,
        "flip_horizontal": apply_flip_horizontal,
        "border":          apply_border,
        "hollow":          apply_hollow,
        "mark_center":     apply_mark_center,
        "center_mark":     apply_mark_center,
    }
    if not grid or not grid[0]:
        return [list(row) for row in grid]
    h, w = len(grid), len(grid[0])
    fn = per_object_fns.get(op_name)
    if fn is None:
        return [list(row) for row in grid]
    out = [[0] * w for _ in range(h)]
    for obj in extract_objects(grid):
        if not obj.get("cells"):
            continue
        patch = [[0] * w for _ in range(h)]
        for r, c in obj["cells"]:
            patch[r][c] = obj["color"]
        patch_objs = extract_objects(patch)
        if patch_objs:
            fn(patch, patch_objs[0], **params)
        for r in range(h):
            for c in range(w):
                if patch[r][c]:
                    out[r][c] = patch[r][c]
    return out

def apply_recolor(
    grid: List[List[int]],
    obj: Dict[str, Any],
    new_color: int,
) -> List[List[int]]:
    """Recolor all ``obj`` cells to ``new_color``.
    """
    cells = obj.get("cells") or []
    if not grid or not grid[0] or not cells:
        return grid
    h, w = len(grid), len(grid[0])
    target = int(new_color)
    for r, c in cells:
        if 0 <= r < h and 0 <= c < w:
            grid[r][c] = target
    return grid

def apply_translate(
    grid: List[List[int]],
    obj: Dict[str, Any],
    dr: int,
    dc: int,
) -> List[List[int]]:
    """Shift ``obj`` cells by ``(dr, dc)``; off-grid cells are dropped. Cells that still hold ``obj.color`` at their original position are cleared first so prior per-object writes aren't stomped."""
    cells = obj.get("cells") or []
    if not grid or not grid[0] or not cells:
        return grid
    h, w = len(grid), len(grid[0])
    color = int(obj.get("color", 0))
    for r, c in cells:
        if 0 <= r < h and 0 <= c < w and grid[r][c] == color:
            grid[r][c] = 0
    for r, c in cells:
        nr, nc = r + int(dr), c + int(dc)
        if 0 <= nr < h and 0 <= nc < w:
            grid[nr][nc] = color
    return grid

def extract_objects(grid: List[List[int]], background: int = 0) -> List[Dict[str, Any]]:
    """Return 4-connected non-background components as dicts ``{cells, color, bbox, size}``, in row-major scan order."""
    if not grid or not grid[0]:
        return []
    h, w = len(grid), len(grid[0])
    seen = [[False] * w for _ in range(h)]
    objects: List[Dict[str, Any]] = []
    for sr in range(h):
        for sc in range(w):
            if seen[sr][sc] or grid[sr][sc] == background:
                continue
            color = grid[sr][sc]
            stack: List[Tuple[int, int]] = [(sr, sc)]
            cells: List[Tuple[int, int]] = []
            while stack:
                r, c = stack.pop()
                if not (0 <= r < h and 0 <= c < w) or seen[r][c] or grid[r][c] != color:
                    continue
                seen[r][c] = True
                cells.append((r, c))
                stack.extend([(r - 1, c), (r + 1, c), (r, c - 1), (r, c + 1)])
            if not cells:
                continue
            top = min(r for r, _ in cells); bottom = max(r for r, _ in cells)
            left = min(c for _, c in cells); right = max(c for _, c in cells)
            objects.append({
                "cells": cells,
                "color": int(color),
                "bbox": {"top": top, "left": left, "bottom": bottom, "right": right},
                "size": len(cells),
            })
    return objects
\end{lstlisting}
\end{tcolorbox}

\subsection{Consolidator decision prompt}
\label{prompt:decision}

After the solver's response is graded passing, the consolidator decision prompt is issued. The history buffer at this point contains three entries: the two carryover trajectories shown in the solver's history block above, plus the trajectory just produced for the current task (one IO pair plus the response above).

\paragraph{Input.}
\begin{tcblisting}{
  promptbox,
  title={Consolidator decision prompt (history bodies elided)},
  listing only,
  listing options={
    basicstyle=\ttfamily\scriptsize,
    breaklines=true,
    columns=fullflexible,
    keepspaces=true,
    showstringspaces=false
  }
}

    You manage a History buffer (recent LLM solve traces) and a Strategy memory
    (distilled patterns). Pick ONE action given the state below
    (1 new this step, 2 carryover).

    ## History buffer (3 entries; indices: 1..2 = carryover, 3..3 = new this step; capacity=50, FIFO; 1 IO + solve code per entry)

[History 1 -- carryover -- task_id=A -- passed]
Example IO pair:
  [Same 15x15 IO pair as solver's [Task A] block above.]
Solve code:
```python
[Same as solve_task_A body above.]
```

---

[History 2 -- carryover -- task_id=B -- passed]
Example IO pair:
  [Same 15x15 IO pair as solver's [Task B] block above.]
Solve code:
```python
[Same as solve_task_B body above.]
```

---

[History 3 -- new this step -- task_id=C -- passed]
Example IO pair:
  [Same 20x20 IO pair as solver's Example 1 above.]
Solve code:
```python
[Same as the solver response above.]
```

    ## Strategy memory (1 entries; capacity=unbounded)

    - Strategy 1: When to use: The task has two same-sized input grids and the output has the same height but double the width, arranged as a left-right concatenation. The left half reproduces the shape pattern from the first input but normalizes every non-background cell to one single fill color, while the right half copies the second input unchanged.

Strategy: (1) Verify that the two inputs have identical dimensions and that the output width should be the sum of their widths. (2) Build the left half from the first input by keeping background cells as background and changing every non-background cell, regardless of its original color, to the common fill color required by the task. This preserves the exact geometry and positions of all shapes from the first input but removes color distinctions. (3) Build the right half by copying the second input exactly without moving or recoloring anything. (4) Concatenate the transformed first grid and the unchanged second grid horizontally, row by row, to form the final output.

    Actions:
    - Keep: leave history and memory unchanged.
    - Remove: drop history entries; "fn_indices" required (>=1, 1..H).
    - Strategy extraction: extract a strategy memory entry from selected history; "fn_indices" required (>=1, 1..H; pick >=2 when entries share a plan). Selected entries are CONSUMED from history; their information moves into strategy memory.

    Return JSON:
    {"action": "Keep" | "Remove" | "Strategy extraction", "reason": "<brief>", "fn_indices": [...]}
    (omit "fn_indices" for Keep; required otherwise)
\end{tcblisting}

\paragraph{Response.} The model selects \texttt{Strategy extraction} on history entries~2 and~3, the two \texttt{detect\_inside\_frame\_relation} trajectories that share a plan:

\begin{tcblisting}{
  outputbox,
  title={Decision response},
  listing only,
  listing options={
    basicstyle=\ttfamily\scriptsize,
    breaklines=true,
    columns=fullflexible,
    keepspaces=true,
    showstringspaces=false
  }
}
{"action":"Strategy extraction","reason":"Histories 2 and 3 share the same reusable plan: detect the largest hollow rectangular frame, remove the frame/outside, and keep only nonzero cells strictly inside (optionally recoloring depending on task). This is a strong recurring pattern worth storing.","fn_indices":[2,3]}
\end{tcblisting}

\subsection{Consolidator extraction prompt}
\label{prompt:extraction}

Because the decision was \texttt{Strategy extraction}, the extraction prompt is then issued on the subset of history selected by the decision step. The current strategy buffer is also supplied so the model can choose to retain, merge, or drop existing entries.

\paragraph{Input.}
\begin{tcblisting}{
  promptbox,
  title={Consolidator extraction prompt (task bodies elided)},
  listing only,
  listing options={
    basicstyle=\ttfamily\scriptsize,
    breaklines=true,
    columns=fullflexible,
    keepspaces=true,
    showstringspaces=false
  }
}
You are converting a batch of K solved ARC-AGI tasks into reusable natural-language memory entries.

You will see:
1. The current strategy buffer (1-based indices 1..N). You may RETAIN entries by index, MERGE
   several into a cleaner entry, or DROP entries by omitting them from the output.
2. K input tasks (1-based indices 1..K), each with description, 5 sample IO pairs, and reference
   solution code.

Goal: produce the **full replacement strategy buffer** as a JSON list of entries. Each entry is
exactly one of:

- Retain unchanged:
    {"from_existing": [i, j, ...]}
  Lists >=1 existing indices; each listed index becomes its own kept-as-is entry. NO other fields.
  Only valid when a '### Current strategy buffer' section appears above; do NOT use if buffer is empty.

- New (distilled from task solutions):
    {"when_to_use": "...", "solve_strategy": "...", "from_functions": [k1, k2, ...]}
  Must have BOTH text fields and >=1 "from_functions" index into the K input tasks.

- Merge (existing entries + optionally task evidence into one cleaner entry):
    {"when_to_use": "...", "solve_strategy": "...",
     "from_existing": [i, ...], "from_functions": [k, ...]}
  Must have BOTH text fields; at least one of "from_existing" / "from_functions" non-empty.

Field definitions:

- "when_to_use": describe the visual/task patterns that should trigger this memory in a future
  ARC task. Focus on observable cues: output size, object movement, color changes, cropping,
  symmetry, counting, markers, repeated patterns, etc.

- "solve_strategy": detailed step-by-step reusable solving strategy. Do NOT summarize a single
  task; generalize the solution into an abstract procedure that could be applied to similar
  tasks.

Content rules (apply to every new/merge entry):
- Do not mention task-specific colors as fixed colors. Replace them with semantic roles such as
  background color, target color, marker color, object color, fill color, etc.
- Do not mention task-specific coordinates unless they express a reusable relation
  (top-left, center, border, same row, same column, inside, adjacent, symmetric position).
- Be concrete and operational.
- The strategy should be detailed enough that another model could apply it to a new ARC task.
- Avoid vague statements like "find the pattern" or "transform the object."
- If a pattern is too task-specific, still emit it, but make "when_to_use" narrow.

Schema / structural rules:
- One input task index may appear in multiple "from_functions" lists (one task can illustrate
  several patterns). Multiple input tasks may collapse into one entry (preferred when they
  share an algorithmic plan).
- Existing indices not referenced anywhere in your output are dropped from the new buffer.
- Do NOT re-emit existing entry text -- reference by index instead.
- You may output an empty list to drop everything.
- Output AT MOST N entries. Pick the most reusable, distinct patterns; merge
  near-duplicates rather than listing them separately.

Reply with a JSON list only. Example (mixing all three entry kinds):
[
  {"from_existing": [1, 4]},
  {"when_to_use": "Output size matches input; one object color is the most frequent non-bg ...",
   "solve_strategy": "(1) detect background, (2) ... (3) recolor",
   "from_functions": [2, 3]},
  {"when_to_use": "...",
   "solve_strategy": "merged plan that subsumes prior entry 2 and adds the diagonal axis case",
   "from_existing": [2], "from_functions": [5]}
]

### Current strategy buffer (1-based indices):
1. When to use: The task has two same-sized input grids and the output has the same height but double the width, arranged as a left-right concatenation. The left half reproduces the shape pattern from the first input but normalizes every non-background cell to one single fill color, while the right half copies the second input unchanged.

Strategy: (1) Verify that the two inputs have identical dimensions and that the output width should be the sum of their widths. (2) Build the left half from the first input by keeping background cells as background and changing every non-background cell, regardless of its original color, to the common fill color required by the task. This preserves the exact geometry and positions of all shapes from the first input but removes color distinctions. (3) Build the right half by copying the second input exactly without moving or recoloring anything. (4) Concatenate the transformed first grid and the unchanged second grid horizontally, row by row, to form the final output.

### Input tasks (1-based indices):

#### Task 1:
Example IO pair:
  [Same 15x15 IO pair as solver's [Task B] block above.]
Solve code:
```python
[Same as solve_task_B body above.]
```
Outcome: passed

#### Task 2:
Example IO pair:
  [Same 20x20 IO pair as solver's Example 1 above.]
Solve code:
```python
[Same as the solver response above.]
```
Outcome: passed
\end{tcblisting}

\paragraph{Response.} The model drops the existing \texttt{compose\_horizontal} entry and emits two new entries, both distilled from the inside-frame trajectories:

\begin{tcblisting}{
  outputbox,
  title={Extraction response},
  listing only,
  listing options={
    basicstyle=\ttfamily\scriptsize,
    breaklines=true,
    columns=fullflexible,
    keepspaces=true,
    showstringspaces=false
  }
}
[
  {
    "when_to_use": "The input contains a large hollow rectangular border or frame made of one color, with other smaller objects both inside and outside it, and the output keeps the same grid size but removes the frame and everything outside it. Use this when the key distinction is whether cells lie strictly inside the largest rectangular enclosure.",
    "solve_strategy": "(1) Identify connected color objects and test which ones form a hollow axis-aligned rectangle: their bounding box must be at least 3x3, and every cell on that box's perimeter must belong to the object. (2) If multiple such frames exist, choose the largest by bounding-box area. (3) Create an all-background output grid of the same size. (4) Consider only cells strictly inside the chosen frame, excluding the frame border itself. (5) Copy the interior non-background cells to the same positions in the output, while leaving all cells on the frame and all cells outside the frame as background. This extracts the contents of the frame without moving them.",
    "from_functions": [2]
  },
  {
    "when_to_use": "The input contains a large hollow rectangular frame enclosing one or more objects, and the output keeps only the objects strictly inside that frame but normalizes their colors to a single target color. Use this when inside-vs-outside relation matters and the surviving interior shapes all become one uniform fill color.",
    "solve_strategy": "(1) Detect the largest hollow rectangular frame by finding a connected single-color object whose bounding box perimeter is completely occupied by that object's cells. Require a non-degenerate rectangle so there is a true interior. (2) Build an all-background output grid with the same dimensions as the input. (3) Scan only the cells strictly inside the frame, not including the border. (4) For every non-background cell found in that interior region, write a cell at the same location in the output using one common fill color, regardless of its original color. (5) Leave the frame itself, all exterior objects, and all background cells as background.",
    "from_functions": [1]
  }
]
\end{tcblisting}

\subsection{Failure recording prompt}
\label{prompt:failed_step_record}

In some experimental conditions the history buffer also records steps whose solve was graded failing. Failed entries carry a Python-comment banner that prepends the failed solve code with one input/output triple from the failure: the input on which the candidate failed, the expected output, and what the candidate produced. Because the banner is comment-only, it is valid Python inside the same code fence the solver prompt's history block already uses. Below is one such entry from a real run; the input grids were 16x16, abbreviated to the first four rows for readability.

\begin{tcblisting}{
  promptbox,
  title={Failed-step entry inside the history buffer},
  listing only,
  listing options={
    basicstyle=\ttfamily\scriptsize,
    breaklines=true,
    columns=fullflexible,
    keepspaces=true,
    showstringspaces=false
  }
}
[Task X]

Solution:

# [FAILED] This solution did not pass all evaluation examples.
# Wrong-IO sample (input / expected / got_or_error):
#   [1] input:
#           0 0 0 0 0 0 0 0 0 0 0 0 0 0 0 0
#           0 0 0 0 0 0 0 0 0 0 0 0 6 0 0 0
#           0 0 0 0 0 0 0 0 0 0 0 0 6 0 0 0
#           0 0 0 0 0 0 0 0 0 0 6 6 6 6 6 0
#           [...12 more rows elided...]
#       expected:
#           0 0 0 0 0 0 0 0 0 0 0 0 8 0 0 0
#           0 0 0 0 0 0 0 0 0 0 0 8 6 8 0 0
#           0 0 0 0 0 0 0 0 0 0 8 8 6 8 8 0
#           0 0 0 0 0 0 0 0 0 8 6 6 6 6 6 8
#           [...12 more rows elided...]
#       got:
#           0 0 0 0 0 0 0 0 0 0 0 8 8 8 0 0
#           0 0 0 0 0 0 0 0 0 0 0 8 6 8 0 0
#           0 0 0 0 0 0 0 0 0 8 8 8 6 8 8 8
#           0 0 0 0 0 0 0 0 0 8 6 6 6 6 6 8
#           [...12 more rows elided...]
# ---
def solve_task_X(grid):
    h, w = len(grid), len(grid[0])
    out = [row[:] for row in grid]

    for r in range(h):
        for c in range(w):
            if grid[r][c] != 0:
                continue
            for dr in (-1, 0, 1):
                for dc in (-1, 0, 1):
                    if dr == 0 and dc == 0:
                        continue
                    nr, nc = r + dr, c + dc
                    if 0 <= nr < h and 0 <= nc < w and grid[nr][nc] != 0:
                        out[r][c] = 8
    return out
\end{tcblisting}

\subsection{Alternative schema for the consolidator extraction prompt}
\label{prompt:extraction_alt_schema}

An alternative schema for the consolidator extraction prompt asks the model to emit a flat JSON list of strategies, where each entry is one of three kinds: a retain (\texttt{from\_existing}), a new pattern (\texttt{strategy} text plus \texttt{from\_functions}), or a merge that combines existing entries with optional new evidence. Strategy text is a single free-form string rather than the structured (\texttt{when\_to\_use}, \texttt{solve\_strategy}) pair used in the prompt above. We reproduce one full call from a real run; the buffer of existing strategies was empty at this step, so the model returns one new entry distilled from the single solve function shown.

\paragraph{Input.}
\begin{tcblisting}{
  promptbox,
  title={Alternative extraction prompt},
  listing only,
  listing options={
    basicstyle=\ttfamily\scriptsize,
    breaklines=true,
    columns=fullflexible,
    keepspaces=true,
    showstringspaces=false
  }
}
Identify the high-level algorithmic strategies used in these functions (numbered 1 to 1).
Output the **FULL replacement strategy buffer** as a JSON list. Each output entry is one of:
- **Retain** existing strategies unchanged: ``{"from_existing": [i, j, ...]}`` --
  list any number of 1-based existing indices (the buffer is currently empty). Each listed
  index becomes its own kept-as-is entry in the new buffer (compact form for
  keeping several entries; equivalent to writing one ``{"from_existing": [i]}``
  per index). Omit "strategy" entirely.
- **New** pattern extracted from the functions: ``{"strategy": "<text>", "from_functions": [k1, k2, ...]}``
  where ``k*`` are 1-based indices into the functions below.
- **Merge** existing entries (and optionally new function evidence) into a single
  cleaner description: ``{"strategy": "<merged text>", "from_existing": [i, j, ...], "from_functions": [k, ...]}``.
  Both index lists may have multiple entries; "from_functions" may be omitted if the merge is purely existing-only.

Hard rules:
- Retain entries have NO "strategy" field and reference >=1 existing index. Multiple
  indices in one retain entry mean "keep all of them as separate entries".
- New / merge entries (with "strategy") MUST cite at least one source via "from_existing" or "from_functions".
- The difference between retain (multi-index) and merge: retain produces N separate
  entries unchanged; merge produces ONE new entry whose text is your "strategy" field.
- Existing indices not referenced anywhere in your output are dropped.
- Do NOT copy existing strategy text verbatim -- use the index instead.
- You may output an empty list to drop everything.
Functions are numbered 1..1. Reference them in "from_functions" when extracting a new pattern. Existing strategies are referenced in "from_existing" by their 1-based index in the current strategy buffer.

### Function 1:
```python
def solve(grid):
    objects = extract_objects(grid)
    output = [list(row) for row in grid]
    if not objects:
        return output
    _max_size = max(int(o.get('size', len(o.get('cells') or []))) for o in objects)
    for obj in objects:
        _obj_size = int(obj.get('size', len(obj.get('cells') or [])))
        if _obj_size == _max_size:
            pass
        else:
            _remove_color = int(obj.get('color', 0))
            for _rr, _cc in (obj.get('cells') or []):
                if 0 <= _rr < len(output) and 0 <= _cc < len(output[0]) and output[_rr][_cc] == _remove_color:
                    output[_rr][_cc] = 0
    return output
```

Reply with a JSON list only. Examples:
[
  {"from_existing": [1, 4, 7]},
  {"strategy": "decompose grid -> process each part -> concatenate", "from_functions": [2, 4]},
  {"strategy": "conditional branch on symmetry: existing rule extended to handle the new diagonal axis seen in functions 3 and 5",
   "from_existing": [2, 3], "from_functions": [3, 5]}
]
\end{tcblisting}

\paragraph{Response.}
\begin{tcblisting}{
  outputbox,
  title={Extraction response},
  listing only,
  listing options={
    basicstyle=\ttfamily\scriptsize,
    breaklines=true,
    columns=fullflexible,
    keepspaces=true,
    showstringspaces=false
  }
}
[
  {
    "strategy": "Extract connected objects, identify the maximum-size object class (keeping all ties), and erase every smaller object by setting its cells to background while leaving the largest object(s) unchanged.",
    "from_functions": [1]
  }
]
\end{tcblisting}

\subsection{Strategy selection prompt}
\label{prompt:selection}

An alternative invocation pattern for the solver decomposes the call into a strategy-selection step followed by a synthesis step. The selection prompt provides three inputs: the current callable helper buffer, half of the task's input-output examples, and the current strategy memory. The model replies with a JSON object that names which existing entry to apply; in the unconstrained variant it may instead propose a new one. The selected strategy text is then injected into the synthesis prompt's memory block in place of the full memory listing. We reproduce one full call from a real run; the helper-function bodies and four of the five input-output example pairs are elided.

\paragraph{Input.}
\begin{tcblisting}{
  promptbox,
  title={Strategy selection prompt (helper bodies and all five IO examples elided)},
  listing only,
  listing options={
    basicstyle=\ttfamily\scriptsize,
    breaklines=true,
    columns=fullflexible,
    keepspaces=true,
    showstringspaces=false
  }
}
Choose a strategy for this task. Below are, in order: (1) the **tool buffer**, (2) **this task's input-output examples** (half of full set), (3) **currently existing strategies**. Then choose how to proceed.

    --- 1. Tool buffer ---
    Tool helper mode: direct_call. Helper memory is callable in strategy selection and should be reused by direct call when applicable.

    Tool buffer (callable helper functions currently available):
    extract_objects, apply_recolor, apply_translate, apply_flip_horizontal,
    apply_border, apply_hollow, apply_mark_center, apply_op_per_object
    [helper signatures and bodies elided here -- reproduced in full in the
    Tool helper definitions subsection.]

    --- 2. This task's input-output examples ---
    [Five 20x20 Input:/Output: example pairs elided -- same format as in the solver prompt above.]

    --- 3. Currently existing strategies ---
      0. Extract connected objects, choose the largest as a frame, classify other objects by whether their bounding boxes lie strictly inside that frame, erase the frame and all outside objects, then hollow out each inside object in place by turning its interior to color 0 while preserving its boundary.

    --- 4. Choose strategy ---
    Options (include a short "reason" in your reply). You MUST pick one existing strategy -- no other action is accepted:
    B) **Use an existing strategy**: {"action": "select", "index": N, "reason": "brief reason"} where N is 0 to 0

    [Guidance bullets elided.]

    Reply with only the JSON, no other text.
\end{tcblisting}

\paragraph{Response.}
\begin{tcblisting}{
  outputbox,
  title={Selection response},
  listing only,
  listing options={
    basicstyle=\ttfamily\scriptsize,
    breaklines=true,
    columns=fullflexible,
    keepspaces=true,
    showstringspaces=false
  }
}
{"action":"select","index":0,"reason":"Examples show keeping only the largest frame-like object, deleting all other objects, and hollowing/removing enclosed non-frame content; this matches the existing inside-frame filtering and hollowing strategy best."}
\end{tcblisting}

\section{Strategy memory evolution over time [ARC-AGI]}
\label{app:evolution}

\subsection{How a strategy is traced across compress steps}
\label{subsec:evo-tracking}

At each compress step we run the extraction prompt (\cref{prompt:extraction}); the consolidator returns the rewritten strategy memory as a JSON list. Each list item declares its \texttt{kind}, one of three values:

\begin{itemize}[leftmargin=*,topsep=2pt,itemsep=2pt]
  \item \textbf{\texttt{retain}} --- keep an existing entry from the current memory; the field \texttt{from\_existing} lists 1-based indices into that memory.
  \item \textbf{\texttt{new}} --- distill a fresh entry from selected input tasks; the field \texttt{from\_functions} lists 1-based indices into this step's input tasks.
  \item \textbf{\texttt{merge}} --- emit a new entry that combines existing entries (\texttt{from\_existing}) with new input tasks (\texttt{from\_functions}).
\end{itemize}

Both \texttt{kind} and the two index fields come directly from the consolidator's JSON output. After parsing the response, we persist the resulting memory as a per-step JSON snapshot; each entry in the snapshot carries this step's \texttt{kind} and its associated indices.

Because both \texttt{merge} and \texttt{new} produce a fresh entry rather than editing an existing one in place, tracking how a strategy evolves over multiple steps requires walking \texttt{from\_existing} pointers backwards across snapshots. Starting from a target entry, we look at its \texttt{kind}: if \texttt{kind = new}, the chain has reached its root and the walk terminates; if \texttt{kind = merge}, the index \texttt{from\_existing: [j]} points to a predecessor in the previous compress step's memory; we look up the predecessor in the snapshot we persisted for that step and take it as the new target. We call a chain reconstructed by this procedure a \emph{strategy lineage}.

\subsection{Worked lineage example}
\label{subsec:evo-example}

We ran GPT-5.4 on a 200-task pool spanning six ARC families, with the same model serving as both solver and consolidator. We trace one strategy lineage by the procedure of \cref{subsec:evo-tracking}, from its \texttt{new} step through its final merge, and show three snapshots: the chain root, an intermediate node, and the chain's final node. In the latter two, yellow shading marks spans that differ from the previous snapshot.

\paragraph{Chain root.} The chain begins with a \texttt{new} step: the consolidator distills four inside-frame tasks into one entry.

\begin{tcolorbox}[memorybox, title={Chain root}]
\textbf{When to use:} A large hollow rectangular frame or border encloses some smaller objects, while other objects may lie outside the frame. The output removes the frame itself and removes all outside objects, preserving only the objects strictly inside the frame. In some variants the kept inside objects are additionally mirrored left-right within their own bounding boxes.

\smallskip
\textbf{Strategy:} (1) Extract all connected non-background objects. (2) Identify the enclosing frame as the largest object, typically a hollow rectangle. (3) For every other object, compare its bounding box with the frame's bounding box; mark it as inside only if its bounding box lies strictly within the frame on all four sides. (4) Start from a copy of the input grid. (5) Erase the frame object from the grid. (6) Erase every object not strictly inside the frame, including all objects outside the frame. (7) For each surviving inside object, determine from training whether it should remain unchanged or be horizontally flipped within its own bounding box. If flipping is required, remap each occupied cell at row offset r and column offset c to row offset r and column offset width-1-c inside that object's box, using the object's own color. (8) Return the full grid with only the interior objects retained.
\end{tcolorbox}

\paragraph{Intermediate node.} Compared with the chain root: the horizontal-flip coordinate formula at root step~(7) is gone, the \emph{When to use} clause about mirroring is gone, and a uniform-recolor step has been added in its place. The overall scaffold --- identify the frame as the largest object, classify other objects by strict bounding-box containment, erase the frame and outside objects, keep the inside objects --- is preserved nearly verbatim.

\begin{tcolorbox}[memorybox, title={Intermediate node}]
\textbf{When to use:} A grid contains one large enclosing rectangular frame and several smaller objects, some strictly inside it and others outside or touching it. The output removes the frame and discards everything not strictly enclosed, leaving only the interior objects; sometimes those kept objects are recolored to a single common color, and sometimes they are simply kept unchanged.

\smallskip
\textbf{Strategy:} (1) Extract all connected non-background objects. (2) Identify the enclosing frame, typically the largest hollow rectangular object. (3) For every other object, compare bounding boxes with the frame's bounding box. Classify an object as inside only if its top, left, bottom, and right bounds are all strictly within the frame bounds; objects touching the frame or lying outside are not inside. (4) Create an output grid, usually starting from the input. (5) Erase the frame itself. (6) Erase every object not strictly inside the frame. (7) Keep the inside objects in their original positions. (8) \hl{If training shows a post-processing step on the kept objects, apply it uniformly, such as recoloring all remaining cells to a single target color}; otherwise leave their internal colors unchanged. (9) Return the resulting grid.
\end{tcolorbox}

\paragraph{Chain end.} Compared with the previous snapshot: the recolor step is gone, replaced by a uniform-translation step at the same position, and the \emph{When to use} now mentions \emph{all shifted by one common offset} in line with that change. The mirror clause that was already missing in the previous snapshot is still absent.

\begin{tcolorbox}[memorybox, title={Chain end}]
\textbf{When to use:} A grid contains one large hollow rectangular frame plus several other objects, and the output removes the frame and all objects outside it, keeping only objects strictly inside the frame. The surviving inside objects may also all be shifted by one common offset.

\smallskip
\textbf{Strategy:} (1) Extract all connected non-background objects. (2) Identify the frame as the large rectangular border object, typically the largest object. (3) For every other object, decide whether it is strictly inside by checking that its bounding box lies strictly within the frame's bounding box on all four sides. (4) Start from a full-size output grid and erase the frame. (5) Erase every object classified as outside the frame. (6) Keep only the inside objects. (7) \hl{If training shows a uniform translation, move every inside object by the same learned row/column offset, clearing original cells before redrawing them}; otherwise leave them in place. (8) Return the full-size grid containing only the surviving processed inside objects.
\end{tcolorbox}

\section{Examples of wrong / low-quality strategies [ARC-AGI]}
\label{app:wrong-examples}

Each subsection below names one failure mode and shows a verbatim
entry from the strategy memory, paired with either a callout that
explains highlighted spans or a held-out solver code comparison.

\subsection{Vacuous abstraction}
\label{subsec:wrong-vacuous}

We ran GPT-5-mini on a 200-task pool spanning six families.
After each task, the model could update its strategy memory or
leave it unchanged; we set no cap on entry count. After task~200,
the memory's first entry read:

\begin{tcolorbox}[memorybox, title={Strategy memory entry}]
\annot{1}{Make a working copy of the input grid (list of row lists)
before mutating, perform all modifications on the copy, and return
the copy to avoid mutating the original input.}
\end{tcolorbox}

\begin{annotations}
\annotitem{1}{Vacuous abstraction}{A defensive Python idiom shared
by every nested-list program. Mentions no color, shape, or rule
that the six families distinguish themselves by.}
\end{annotations}

\subsection{Concrete-vocabulary erosion}
\label{subsec:wrong-strip}

We showed GPT-5-mini one training task (``recolor the largest
object'') for 50 consecutive rewrites of its memory. Round~1
still mentions ``max size'' --- the only concrete property the
task actually selects on --- as a parenthetical example of the
metric to compute. By round~50, the same lineage has no word left
for it:

\begin{tcolorbox}[memorybox, title={Strategy memory entry --- round 1}]
identify and extract structured elements from input
\textrightarrow{} compute a global metric (e.g., max size)
\textrightarrow{} iterate elements and selectively apply targeted
edits
\end{tcolorbox}

\begin{tcolorbox}[memorybox, title={Strategy memory entry --- round 50}]
Find the \annot{1}{maximum value of a derived per-object numeric
attribute} and apply a uniform transformation to every object
whose attribute equals that maximum.
\end{tcolorbox}

\begin{annotations}
\annotitem{1}{Concrete feature stripped}{Round~1's ``max size''
names a property a solver can compute in cells. The round~50
phrasing names no specific feature; the entry no longer records
\emph{which} attribute to maximize.}
\end{annotations}

\subsection{Foreign-family injection}
\label{subsec:wrong-mix}

We presented training tasks to GPT-5.4 one at a time, each paired
with its ground-truth solver code. We required the model to
compress every task into a single memory entry on receipt. The compressor distilled the task
into an entry whose Strategy steps mix in content from multiple
families.

\begin{tcolorbox}[memorybox, title={Strategy memory entry}]
\textbf{When to use:} A large hollow rectangular frame encloses
some objects while other objects lie outside it, and the output
keeps only the interior objects in their original positions while
deleting the frame and all exterior objects. \annot{1}{In the kept
interior objects, a single distinguished cell is changed based on a
relation to a matching object outside the frame, often when an
outside object has the same shape as an inside object.}

\smallskip
\textbf{Strategy:} (1) Detect the background color and extract
connected non-background objects.
(2) Identify the frame as the large hollow rectangle.
(3) Classify every other object as inside or outside the frame by
strict bounding-box containment.
(4) Erase the frame and every object outside the frame.
(5) For each interior object, \annot{2a}{look for an exterior object
with the same shape signature, allowing translation but not
reshaping; compare normalized cell sets or bounding-box-relative
coordinates.}
(6) \annot{2b}{If an interior object has such a matching exterior
counterpart, mark the center cell of the interior object's bounding
box with the exterior object's color.} If no such counterpart
exists, leave the interior object unchanged.
(7) Return the full-size grid with everything else set to background.
\end{tcolorbox}

\begin{annotations}
\annotitem{1}{Foreign-family extension to the trigger}{The second
sentence of the trigger introduces a shape-relation step that the
inside-frame source task does not perform; it pre-commits the
solver to look for a cross-frame shape match before any retrieval
filter can reject the entry.}
\annotitem{2a--2b}{Composite absent from every family}{Spans (2a) and
(2b) together form a single composite: a shape-signature lookup
across the frame (2a), followed by a marker-style color write into
the matched interior object's center (2b). The lookup belongs to
the group-by-shape family and the color write to the
key-marker family; neither is part of the inside-frame
source task, and no family in the pool prescribes this
composite.}
\end{annotations}

\subsection{Phantom strategies from failed attempts}
\label{subsec:wrong-phantom}

In a 200-task GPT-5.4 run, we exposed the strategy-update prompt
to both successful and failed past attempts (each failure paired
with the wrong output it produced). We expected this to let the
model prune bad entries by surfacing what had not worked. The output
instead distilled recurring patterns from the failed code into new
entries. The strategy library at
the end of the run held five entries; two were paraphrases of the
same fictional rule. We reproduce the longer paraphrase below:

\begin{tcolorbox}[memorybox, title={Strategy memory entry}]
\textbf{When to use:} A same-size output contains \annot{1}{glyph-like
objects made of horizontal and vertical strokes}, and only one side
stroke or partial row/column segment is shifted within the object's
own bounding box to \annot{1}{normalize the glyph}. The object is
not translated as a whole and not hollowed. Often one full-width or
full-height stroke stays fixed while shorter spans attached to one
side move inward or to the opposite side.

\smallskip
\textbf{Strategy:} (1) Segment connected non-background objects and
treat isolated marker cells separately.
(2) For each candidate glyph, compute its bounding box plus row
spans and column spans.
(3) Identify \annot{2}{anchor strokes that remain fixed, such as a
full-width top or bottom bar or another span matching the final
examples.}
(4) Detect \annot{2}{shorter side-attached spans that differ from
the anchor-aligned placement by a constant horizontal or vertical
offset.}
(5) Infer from training which subset of rows or columns should move
and in which direction; do not assume all non-anchor spans move.
Often the moved spans are the interior partial strokes while an
outer full bar stays unchanged.
(6) \annot{3}{Clear the original glyph and redraw it inside the
same bounding box using the unchanged anchor spans plus the
shifted partial spans.}
(7) Apply the same \annot{1}{local stroke-shift rule} to each
target glyph class, preserving color and overall bounding-box
location, and leave unrelated objects unchanged.
\end{tcolorbox}

\begin{annotations}
\annotitem{1}{Fictional family and operation traced to failure code}{Neither
``glyph-like'' (the family) nor ``normalize the glyph'' /
``stroke-shift rule'' (the operation) matches anything in the six
families the run contained; the consolidator produced both. The
lineage traces to a recurring fragment in failed solver code:
\texttt{out[r][min(cols)]\,=\,0} (``erase the leftmost cell of this
row''). The strategy-update step read this fragment as evidence of
an intentional row-level edit rule; the font-rendering metaphor (a
glyph being normalized by a stroke shift) is the abstraction it
produced.}
\annotitem{2}{Structural vocabulary borrowed from neighboring
families}{``Anchor stroke'' and ``side-attached span'' resemble
the geometric vocabulary used by the inside-frame and
compose-horizontal families (frame edges, panel borders), but
the entry applies these terms to a different referent ---
internal parts of a single object --- where no real family does
so.}
\annotitem{3}{Operational scaffolding for a non-existent rule}{Step
(6) gives the solver concrete, copy-paste-ready instructions
(``clear the original glyph'', ``redraw inside the same bounding
box''), but the instructions are wrong.}
\end{annotations}

\subsection{Single-strategy collapse on multi-family compression}
\label{subsec:wrong-overfocus}

The cases above describe entries whose content is itself
problematic. This subsection describes a different failure with
two parts: (i) compression collapses multi-family training tasks
into one single-family playbook; (ii) retrieval applies that entry
verbatim to held-out tasks from other families. The entry text
itself is internally consistent and appropriate for its source
family.

\paragraph{Example 1: 19 tasks revisited 10 times collapse to a
single inside-frame plus hollow strategy.}
We presented a fixed pool of 19 tasks, spanning the six families,
to GPT-5.4 10 times. The model viewed one task at a time alongside
its ground-truth solver code, and updated its memory after each
task. After 10 passes (190 steps), the memory contained a single
entry:

\begin{tcolorbox}[memorybox, title={Strategy memory entry}]
Extract connected objects, choose the largest as a frame, classify
other objects by whether their bounding boxes lie strictly inside
that frame, erase the frame and all outside objects, then hollow
out each inside object in place by turning its interior to
color~$0$ while preserving its boundary.
\end{tcolorbox}

\noindent
On a held-out task whose true rule was ``erase every object of one
specific color'' --- unrelated to the inside-frame pattern in the
entry --- the same model produced:

\begin{tcolorbox}[outputbox, title={Held-out solver code with memory (color-removal task)}]
\begin{lstlisting}[style=solverpy]
def solve(grid):
    # Keep only non-frame objects; frames are the largest
    # connected components (color 6 in the examples).
    # Then hollow the kept objects.
    h = len(grid); w = len(grid[0]) if h else 0
    objs = extract_objects(grid)
    if not objs:
        return [row[:] for row in grid]

    frame = max(objs, key=lambda o: o["size"])
    fb = frame["bbox"]

    out = [[0] * w for _ in range(h)]
    for obj in objs:
        if obj is frame: continue
        bb = obj["bbox"]
        inside = (bb["top"]  > fb["top"]
              and bb["bottom"] < fb["bottom"]
              and bb["left"] > fb["left"]
              and bb["right"] < fb["right"])
        if inside:
            for r, c in obj["cells"]:
                out[r][c] = obj["color"]

    for obj in extract_objects(out):
        apply_hollow(out, obj, fill_color=0)
    return out
\end{lstlisting}
\end{tcolorbox}

\noindent
The first comment reproduces the entry's framing verbatim. The
second comment records the active color from the examples. The
code below treats color~$6$ as the frame color rather than as the
color to remove. The code follows the entry's
frame-and-inside structure: it picks a frame by size, copies only
strictly-interior objects onto a blank canvas, and hollows them in
place. On this input no color-$3$ marker passes the strict-inside
check, so the canvas stays empty. The same model without strategy
memory solved the task in eight lines:

\begin{tcolorbox}[outputbox, title={Same model on the same task, no strategy memory}]
\begin{lstlisting}[style=solverpy]
def solve(grid):
    # The transformation removes every connected component of color 6
    # and leaves all other cells unchanged.
    out = [row[:] for row in grid]
    for obj in extract_objects(grid):
        if obj["color"] == 6:
            for r, c in obj["cells"]:
                out[r][c] = 0
    return out
\end{lstlisting}
\end{tcolorbox}

\begin{figure}[h]
\centering
\includegraphics[width=\linewidth]{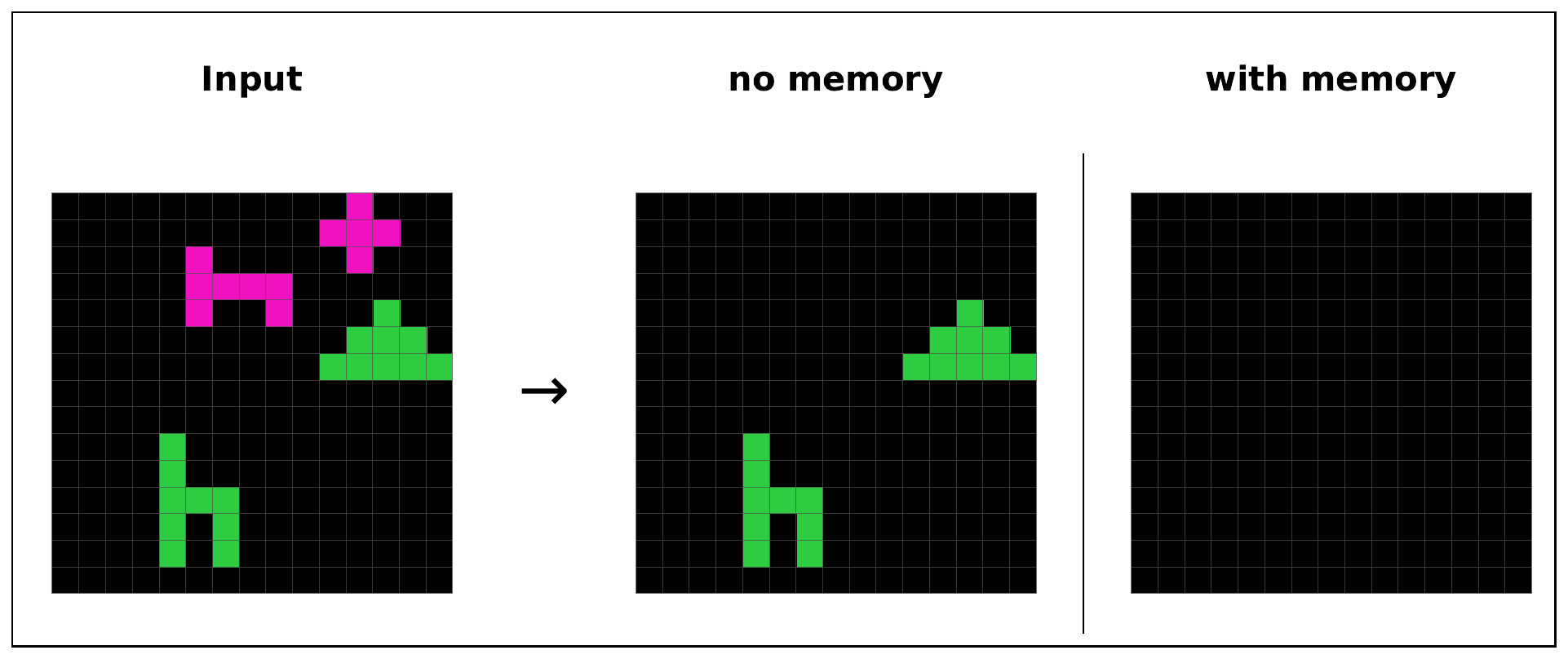}
\caption{Example 1, held-out task. The input contains a hollow
color-$6$ frame and several smaller color-$3$ markers. The same GPT-5.4
solver passes this task without strategy memory (center, 10/10) by
erasing every color-$6$ component and keeping the markers; with
memory (right, 0/10) the solver instead executes the entry's
frame-and-hollow playbook and produces an empty grid.}
\label{fig:wrong-overfocus-task224}
\end{figure}

\paragraph{Example 2: 250 tasks streamed once collapse to a
group-by-shape strategy.}
In a separate run, we presented 250 training tasks to GPT-5.4
one at a time, each paired with its ground-truth solver code.
After each task the model chose whether to update the memory. Training tasks again
spanned the six families. After 250 steps the memory contained
two entries, both from the group-by-shape family. Each
entry prescribed partitioning objects by translation-normalized
shape signature, keeping the most frequent signature class, and
applying a decoration to the survivors. The other five families
(largest-object, color-property, key-marker, inside-frame,
compose-horizontal) had no entry in the memory.

\begin{tcolorbox}[memorybox, title={Strategy memory entry}]
\textbf{When to use:} The grid contains several disconnected
non-background objects. The output keeps only the objects whose
translation-normalized shape occurs most often, erases the rest,
and applies a single decoration to each survivor.

\smallskip
\textbf{Strategy:} (1) Extract all connected non-background
objects.
(2) For each object, compute a position-independent shape
signature by translating its cells so its topmost-leftmost cell
becomes the origin, then sorting the relative coordinates.
(3) Count how many objects share each signature; pick the
majority signature, breaking ties by first appearance.
(4) Starting from a copy of the input, erase every object whose
signature is not the majority signature.
(5) Keep all objects with the majority signature unchanged in
place and color.
(6) Apply the task's required common decoration to each survivor;
in one common variant, draw a one-cell outline around the object
using a marker color without overwriting the object itself.
\end{tcolorbox}

\noindent
On a held-out task from the largest-object family --- true rule
``keep only the largest connected component, erase the rest'' ---
the same model with this memory produced:

\begin{tcolorbox}[outputbox, title={Held-out solver code with memory (keep-largest task)}]
\begin{lstlisting}[style=solverpy]
def solve(grid):
    objs = extract_objects(grid)
    if not objs:
        return [row[:] for row in grid]

    def signature(obj):
        cells = obj["cells"]
        top = min(r for r, c in cells)
        left = min(c for r, c in cells)
        return tuple(sorted((r - top, c - left) for r, c in cells))

    # Count normalized shapes, regardless of color/position.
    counts = {}
    first_seen = {}
    for i, obj in enumerate(objs):
        sig = signature(obj)
        counts[sig] = counts.get(sig, 0) + 1
        if sig not in first_seen:
            first_seen[sig] = i

    # Keep the most frequent shape; break ties by first appearance.
    best_sig = min(
        counts,
        key=lambda s: (-counts[s], first_seen[s])
    )

    h, w = len(grid), len(grid[0])
    out = [[0] * w for _ in range(h)]

    for obj in objs:
        if signature(obj) == best_sig:
            color = obj["color"]
            for r, c in obj["cells"]:
                out[r][c] = color

    return out
\end{lstlisting}
\end{tcolorbox}

\noindent
The solver transcribes the entry's selection logic into Python:
\texttt{signature()} implements step~(2), the dictionary count
implements step~(3), and \texttt{best\_sig} implements step~(4).
The downstream operation --- writing each kept object's cells onto
a blank canvas --- is identical to what the same model produces
without memory, shown next:

\begin{tcolorbox}[outputbox, title={Same model on the same task, no strategy memory}]
\begin{lstlisting}[style=solverpy]
def solve(grid):
    objs = extract_objects(grid)
    if not objs:
        return [row[:] for row in grid]
    # Keep only the largest object(s); in all examples the output is exactly
    # the maximal connected component and everything else is removed.
    max_size = max(obj["size"] for obj in objs)
    out = [[0 for _ in row] for row in grid]
    for obj in objs:
        if obj["size"] == max_size:
            color = obj["color"]
            for r, c in obj["cells"]:
                out[r][c] = color
    return out
\end{lstlisting}
\end{tcolorbox}

\noindent
Without memory the model identifies the correct selection
criterion (\texttt{obj["size"] == max\_size}). With memory the
selection criterion is replaced by the entry's
\texttt{signature(obj) == best\_sig}. The action that follows the
selection is unchanged. Replacing the selection criterion alone,
while keeping the action identical, drops the solver from 10/10 to
2/10 on this task's 10 test grids.

\begin{figure}[h]
\centering
\includegraphics[width=\linewidth]{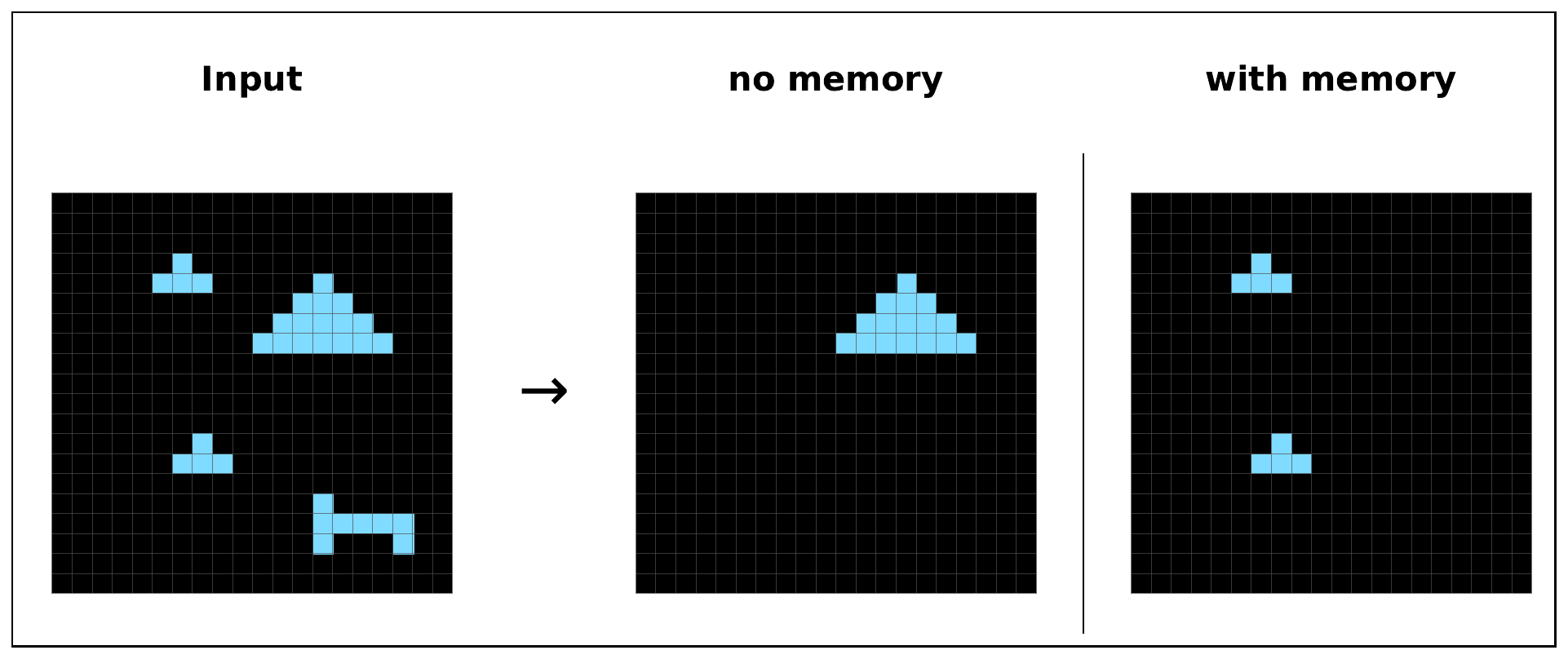}
\caption{Example 2, held-out task. The input contains one large
unique triangle (top-right), one medium unique H-shape
(bottom-right), and two small same-shape T-shapes (top-left and
mid-left). The same GPT-5.4 solver passes this task without
memory (center, 10/10) by keeping only the largest object; with
memory (right, 2/10) the solver keeps only the two same-shape
T-shapes, since the entry's selection criterion is shape-signature
majority rather than size.}
\label{fig:wrong-overfocus-task207}
\end{figure}

\section{Examples of useful / well-formed strategies [ARC-AGI]}
\label{app:useful-examples}

Below are four memory entries that share one property: their
procedural body is concrete and directly operational --- each step
names an action the solver can translate into a Python helper, and
the directions, boundaries, and object selections it depends on
are stated specifically rather than left implicit. Yellow
highlights mark the phrases that most directly drive this
property.

\begin{tcolorbox}[memorybox, title={Strategy memory entry --- compose-horizontal (operation left as a slot)}]
\textbf{When to use:} \hl{The task has two separate input grids of the
same height, and the output width is exactly the sum of their widths,
indicating horizontal concatenation. The left half of the output is
just the first input, while the right half is the second input after
a uniform transformation applied only there}: either every
non-background cell is recolored to one fixed color, or each
connected component keeps its shape but receives a center marker at
its bounding-box midpoint.

\smallskip
\textbf{Strategy:} (1) Verify that the output is formed by placing the
first input on the left and the second input on the right with no
vertical offset.
(2) Build the output by horizontally concatenating corresponding rows
of the two inputs.
(3) Determine what extra rule applies only to the right half by
comparing examples.
(4) If the right-half rule is uniform recoloring, \hl{replace every
non-background cell from the second input by the designated output
color while preserving all positions and shapes.}
This can be done cellwise or per connected component.
(5) If the right-half rule is center marking, extract each
non-background 4-connected component in the second input,
\hl{compute its bounding box, and find the bbox center using the
midpoint row and midpoint column.}
Shift that center into the concatenated output's right half and write
the marker color there, leaving the rest of the component unchanged.
(6) Keep the left half identical to the first input throughout.
\end{tcolorbox}

\begin{tcolorbox}[memorybox, title={Strategy memory entry --- largest-object}]
\textbf{When to use:} \hl{The grid contains several disconnected
objects on the same background, and the output changes only the
object or objects with the greatest area/cell count while leaving all
smaller objects untouched}. The transformation on the selected
objects may be marking a special point, recoloring the whole object,
or mirroring the object within its own bounding box.

\smallskip
\textbf{Strategy:} (1) Extract all connected non-background objects.
(2) For each object, \hl{compute its size as the number of occupied
cells.}
(3) Find the maximum size across all objects; if several objects tie
for largest, select all of them.
(4) Copy the input grid to the output.
(5) \hl{Apply the task-specific operation only to the selected largest
objects, preserving their positions}: for example, recolor every
cell of each selected object, mark the object's center cell if the
examples show a single inserted marker, or horizontally mirror the
object's occupied cells within its own bounding box.
(6) Leave every non-selected object and all background cells unchanged.
\end{tcolorbox}

\begin{tcolorbox}[memorybox, title={Strategy memory entry --- inside-frame}]
\textbf{When to use:} \hl{A grid contains one large hollow
rectangular frame and several separate objects both inside and
outside it. The output keeps only the objects strictly enclosed
by the frame}, removes the frame itself, removes every outside
object, and leaves the kept inner objects unchanged in their
original positions on a same-size grid.

\smallskip
\textbf{Strategy:} (1) Find all connected non-background objects.
(2) \hl{Identify the frame as the large hollow rectangle, typically
the largest object by size and with an empty interior.}
(3) For every other object, compare its bounding box to the frame's
bounding box; \hl{keep only objects whose top, left, bottom, and
right bounds lie strictly inside the frame bounds, not touching the
frame border.}
(4) Start from a copy of the input or from an empty same-size grid.
(5) Erase the frame cells.
(6) Erase every object not strictly inside the frame.
(7) Preserve each inside object exactly as is: same shape, same
color layout, same absolute position.
(8) Leave all remaining cells as background.
\end{tcolorbox}

\begin{tcolorbox}[memorybox, title={Strategy memory entry --- key-marker (earlier extraction schema)}]
Use \hl{the top-left cell as a preserved control marker}; for every
other object, choose between two transforms according to whether
that marker matches a designated color.
\end{tcolorbox}

\section{Additional Experiment Details}

\subsection{ALFWorld}
\label{sec:alfworld-details}

\subsubsection*{Training data}

We use trajectories on the six standard ALFWorld task types, generated by gpt-4.1 or gpt-5-nano (see \cref{tab:alfworld-train}):
\begin{itemize}\itemsep=2pt
  \item \texttt{pick\_and\_place}
  \item \texttt{pick\_clean\_then\_place\_in\_recep}
  \item \texttt{pick\_two\_obj\_and\_place}
  \item \texttt{look\_at\_obj\_in\_light}
  \item \texttt{pick\_cool\_then\_place\_in\_recep}
  \item \texttt{pick\_heat\_then\_place\_in\_recep}
\end{itemize}
We use script-normalized task names; the upstream ALFWorld task type for \texttt{pick\_and\_place} is \texttt{pick\_and\_place\_simple}, with the \texttt{\_simple} suffix stripped during data preparation.

``$n{=}64$'' and ``$n{=}128$'' below refer to the number of trajectories \emph{per task type}, so a $n{=}64$ set contains $64 \times 6 = 384$ trajectories in total. All rollouts use \texttt{max\_steps}=50.

Three training-data variants are used across the experiments (\cref{tab:alfworld-train}):

\begin{table}[h]
\centering
\small
\begin{tabular}{lllrrl}
\toprule
Variant & generator & per type & total & filtering \\
\midrule
$n{=}64$ golden            & gpt-4.1      & 64  & 384 & only success-trajectory kept \\
$n{=}128$ golden           & gpt-4.1      & 128 & 768 & only success-trajectory kept \\
$n{=}64$ \texttt{with\_failures}    & gpt-5-nano   & 64  & 384 & none (mixed success/failure)  \\
\bottomrule
\end{tabular}
\caption{The three ALFWorld training-data variants differ on two axes: total size (384 vs 768 trajectories) and reward filtering. The two golden variants retain only success trajectories from gpt-4.1 rollouts; \texttt{with\_failures} retains both successes and failures from gpt-5-nano rollouts and is used for probes that test consolidation behaviour under imperfect inputs.}
\label{tab:alfworld-train}
\end{table}

After collection, every raw trajectory is converted into a structured memory item via the abstraction step of the SkillRL pipeline. We use gpt-5.4 as the abstraction LLM. The abstraction makes 3 LLM calls per trajectory and produces one item with three fields: \texttt{contextual\_description}, \texttt{refined\_trajectory}, and \texttt{strategic\_guidelines}.

\subsubsection*{Two-task-type probe set}

For one probe experiment we use a focused training subset that pairs two task types: \textbf{\texttt{pick\_and\_place}} (focal) and \textbf{\texttt{pick\_clean\_then\_place\_in\_recep}} (probe). The base memory at every probe point is a single-type consolidation that fixes the per-stage task type to \texttt{pick\_and\_place} and samples 32 abstracts per stage from the $n{=}64$ abstract pool; snapshots are taken at consolidation stage $S \in \{1, 20, 50, 100\}$. From each snapshot we run one additional consolidation iteration with 32 abstracts. The management model is gpt-5-nano and it has a max capacity of memory, which is 50.

\subsubsection*{Evaluation set}

We evaluate on \texttt{eval\_id\_balanced\_48}, a 48-episode subset of ALFWorld's in-distribution test split with 8 episodes per task type for the 6 task types listed above.

\subsubsection*{Metric}

Each ALFWorld episode has a binary success indicator (env reward $=1$ when the agent satisfies the goal within \texttt{max\_steps}=50, else 0). We report \textbf{wins/48} (the count of successful episodes) and the corresponding success rate in $[0,1]$ as the per-condition score on this evaluation set.

\subsection{WebShop}
\label{sec:webshop-details}

\subsubsection*{Training data}

WebShop training data come in two variants (\cref{tab:webshop-train}).

\textbf{Golden 128.} We sample 128 trajectories (seed $=42$) from the WebShop subset of the AgentGym AgentTraj-L corpus, filtered to score $=1.0$ (perfect-reward demonstrations only). We replay these trajectories through our WebShop environment instance (\texttt{num\_products}=1000) using a replay script that re-executes each AgentGym action sequence to obtain matching raw trajectories. The 128 sampled tasks are drawn sparsely from AgentGym's success-filtered train pool; their underlying WebShop session identifiers span the range $72$--$6772$ (sparse; 128 distinct IDs). This is the training set used for the AWM-style Workflow~8 ablation in \cref{sec:surgery-awm-w8}.

\textbf{Non-golden 128.} We use a separate, contiguous block of 128 WebShop sessions (session IDs $1500$--$1627$), disjoint from the test pool $[0, 500)$. On these sessions we collect 128 trajectories generated by gpt-5.4-mini directly, with no reward filtering. Most resulting trajectories have score $<1.0$ (\cref{tab:webshop-train}). This set tests how memory consolidation behaves when the input training trajectories are themselves imperfect.

\begin{table}[h]
\centering
\small
\begin{tabular}{p{2.2cm}p{3.6cm}p{2.5cm}p{3.4cm}}
\toprule
Variant & source / generator & session IDs & filtering / note \\
\midrule
golden 128     & replay from AgentGym's & $72$--$6772$ (sparse) & only score $=1.0$ kept; all expert demonstrations \\
non-golden 128 & gpt-5.4-mini self-rollout       & $1500$--$1627$ (contiguous) & none; $20/128$ with score $=1.0$ \\
\bottomrule
\end{tabular}
\caption{Composition of the WebShop training-data variants. The two variants use \emph{different} sets of 128 WebShop session IDs and serve different probes: golden 128 supplies expert demonstrations for AWM-style consolidation; non-golden 128 supplies a mixed-success training set for studying consolidation under imperfect inputs.}
\label{tab:webshop-train}
\end{table}

After collection, raw trajectories are abstracted into structured memory items via the same SkillRL abstraction pipeline used for ALFWorld.

\subsubsection*{Evaluation set}

We evaluate on \texttt{test50\_seed42\_no\_leak.json}, a 50-session subset of WebShop's test pool (\texttt{num\_products}=1000, seed $=42$) drawn from the index range $[0, 500)$. The 13 training-set session IDs (all from \textbf{Golden 128}) that fall within $[0, 500)$ are excluded from the sampling pool before drawing. Final session indices span $3$--$467$.

\subsubsection*{Metric}

Each WebShop episode terminates with a reward in $[0,1]$ computed from attribute match between the purchased product and the instruction (a perfect match yields reward $1.0$). If the agent never issues a \texttt{click[Buy Now]} action and the episode ends without a purchase (e.g., the agent times out at \texttt{eval\_max\_steps}), the reward is $0$. We report two quantities per condition: \textbf{wins/50} (the count of episodes with reward $=1.0$) and \textbf{mean reward} (arithmetic mean of episode rewards over the 50 evaluation sessions).

\subsection{ScienceWorld}
We select 15 medium-difficulty tasks from ScienceWorld for the memory consolidation experiments, where each selected task satisfies an average optimal action sequence length between 20 and 100. In total, these tasks contain \textbf{485} rollout cases. The task names and the number of rollout cases for each task are shown in Table~\ref{tab:scienceworld_tasks}.

Our ScienceWorld experiments are based on the CLIN setting, with several adaptations for studying long-horizon memory consolidation. Following CLIN, each rollout is presented together with an \annot{1}{\textit{evaluation report}}, which provides feedback about the success or failure of the corresponding attempt. We also follow \annot{2}{the CLIN-style constrained memory format}, where each memory is written as a unified natural-language rule using one of a small set of relation phrases, such as \texttt{MAY BE NECCESSARY to}, \texttt{SHOULD BE NECCESSARY to}, \texttt{MAY BE CONTRIBUTE to}, and \texttt{DOES NOT CONTRIBUTE to}.

Compared with the original CLIN update procedure, we introduce several modifications. First, we maintain a bounded memory bank and explicitly inform the LLM of the maximum memory-bank size and FIFO replacement policy, as highlighted in \annot{3}{the memory-capacity instruction}. Second, instead of only adding new memories, we allow multiple memory-bank operations, including \annot{4}{delete}, \annot{5}{replace}, \annot{6}{add}, and \annot{7}{retain}. Third, we feed rollouts to the LLM in mini-batches and ask the model to compare the buffered rollouts jointly before editing the memory bank, as shown in \annot{8}{the mini-batch consolidation instruction}. Finally, after each full pass over all rollouts for a given number of consolidation rounds, we evaluate the current memory bank on a fixed ScienceWorld test set. This test set contains \textbf{44} cases sampled from the same 15 selected tasks, with 2--4 cases per task depending on the number of available variants.

\begin{tcolorbox}[memorybox, title={Memory update prompt template for ScienceWorld}]
\scriptsize
\ttfamily
\setlength{\parindent}{0pt}
\setlength{\parskip}{3pt}

You are given BUFFERED ROLLOUTS, which are multiple recent rollouts that have NOT been summarized yet. Each rollout is presented as a surface-level summary (compressed narrative), and is accompanied by an \annot{7}{EVALUATION REPORT} indicating the success of that attempt.

You are also provided with PREVIOUS LEARNINGS collected from the agent's earlier attempts across interleaved tasks and environment variants. Some memories may be specific to one task or one environment configuration, while others may generalize across related tasks or across multiple environments. TASK indicates the task description. EPISODE indicates when that learning was created. PREVIOUS LEARNINGS also have \annot{1}{EVALUATION REPORTs} indicating how successful the respective attempt was.

Your job is to UPDATE the FULL MEMORY BANK now, based on the BUFFER of recent rollouts that have NOT been summarized yet. You may:
(a) \annot{4}{delete outdated memories},
(b) \annot{5}{replace several memories with a NEW set of memories},
(c) \annot{6}{add new memories based on the buffered rollouts},
(d) \annot{7}{retain selected buffered rollouts as concrete case records in memory}.

\annot{8}{In this memory-update call, process this MINI-BATCH of <buffered\_rollout\_count> buffered rollouts together (configured batch size k=<pool\_batch\_size>). Compare the rollout summaries jointly before editing the MEMORY BANK. If several rollouts support the same general lesson, prefer a single compact memory that captures the shared pattern rather than duplicating near-identical memories. If the rollouts disagree, preserve only the parts that are clearly supported by the evidence.}

\annot{3}{The MEMORY BANK can hold at most <memory\_bank\_max\_size> entries and currently contains <current\_size> entries.
The MEMORY BANK is maintained with a FIFO policy: if it grows beyond capacity, the oldest entries will be dropped first.}

When adding or replacing memories, ONLY use rollouts you explicitly select from the buffer (used\_rollout\_ids). Do NOT use or rely on any rollout outside the buffer.

Because the training schedule alternates across tasks and variants, prefer memories that capture reusable preconditions, transferable mistakes, or environment knowledge that can generalize to related problems. Keep a memory narrowly task-specific only when the evidence strongly suggests it.

\annot{2}{When writing new memories, use the SAME format as the examples below:
Each numbered item in the summary can ONLY be of the form:
X MAY BE NECCESSARY to Y.
X SHOULD BE NECCESSARY to Y.
X MAY BE CONTRIBUTE to Y.
X DOES NOT CONTRIBUTE to Y.}

Use retain when a rollout's raw narrative is itself worth keeping as a concrete case record. Use add / replace when you want abstracted lessons instead.
A single rollout may support both add and retain in the same response.

Return ONLY a JSON object with this schema:

\{
  "actions": [
    \{"type": "delete", "ids": [int, ...]\},
    \{"type": "replace", "ids": [int, ...], "new\_memories": ["<memory1>", "<memory2>", ...], "used\_rollout\_ids": [int, ...]\},
    \{"type": "add", "new\_memories": ["<memory1>", "<memory2>", ...], "used\_rollout\_ids": [int, ...]\},
    \{"type": "retain", "rollout\_ids": [int, ...]\}
  ]
\}

Notes:
- IDs refer to the MEMORY BANK entry IDs shown below.
- used\_rollout\_ids refer to BUFFERED ROLLOUT IDs shown below, and indicate which rollouts were used to create the new memories.
- Include an empty actions list only if you judge that no edit is needed this round.
- For each retained rollout, the corresponding SURFACE SUMMARY will be inserted into the MEMORY BANK verbatim as its own memory entry.
\end{tcolorbox}

\begin{table}[t]
\centering
\small
\caption{ScienceWorld tasks used for memory consolidation. }
\label{tab:scienceworld_tasks}
\begin{tabular}{lc}
\toprule
\textbf{Task name} & \textbf{\# Rollout cases} \\
\midrule
\texttt{boil} & 28 \\
\texttt{change-the-state-of-matter-of} & 28 \\
\texttt{chemistry-mix} & 30 \\
\texttt{chemistry-mix-paint-tertiary-color} & 34 \\
\texttt{freeze} & 28 \\
\texttt{grow-fruit} & 40 \\
\texttt{grow-plant} & 40 \\
\texttt{identify-life-stages-1} & 12 \\
\texttt{measure-melting-point-known-substance} & 40 \\
\texttt{measure-melting-point-unknown-substance} & 40 \\
\texttt{melt} & 27 \\
\texttt{power-component-renewable-vs-nonrenewable-energy} & 18 \\
\texttt{test-conductivity} & 40 \\
\texttt{test-conductivity-of-unknown-substances} & 40 \\
\texttt{use-thermometer} & 40 \\
\midrule
\textbf{Total} & \textbf{485} \\
\bottomrule
\end{tabular}
\end{table}

\subsection{AppWorld}
\label{sec:appworld-details}

\subsubsection*{Training data}

We use 140 trajectories collected on AppWorld's official \texttt{train} and \texttt{dev} splits (Trivedi et al., 2024) as the shared training set across all memory-induction methods evaluated in this work.

The 90-task train segment uses AppWorld's officially released ReAct baseline trajectories, produced with GPT-4o (gpt-4o-2024-05-13). The 50-task dev segment uses ReAct trajectories collected with Qwen3.5-27B on a stratified-by-difficulty sample of the dev split, run under AppWorld's standard ReAct interface. Composition is summarized in \cref{tab:appworld-train}. The mean Task Goal Completion (TGC) over the combined 140 trajectories is $61.4\%$ ($86 / 140$): $52.2\%$ on the train segment and $78.0\%$ on the dev segment.

For methods that require gold-quality input, we additionally collect a parallel set of 140 trajectories on the same task IDs by replaying AppWorld's released ground-truth solutions through the environment API, yielding a $100\%$-pass training set with identical task IDs and ordering.

\begin{table}[h]
\centering
\small
\begin{tabular}{lrll}
\toprule
Subset & N & Source split & Trajectory backend \\
\midrule
Train segment & 90  & AppWorld \texttt{train} (90 tasks) & GPT-4o      \\
Dev segment   & 50  & AppWorld \texttt{dev}\ \,(57 tasks) & Qwen3.5-27B \\
\midrule
Combined      & 140 & ---                     & ---         \\
\bottomrule
\end{tabular}
\caption{Composition of the 140-trajectory AppWorld training set.}
\label{tab:appworld-train}
\end{table}

Each training datapoint consists of a task identifier, a final pass/fail outcome, and a Qwen3.5-27B--generated prose summary of the full ReAct trajectory (thoughts, actions, and observations of all steps), used in place of the raw token-level trajectory to keep per-entry length tractable for prompt-time concatenation.

\subsubsection*{Evaluation set}
We evaluate on a 100-task subset drawn from AppWorld's two official test splits (\texttt{test\_normal} with $168$ tasks and \texttt{test\_challenge} with $417$) by proportional stratified sampling on difficulty: 50 tasks from each split, seed $=42$, with no overlap. All conditions are evaluated on the same 100 task IDs.

\subsubsection*{Metric}
Each AppWorld task is paired with a set of Python unit tests. The official evaluator replays the agent's trajectory against the resulting database state and runs these tests; a task is counted as a success only when all of its unit tests pass. Following AppWorld's official protocol, we report Task Goal Completion (TGC) as the sole metric: the percentage of evaluation tasks whose success indicator equals one. All TGC values reported in the main tables are arithmetic means over the 100 evaluation tasks.
\section{Case studies of memory failure modes}
\label{sec:memory-case-studies}

\subsection{Erosion: good memory degrading over time}
\label{sec:case-erosion}

\paragraph{Setup.} The memory under study is produced by a 200-stage iterative consolidation run on ALFWorld. The base memory at stage~0 is a 10-item seed obtained by passing all 768 abstracts of the $n{=}128$ golden ALFWorld training set (\cref{sec:alfworld-details}) through a single gpt-5.4 consolidation call that emits a free-count list of task-type-agnostic insights; the model picks the count, with no fixed target. Each subsequent stage feeds gpt-5-nano a fresh mini-batch of $6$ abstracts ($1$ per task type, drawn from the $n{=}64$ subset of the same six-type pool) together with the current memory; the management model rewrites the entire memory at every stage and forces overwrites rather than appends. Besides, the memory capacity is 50. We evaluate the consolidated memory at every 20 stages on the held-out balanced ALFWorld split (defined in \cref{sec:alfworld-details}; n=48 episodes) using three Qwen3.5 sizes as rollout policies --- Qwen3.5-4B, Qwen3.5-9B, and Qwen3.5-27B --- with each model's reasoning mode disabled (the ``no-thinking'' setting). The full memory is supplied in the prompt context at every step, so the rollout sees every item in every snapshot.

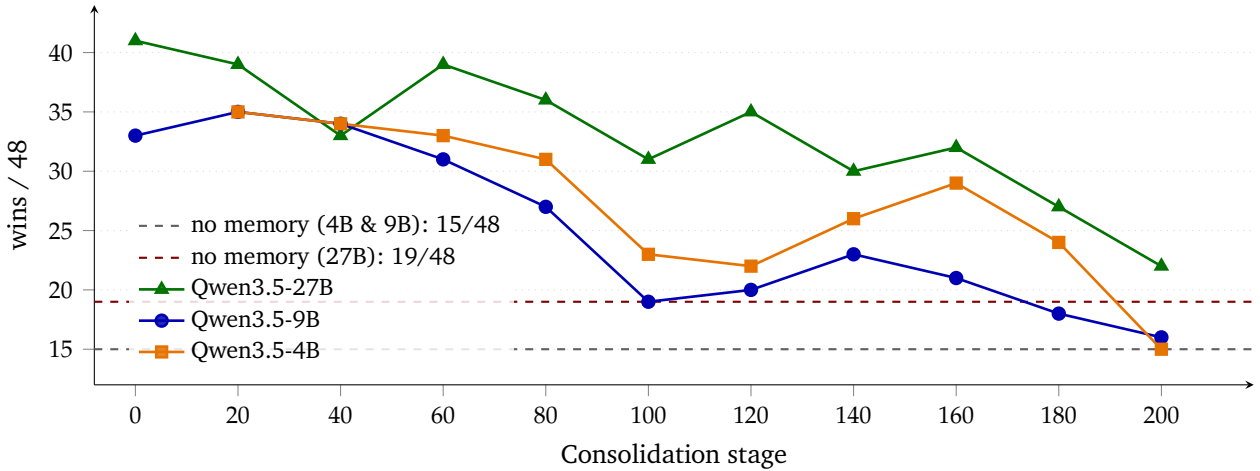
\begin{figure}[h]
\centering
\begin{tikzpicture}
\begin{axis}[
  width=0.95\linewidth,
  height=6.6cm,
  xlabel={Consolidation stage},
  ylabel={wins / 48},
  xmin=-8, xmax=218,
  ymin=12, ymax=44,
  xtick={0,20,40,60,80,100,120,140,160,180,200},
  ytick={15,20,25,30,35,40},
  ymajorgrids=true,
  grid style={dotted,gray!30},
  axis x line=bottom,
  axis y line=left,
  every axis label/.append style={font=\small},
  every tick label/.append style={font=\footnotesize},
  tick align=outside,
  tick pos=left,
  legend pos=south west,
  legend cell align=left,
  legend style={font=\footnotesize, draw=none, fill opacity=0.85, text opacity=1},
]
\addplot[no marks, dashed, gray!75!black, thick] coordinates {(-8,15) (218,15)};
\addlegendentry{no memory (4B \& 9B): 15/48}
\addplot[no marks, dashed, red!50!black, thick] coordinates {(-8,19) (218,19)};
\addlegendentry{no memory (27B): 19/48}
\addplot[mark=triangle*, line width=1.0pt, mark size=2.4pt, green!45!black] coordinates {
  (0,41) (20,39) (40,33) (60,39) (80,36) (100,31) (120,35) (140,30) (160,32) (180,27) (200,22)
};
\addlegendentry{Qwen3.5-27B}
\addplot[mark=*, line width=1.0pt, mark size=2.2pt, blue!70!black] coordinates {
  (0,33) (20,35) (40,34) (60,31) (80,27) (100,19) (120,20) (140,23) (160,21) (180,18) (200,16)
};
\addlegendentry{Qwen3.5-9B}
\addplot[mark=square*, line width=1.0pt, mark size=2.0pt, orange!90!black] coordinates {
  (20,35) (40,34) (60,33) (80,31) (100,23) (120,22) (140,26) (160,29) (180,24) (200,15)
};
\addlegendentry{Qwen3.5-4B}
\end{axis}
\end{tikzpicture}
\caption{Iterative consolidation erodes the memory's value across three rollout sizes. By stage~200, the 4B and 9B rollouts collapse to within $1$ win of their no-memory baselines (15 vs.\ 15; 16 vs.\ 15); the 27B rollout retains a $3$-win gap above its baseline (22 vs.\ 19). The 4B and 9B rollouts peak early (stage~20, 35/48 wins each) before declining; the 27B rollout's peak is the un-consolidated base seed at stage~0 (41/48), so consolidation never helps it. Memory size is bounded by \texttt{max\_memory\_size=50} throughout, so the loss is in content quality, not item count.}
\label{fig:erosion-curve}
\end{figure}

\paragraph{The erosion curve.} \cref{fig:erosion-curve} plots wins/48 for the three rollout sizes at all evaluated stages, against the corresponding no-memory baselines (15/48 for both 4B and 9B; 19/48 for 27B). All three sizes show the same qualitative pattern: an early peak followed by a general decline (with small recoveries) to stage~200. The 4B and 9B rollouts each gain a small bump at stage~20 (35/48 wins), then erode to 15/48 (4B) and 16/48 (9B). The 27B rollout's peak is at stage~0 (41/48), so for the 27B rollout consolidation never helps; it erodes to 22/48 by stage~200. By stage~200 the gap to the no-memory baseline collapses for the 4B and 9B rollouts ($\Delta{=}0$ and $\Delta{=}1$ respectively); the 27B rollout retains a $\Delta{=}3$ gap above its baseline (22/48 vs.\ 19/48). Memory size stays bounded at 38--50 items by the cap throughout, so the erosion is in content quality, not item count.

\paragraph{What is in the memory at the peak.} At stage~20 the 42 items each cover a distinct decision-relevant insight. We reproduce items~0--5 verbatim below; items~6--41 expand on this cover set with concrete recipes for heat/cool/clean tasks, look-at-light tasks, and multi-object planning.

\begin{tcolorbox}[memorybox, title={Stage 20 memory (peak, 42 items) --- items 0--5 verbatim}, label={box:erosion-stage20}]
\textbf{Item 0.} Most ALFWorld tasks reduce to a small set of reusable plans. For plain pick-and-place: find object, pick it up, go to destination, place it. For cool/heat/clean tasks: find object, pick it up, apply the required state change at the correct appliance/fixture, then move it to the goal receptacle. For look-at-obj-in-light tasks: get the target under or near the desklamp, turn on/use the desklamp, then examine or hold the target there. Recognizing the task type early lets you prioritize the right fixtures: fridge for cooling, microwave for heating, sinkbasin for cleaning, desklamp for light-examination.

\smallskip
\textbf{Item 1.} When the object location is unknown, systematic search beats wandering. Common sources recur by room type: kitchen objects often appear on countertop, diningtable, in fridge, cabinet, drawer, sinkbasin, stoveburner, or microwave; living-room objects often appear on sofa, armchair, coffeetable, sidetable, desk, shelf, dresser, or tvstand; bathroom objects often appear on toilet, toiletpaperhanger, handtowelholder, bathtubbasin, countertop, cabinet, or drawer. Open closed containers when needed, and once you find one instance of the target type, revisit that source because the second instance is often in the same place.

\smallskip
\textbf{Item 2.} If the goal location is a closed container like a drawer, cabinet, fridge, microwave, or safe, opening it early can simplify the rest of the task. This avoids carrying the object to the destination only to discover it is closed, and it is especially helpful in two-object tasks because the container remains ready for repeated placements. The same applies when the destination is also the object source: open once, retrieve object, do the state change, then return and place.

\smallskip
\textbf{Item 3.} \anchor{Cooling and cleaning are usually simpler than heating}: once you hold the target object, go to the fridge or sinkbasin and apply `cool X with fridge' or `clean X with sinkbasin'. You often do not need to insert the object into the appliance/container first. This means a strong default is: pick up target, go to the relevant fixture, apply the state change directly, then place at destination.

\smallskip
\textbf{Item 4.} Many state-change tasks start with the target object already on or in the goal receptacle. In these cases, the object still must be picked up, changed to the required state, and then placed back. Examples include cooling an object already on a diningtable or in a coffeemachine, heating an object that ends up back in the fridge or diningtable, and cleaning an item that returns to the same surface. Do not stop just because the object is already at the destination.

\smallskip
\textbf{Item 5.} \anchor{The robust default for `pick two obj and place'} is: find one target, pick it up, carry it to the destination, place it, return for the second, and repeat. If both objects are visible and inventory permits, carrying both can help, but many successful trajectories simply shuttle one at a time. Once a source location is known to contain multiple copies, revisit it directly instead of re-searching the whole room.
\end{tcolorbox}

\paragraph{What is in the memory after erosion.} At stage~200 the 38 items collapse onto a small number of overlapping templates. By keyword scan, 22 of the 38 items mention the desklamp, 28 mention multi-object handling, and 33 mention systematic room-by-room search; the templates overlap heavily (most items mention more than one). Items~3 and~5 are byte-identical. The concrete meta-strategies of stage~20 --- the cooling-vs-heating asymmetry of item~3, the shuttle default of item~5, the recognize-task-type-early principle of item~0 --- do not appear in concrete actionable form in any of the five representative items below, and a full scan of the 38 items finds none that retain the cooling-vs-heating asymmetry as a separable insight.

\begin{tcolorbox}[memorybox, title={Stage 200 memory (eroded, 38 items) --- representative items}, label={box:erosion-stage200}]
\textbf{Item 3.} \annot{1}{Pattern: For a single target inside a container, open the container to access the object, pick it into the inventory, move directly to the destination surface, and place the object there.} Planning: always consider planning/visibility by keeping the destination in view or planning a route that preserves visibility; open destinations early to simplify access. Unknown-location handling: if location is unknown, perform a systematic room-by-room search to locate and acquire the target before planning. Example: retrieve a candle from a cabinet and place it in the toilet.

\smallskip
\textbf{Item 5.} \annot{1}{Pattern: For a single target inside a container, open the container to access the object, pick it into the inventory, move directly to the destination surface, and place the object there.} Planning: always consider planning/visibility by keeping the destination in view or planning a route that preserves visibility; open destinations early to simplify access. Unknown-location handling: if location is unknown, perform a systematic room-by-room search to locate and acquire the target before planning. Example: retrieve a candle from a cabinet and place it in the toilet.

\smallskip
\textbf{Item 8.} \annot{2}{Always illuminate targets with the desklamp during planning and fetching}; relocate targets onto a lamp-hosting surface or move the lamp to the target surface and switch it on; use lamp-assisted look-at to maximize visibility and minimize wandering; Planning/visibility: keep targets in view during planning and fetching; open destinations early; In multi-object tasks, plan the fetch order to minimize wandering and travel distance; plan their final arrangement on the destination surface when carrying two objects; Unknown-location handling: perform a systematic room-by-room search to locate targets and the lamp, acquire targets, and carry them to the destination while preserving visibility of destinations during planning.

\smallskip
\textbf{Item 9.} \annot{2}{Always illuminate targets with the desklamp during planning and fetching}; relocate targets onto a lamp-hosting surface or move the lamp to the target surface and switch it on; use lamp-assisted look-at to maximize visibility and minimize wandering. Planning/visibility: keep targets in view during planning and fetching; open destinations early. Unknown-location handling: perform a systematic room-by-room search to locate targets and the lamp, acquire targets, and carry them to the destination while preserving visibility of destinations during planning. For two-object tasks: plan final arrangement on the destination surface when carrying two objects; if the inventory can carry two objects, carry two; otherwise place the first object, fetch the second, and then place it. Distractors: set aside held distractors to free space.

\smallskip
\textbf{Item 10.} Pattern: \annot{2}{Always illuminate targets with the desklamp during planning and fetching}; relocate targets onto a lamp-hosting surface or move the lamp to the target surface and switch it on; use lamp-assisted look-at to maximize visibility and minimize wandering. Implementation: 1) Before approaching a target, ensure the desklamp is on. 2) If needed, reposition the lamp or move the target onto a surface that can host the lamp to achieve optimal lighting. 3) Use lamp-assisted look-at to orient yourself and inspect the target under the light, reducing uncertainty and stray movement. 4) Planning/visibility: keep the target in view during planning and fetching; open destinations early to simplify access. Unknown-location reinforcement: if location is unknown, perform a systematic search to locate the target and the lamp, acquire targets, and carry them to the destination while preserving destination visibility during planning.
\end{tcolorbox}

\begin{annotations}[Erosion modes]
\annotitem{1}{Byte-identical duplicate}{Items~3 and~5 of stage~200 (\cref{box:erosion-stage200}) are character-for-character identical, occupying two of the 50 cap slots with the same content. The pipeline applies no post-hoc deduplication step, so gpt-5-nano emitting the same string twice within one consolidation stage produces two stored copies.}
\annotitem{2}{Paraphrase cluster (desklamp directive)}{Items~8, 9, and 10 of stage~200 each open with a near-verbatim restatement of ``always illuminate targets with the desklamp during planning and fetching'', followed by progressively rephrased downstream advice. 21 of the 38 items at stage~200 contain this opener. The desklamp directive only applies to look-at-in-light tasks, which are 8/48 of the eval split; the memory devotes 21/38 ($>$55\%) of its prompt-context slots to it, crowding out the diverse meta-strategies of stage~20.}
\annotitem{3}{Lost meta-strategy}{The cooling-vs-heating asymmetry of stage~20 item~3 (\anchor{Cooling and cleaning are usually simpler than heating} --- the agent can apply \texttt{cool X with fridge} or \texttt{clean X with sinkbasin} \emph{without} inserting the object first) is absent from every item at stage~200. The 50-item cap evicted it to make room for paraphrases of the desklamp template. The shuttle default for two-object tasks of stage~20 item~5 (\anchor{The robust default for `pick two obj and place'}) is also absent in this concrete actionable form, replaced by 15 paraphrases of a high-level ``carry two in a single trip'' directive that does not specify the fallback.}
\end{annotations}

\subsection{Collapsed memory: single-stage merge to one item}
\label{sec:case-collapse}

\paragraph{Setup.} The iterative-consolidation pipeline matches \cref{sec:case-erosion}: gpt-5-nano management model, memory capacity is 50 items, the same six ALFWorld task types, and the same gpt-5.4 abstraction step. The per-stage data sampling differs in one respect. Each stage feeds the manager a mini-batch of $6$ trajectories drawn from a single (randomly chosen) task type, rather than a flat sample across all six types. Rollout policies are the three Qwen3.5 sizes (Qwen3.5-4B, Qwen3.5-9B, Qwen3.5-27B); the evaluation split is the same balanced ALFWorld split as in \cref{sec:case-erosion}. We focus on the single-stage transition $S{=}168 \to S{=}169$ in this run.

\paragraph{The collapse event.} At stage $168$ the memory contains $50$ items totalling $48{,}506$ characters (i.e., the cap is saturated, average $\sim$$970$ characters per item). One consolidation step later, at stage $169$, the memory contains \emph{a single item} of $1{,}960$ characters. The manager has merged $50$ structured items into one numbered ``unified loop'' that tries to cover every ALFWorld task type at once. The collapse is concentrated in this single stage. The preceding stage $167$ has $49$ items; the following stage $170$ already begins to rebuild ($2$ items, $3{,}410$ chars). \cref{tab:collapse-event} reports the rollout-level effect of this single-stage transition under the three Qwen3.5 sizes, against the corresponding no-memory baselines. \cref{box:collapse-stage169} reproduces the single stage-$169$ item verbatim.

\begin{table}[h]
\centering
\small
\begin{tabular}{lcccc}
\toprule
Rollout policy & no memory & stage 168 (50 items) & stage 169 (1 item) & $\Delta$ \\
\midrule
Qwen3.5-4B  & 15/48 & 35/48 & 29/48 & $-6$ \\
Qwen3.5-9B  & 15/48 & 36/48 & 26/48 & $-10$ \\
Qwen3.5-27B & 19/48 & 37/48 & 24/48 & $-13$ \\
\bottomrule
\end{tabular}
\caption{One consolidation step (50 items $\to$ 1 super-item) costs every rollout $6$--$13$ wins, with the largest absolute drop at the largest rollout (Qwen3.5-27B, $-13$). Across all three sizes the post-collapse memory still sits above the no-memory baseline, so the collapse degrades the memory's value but does not zero it out. Stage 168 is the last pre-collapse stage; stage 169 is the post-collapse single-item state. wins/48 on the balanced ALFWorld split, n=48 episodes per cell, single run.}
\label{tab:collapse-event}
\end{table}

\begin{tcolorbox}[memorybox, title={Stage 169 memory (after collapse, 1 item, 1{,}960 chars) --- verbatim}, label={box:collapse-stage169}]
\textbf{Item 0.} \annot{1}{Use a single, repeatable loop to handle both single and multi-object tasks across cleaning, cooling, heating, and look-at-in-light.} 1)~Systematically search common room sources (kitchen countertop, fridge, cabinet, drawer, dining table, living surfaces) to locate each target; if a known source can hold multiple copies, reuse it for subsequent objects to avoid re-search. 2)~Pick each target; if a state change is required (cool X with fridge, heat X with microwave, or clean X with sinkbasin), perform it while holding or immediately after acquisition; open appliances early as needed. 3)~Navigate to the destination surface or container; if the destination is a container, open it before placing. 4)~Place the object; for multi-object tasks, repeat for each item. 5)~If two identical objects must go to the same destination, fetch Object1 from a known multi-copy source and place it, then fetch Object2 from a similarly likely multi-copy source (prefer reusing the same source) and place it at the same destination; carry both if inventory permits. 6)~If an object starts at the destination (or inside a container), re-pick it and place again after applying any necessary state change to ensure task closure. 7)~Manage distractors by moving unused tools aside to free space; batch carries when possible and reuse multi-copy sources for faster throughput. 8)~After completing a batch, return to a known multi-copy source for the next run. 9)~\annot{2}{Look-at-in-light-specific extension: anchor a central desklamp location, co-locate the target and lamp by moving one to the other's surface if needed, activate the lamp, and examine/hold the object under illumination; batch look-at tasks using the same lamp position and multi-copy sources when possible.} 10)~Pitfalls to avoid: forgetting to re-pick items that started on the destination; forgetting to reopen microwaves or containers when needed; forgetting to retrieve heated/cooled/cleaned objects before placing.
\end{tcolorbox}

\begin{annotations}[Collapse modes]
\annotitem{1}{Over-merge into a single ``unified loop''}{The $50$ structured items of stage~$168$ each described one decision-relevant insight (task taxonomy, state-change shortcuts, two-object recipes, look-at protocol, etc.). Stage~$169$ replaces all of them with a single $10$-step procedure framed as ``a single, repeatable loop to handle both single and multi-object tasks across cleaning, cooling, heating, and look-at-in-light''. Distinctions that were separate items in stage~$168$ (e.g., the cooling-vs-heating asymmetry, the inventory-permitting two-object shuttle, the lamp-co-location requirement) are reduced to numbered sub-steps with no inter-step preference, no failure cases, and no domain conditions.}
\annotitem{2}{All task-type-specific procedures collapsed into one item}{Stage~$169$'s sub-step~$9$ encodes the entire look-at-in-light protocol (lamp anchoring, co-location, illumination, batching) as a single sub-clause inside an item that primarily reads as a generic pick-and-place loop. Look-at-in-light tasks ($8/48$ of the eval split) now share retrieval bandwidth with a memory whose lead framing is unrelated. The same compression pattern occurs for cleaning, cooling, and heating: each becomes a half-sentence inside sub-step~$2$.}
\annotitem{3}{Magnitude scales with rollout strength}{In \cref{tab:collapse-event} the absolute win drop at $S{=}168 \to 169$ grows monotonically with rollout size ($-6$ for Qwen3.5-4B, $-10$ for 9B, $-13$ for 27B; n=3 sizes, single run, no significance test). One reading is that stronger rollouts extract more from the structured $50$-item memory and therefore lose more when those items are merged into one. The post-collapse memory still beats no-memory for every size, so the collapse removes most of the consolidation's value but does not destroy it entirely.}
\end{annotations}

\subsection{Useless memory items}
\label{sec:case-useless}

We distinguish two failure modes for memory items that are syntactically well-formed but supply no decision-relevant information. Both modes survive across stages because nothing contradicts them, and both consume retrieval bandwidth without shifting the action distribution.

\paragraph{Overgeneralized.} Items whose recommendation applies to any state (``be careful'', ``consider all options'', ``plan ahead''). The clause names no observable cue and shifts no prior over candidate actions when retrieved.

We pull three verbatim items from an ExpeL iterative consolidation run on ALFWorld ($n{=}64$ trajectories per task type, gpt-5.4 base model, gpt-5-nano management, 200 mini-batch consolidation stages) at stage~$200$. The stage-200 memory has 9 active rules; we display the top-1, top-2, and a low-vote item (Item~7, vote~$=2$; the lowest is Item~8 with vote~$=1$).

\begin{tcolorbox}[memorybox, title={Overgeneralized memory items (ExpeL ALFWorld stage 200, verbatim)}, label={box:useless-overgen}]
\textbf{Item 0 (vote $=99$).} \annot{1}{Prioritize actions that directly advance the current subgoal} using \annot{2}{the nearest feasible fixture or item}; verify prerequisites before acting; minimize travel; \annot{3}{interleave subgoals when beneficial}; attempt to combine state-change and final placement in a single near-term step when possible; avoid assuming an item's state without explicit verification.

\smallskip
\textbf{Item 1 (vote $=48$).} Maintain a lightweight explicit world model of item locations, containment relationships, and state attributes across environments; update it after every action; verify prerequisites before use; use the model to guide planning, detect contradictions, and resolve ambiguities when multiple identical items exist; keep post-action states consistent for robust planning.

\smallskip
\textbf{Item 7 (vote $=2$).} Prefer reversible actions and plan with contingencies; always consider alternative paths if the primary plan might fail, to minimize backtracking and avoid dead ends.
\end{tcolorbox}

\begin{annotations}[Overgeneralization modes]
\annotitem{1}{Tautological directive}{An agent already pursuing a subgoal already prefers actions that advance it; the clause names no observable cue and so leaves the action distribution unchanged at retrieval.}
\annotitem{2}{Vacuous ranking criterion}{``Nearest feasible fixture or item'' offers a preference but no operational definition of feasibility or distance, no rule for selecting among candidates that all look feasible, and no fallback when none is.}
\annotitem{3}{Meta-directive without operationalization}{``Interleave subgoals when beneficial'' is a control-flow recommendation with no condition for ``beneficial''; the item supplies no procedure that produces the benefit signal at action-selection time.}
\end{annotations}

\paragraph{Worked example: how an ExpeL highest-voted insight collapsed.} The same run produced Item~0 above by repeatedly applying ExpeL's \texttt{EDIT} operation to the top slot. By stage~$200$ the slot had accumulated $99$ \texttt{AGREE}/\texttt{EDIT} votes, which would naively suggest extreme value. The vote counter, however, increments on \texttt{EDIT} but is not reset when an \texttt{EDIT} replaces the underlying concept. The score therefore tracks slot edit-volume rather than current content quality. Tracing the slot through stages, its content was rewritten through at least three distinct concepts (\cref{box:expel-drift}).

\begin{tcolorbox}[memorybox, title={ExpeL top-1 slot content drift across stages (verbatim)}, label={box:expel-drift}]
\textbf{Stage 0 (vote $=28$).} For tasks phrased as ``examine/look at X with/by/using Y,'' first locate both objects, then prefer the environment's direct task-relevant interaction (e.g., use Y, examine X with Y, or examine X if Y has been activated); only move objects if explicit admissible actions or clear failure show that relocation is required.\\
\textit{$\to$ Concrete: explicitly names the \texttt{look\_at\_obj\_in\_light} examine-with-Y interaction pattern.}

\smallskip
\textbf{Stage 80 (vote $=46$).} Prioritize the simplest action that directly advances the current subgoal with the fewest prerequisites and minimal detours, while also favoring actions that reuse already-opened receptacles or inventory-held items to minimize movement; also consider steps that reduce overall future effort.\\
\textit{$\to$ Generic planning heuristic; the task-type cue is gone.}

\smallskip
\textbf{Stage 200 (vote $=99$).} Prioritize actions that directly advance the current subgoal using the nearest feasible fixture or item; verify prerequisites before acting; minimize travel; interleave subgoals when beneficial; attempt to combine state-change and final placement in a single near-term step when possible; avoid assuming an item's state without explicit verification.\\
\textit{$\to$ Tautology; applies to any agent benchmark.}
\end{tcolorbox}

The slot's $99$ votes were earned across three different concepts (examine-with-Y interaction $\to$ generic action selection $\to$ tautology). At stage~$200$ only the third concept is in the slot; the other $\sim 80$ votes were accumulated by predecessors that the \texttt{EDIT} operation has since overwritten. The vote score is therefore not a measure of the current rule's value to the agent.

\paragraph{Overspecified.} Items so narrowly tied to a past trajectory state or to one (object, surface, appliance) tuple that they cannot transfer once that condition is not present. They survive in the cap because nothing contradicts them; they take up retrieval bandwidth on tasks they do not match.

We pull three verbatim items from an ACE iterative consolidation run on ALFWorld (6 abstracts sampled flat-randomly across all six task types per stage; $n{=}128$ trajectories per type; analyzer dedup threshold $0.75$; both reflector and curator $=$ gpt-5.4-mini). Each item is an \texttt{ADD} operation recorded in the curator's operation log. Reflector and curator are the two ACE roles: the reflector emits bullet candidates from each trajectory, the curator decides which are added to or modified in the running memory, and the analyzer merges any pair of items whose embedding similarity exceeds the dedup threshold.

\begin{tcolorbox}[memorybox, title={Overspecified memory items (ACE ALFWorld curator ADDs, verbatim)}, label={box:useless-overspec}]
\textbf{Item A.} \annot{1}{In \texttt{look\_at\_obj\_in\_light} tasks, follow the environment's explicit affordances to complete the setup: if the vase has been moved onto the desk, turn on the desklamp, then verify the vase is on the desk and finish with the available `take vase from desk' action.} Treat the final take-from-desk step as the success-confirming move rather than assuming relocation alone is enough.

\smallskip
\textbf{Item B.} \annot{2}{For \texttt{pick\_heat\_then\_place\_in\_recep} tasks, once the target mug is confirmed hot, stop doing additional checks or searches and immediately place/move it into the target receptacle.} If the mug is already in hand and hot, the completion action is usually a direct `put/place mug in cabinet' step; do not reopen the microwave, re-verify the temperature, or look for another intermediary action.

\smallskip
\textbf{Item C.} \annot{3}{For \texttt{pick\_cool\_then\_place\_in\_recep} tasks involving a mug, prioritize exposed-surface search for the exact mug first, then perform one supported cooling action with the fridge, immediately confirm the same mug is still in inventory, and go straight to the already-open cabinet for final placement.} Do not keep searching after the correct mug is found, and do not return to the fridge once cooling has succeeded unless the environment explicitly says the cool action failed.
\end{tcolorbox}

\begin{annotations}[Overspecification modes]
\annotitem{1}{Single-instance object$+$surface pinning}{Item~A names two specific object instances (\texttt{vase}, \texttt{desk}) and one specific completion action (\texttt{take vase from desk}). The action sequence does not transfer to a look-at task with a different target (e.g., \texttt{cd} on \texttt{shelf}) because both the object handle and the surface change.}
\annotitem{2}{Single-receptacle completion pinning}{Item~B fixes the heating appliance to \texttt{microwave} and the destination of \texttt{pick\_heat\_then\_place} to \texttt{cabinet}. ALFWorld's actual destinations for this task type include \texttt{countertop}, \texttt{fridge}, \texttt{sidetable}, \texttt{shelf}, etc.; the rule's \texttt{put/place mug in cabinet} phrasing fires only on (mug, cabinet) instances.}
\annotitem{3}{Compound condition over object$+$appliance$+$destination}{Item~C requires (mug \emph{and} fridge \emph{and} an already-open cabinet) for its action sequence to apply. Each conjunct narrows the matching set further; the resulting rule fires on a tiny subset of the eval split and contributes nothing on tasks that do not involve all three.}
\end{annotations}

\section{WebShop}
\label{sec:webshop}

\subsection{Memory artifacts from prior methods}
\label{sec:webshop-artifacts}

This section reproduces, verbatim, two memory artifacts from prior consolidation methods on WebShop (Yao et al., 2022). They are intended as concrete reference points for the failure-mode taxonomy in \cref{sec:memory-case-studies} --- not as quantitative comparisons. For each artifact, report: method version and configuration (model, rewrite cadence), WebShop split and seed, and the trajectory step or consolidation round at which the snapshot was taken.

\subsubsection{Healthy snapshots: ACE and Dynamic Cheatsheet}
\label{sec:webshop-ace-dc}

We display one mid-run snapshot per method, taken after enough consolidation has happened to populate the memory but before any visible degradation. Both runs use $n{=}64$ trajectories from the golden WebShop set and gpt-5.4-mini as the consolidation model.

\paragraph{ACE healthy.} ACE keeps a flat list of bullet items grouped into named sections (\textsc{Common mistakes to avoid}, \textsc{Problem-solving heuristics}). At stage~$100$ of a $192$-stage run with dedup threshold $0.75$, the memory contains $32$ bullets totalling about $14$k characters. We reproduce one short bullet from each section.

\begin{tcolorbox}[memorybox, title={ACE on WebShop, stage 100 (32 bullets, $\sim$14k chars) --- representative items}, label={box:ace-healthy}]
\textbf{Section: Common mistakes to avoid.} Do not invent a reasoning error when the predicted action and ground truth are the same and the environment confirms success. In that case, the correct diagnosis is that no substantive error occurred.

\smallskip
\textbf{Section: Problem-solving heuristics.} For apparel tasks that specify both fit type and size, treat them as a linked final gate: confirm the requested fit type (e.g., youth) and the exact size (e.g., x-small) on the product page before purchase, along with color and price, rather than validating each in isolation.
\end{tcolorbox}

\paragraph{Dynamic Cheatsheet healthy.} Dynamic Cheatsheet keeps a single living document and rewrites it after every trajectory. The document at stage~$100$ of the same run is roughly $8.3$k characters, organised under a \textsc{Action patterns and interaction sequences} heading with one \texttt{<memory\_item>} per consolidated insight (each with \texttt{<description>}, \texttt{<example>}, and a \texttt{Count} field). We reproduce two consecutive \texttt{<memory\_item>} blocks.

\begin{tcolorbox}[memorybox, title={Dynamic Cheatsheet on WebShop, stage 100 ($\sim$8.3k chars) --- two-item blocks, verbatim}, label={box:dc-healthy}]
\texttt{<memory\_item>}\\
\texttt{<description>}\\
For apparel tasks with multiple exact requirements, search broad by category plus the most discriminative attributes first, then verify the remaining constraints on the product page. Do not trust title-only matches for fabric, fit, care, color, size, or price. (Reference: current input)\\
\texttt{</description>}\\
\texttt{<example>}\\
Template:\\
1) Search: men's dress shirt cotton spandex classic fit short sleeve machine wash melon berry 6x under 60\\
2) Open best-looking product\\
3) Verify category = men's dress shirt\\
4) Verify fabric = cotton spandex\\
5) Verify fit = classic fit\\
6) Verify sleeve = short sleeve\\
7) Verify care = machine wash\\
8) Verify color = melon berry\\
9) Verify size = 6x\\
10) Verify price $<$ 60.00\\
11) Buy only after all checks pass\\
\texttt{</example>}\\
\texttt{** Count: 1}\\
\texttt{</memory\_item>}

\smallskip
\texttt{<memory\_item>}\\
\texttt{<description>}\\
Exact color variants are non-substitutable. If the task requests a precise color name like ``melon berry,'' do not accept nearby pink/red/coral shades, generic berry tones, or other ``melon'' variants unless the swatch label matches exactly. (Reference: current input)\\
\texttt{</description>}\\
\texttt{<example>}\\
Accept only:\\
- melon berry\\
\\
Reject nearby variants:\\
- berry\\
- melon\\
- rose\\
- coral\\
- salmon\\
- watermelon\\
- dusty rose\\
- light red\\
\texttt{</example>}\\
\texttt{** Count: 1}\\
\texttt{</memory\_item>}
\end{tcolorbox}

These two snapshots illustrate the methods' design intent: ACE produces structured bullet lists with explicit failure-mode categories, while Dynamic Cheatsheet produces one continuously-edited document with embedded action templates and counter-examples. Both show concrete domain content (apparel, sizes, colour names, price thresholds) and avoid the empty-content / paraphrase failure modes of \cref{sec:case-useless,sec:case-erosion}.

\subsubsection{Collapsed memory: Agent Workflow Memory}
\label{sec:webshop-awm-collapse}

\paragraph{Setup.} The artifact is the AWM-style consolidated workflow file at \emph{epoch~$3$} of the same training run that supplies the epoch-$1$ snapshot used in the W8 ablation of \cref{sec:surgery-awm-w8}: $128$ golden WebShop trajectories ($n{=}128$, all reward $=1.0$), gpt-5.4-mini consolidator, batch size $16$, 3 epochs. Epoch~$1$ contains $8$ workflows (W1--W8 in \cref{box:awm-memory}); epoch~$3$ contains $16$ workflows (W1--W16 below). The first $8$ workflows of epoch~$3$ share titles and substantive content with epoch~$1$ (W2--W6 and W8 are character-identical modulo trailing whitespace; W1 and W7 receive minor wording edits over the next two epochs). The additional W9--W16 are new entries the consolidator appended over those epochs.

\paragraph{The collapsed snapshot.} We list epoch~$3$'s $16$ workflow titles below. The eight W9--W16 entries (highlighted) are category-specific specialisations of W1--W7; none introduces a new control-flow pattern or stop criterion not already present in W1--W7. We treat this as collapsed by accretion: the cap of the consolidation pipeline does not bind here, but the consolidator nevertheless stores redundant restatements of the same templates.

\begin{tcolorbox}[memorybox, title={AWM on WebShop, epoch 3 (16 workflows, $\sim$8.2k chars) --- workflow titles, verbatim}, label={box:awm-collapse}]
\textbf{W1.} Search by attribute-rich query. \\
\textbf{W2.} Open candidate item to inspect options. \\
\textbf{W3.} Select required attributes before buying. \\
\textbf{W4.} Select required size, color, and other variant options before buying. \\
\textbf{W5.} Search and select clothing variants with fit type. \\
\textbf{W6.} Search and select home decor variants. \\
\textbf{W7.} Search and select multi-part apparel sizing variants. \\
\textbf{W8.} Search across pages when the first results do not match. \\
\smallskip
\textbf{W9.} \annot{1}{Select apparel color, size, and fit/order-specific variants before buying.} \\
\textbf{W10.} \annot{2}{Select non-apparel flavor and size variants before buying.} \\
\textbf{W11.} \annot{2}{Select pack-count and color variants before buying.} \\
\textbf{W12.} \annot{2}{Search and select color, size, and shape variants for home goods.} \\
\textbf{W13.} \annot{2}{Search and select shoes with color and size.} \\
\textbf{W14.} \annot{2}{Search and select apparel color, size, and purchase.} \\
\textbf{W15.} \annot{2}{Search and select electronics memory/storage variants.} \\
\textbf{W16.} \annot{2}{Search and select activewear or performance tops variants.}
\end{tcolorbox}

\begin{annotations}[Collapse modes]
\annotitem{1}{Re-derivation of an existing workflow}{W9 ``Select apparel color, size, and fit/order-specific variants before buying'' is the union of W3 (``Select required attributes before buying''), W4 (``size, color, and other variant options''), and W5 (clothing fit type) restated in one entry. The action sequence inside W9 reuses the same primitives (\texttt{click[<required\_color>]}, \texttt{click[<required\_size>]}, etc.) as those three predecessors; no new primitive or guard is added.}
\annotitem{2}{Per-category specialisations of W1--W4}{W10--W16 each pin the search-then-variant-select template (W1--W4) to one product category (food/flavour, pack-count, home goods, shoes, apparel, electronics, activewear). Their action sequences differ only in which placeholder names appear inside \texttt{search[\dots]}; the control flow is identical to W2 followed by W3/W4. The eight new entries shorten the retrieval bandwidth available for the eight original abstract workflows without adding any control-flow generality. The W8 page-turn pathology of \cref{sec:surgery-awm-w8} persists unchanged in epoch~$3$.}
\end{annotations}
\section{Memory surgery}
\label{sec:memory-surgery}

We isolate the contribution of individual memory items to overall failure by removing one item from a memory snapshot and re-evaluating, holding the rollout policy and all other items fixed.

\subsection{Workflow~8 ablation on WebShop}
\label{sec:surgery-awm-w8}

\paragraph{Setup.} The memory under study is a consolidated workflow file (8 workflows W1--W8, reproduced verbatim in \cref{box:awm-memory}) produced after one epoch of AWM-style consolidation. AWM-style consolidation abstracts each input trajectory into a workflow template that pairs a thought rationale with an action sequence; we ablate Workflow~8 (W8) of the resulting file. The training data are 128 expert (task reward $1.0$) trajectories sampled from the AgentGym AgentTraj-L WebShop corpus (seed $42$). The eval split is a 50-session subset of WebShop's test pool (seed $42$) with overlap to the training sessions removed. We construct two evaluation conditions on this split:
\begin{itemize}
    \item \textbf{Full memory}: all 8 workflows.
    \item \textbf{Memory minus W8}: the 7 workflows from the same file with Workflow~8 removed (no other edits).
\end{itemize}
All conditions use \texttt{eval\_max\_steps}=30, \texttt{eval\_max\_tokens}=4096, and \texttt{eval\_num\_runs}=1. We do not set a temperature parameter. Each condition is evaluated under two rollout policies, gpt-5.4-mini and gpt-5-mini. The memory artifact is byte-identical across the two rollouts.

The WebShop action space used in \cref{tab:surgery-awm-w8-behavior} groups raw \texttt{click[\dots]} actions into six categories by the type of UI element clicked:
\begin{itemize}\itemsep=2pt
    \item \textbf{search}: issue a query against the catalog, e.g., \texttt{search[red running shoes size 10 < \$60]}. Returns a paginated list of product cards.
    \item \textbf{click ASIN}: open a product detail page from a search-result card by its ASIN (Amazon Standard Identification Number, a 10-character product code such as \texttt{b09npml43m}), e.g., \texttt{click[b09npml43m]}. This is the only way to inspect a product's variants, price, attributes, and Buy~Now button.
    \item \textbf{click variant}: select a required option on a product page before purchase, e.g., \texttt{click[medium]}, \texttt{click[b16-red]}. Many products require one or more such selections; missing a required variant prevents reward~1.0 even if Buy~Now succeeds.
    \item \textbf{click [Next >]}: advance to the next page of search results without opening any product, e.g., \texttt{click[Next >]}, \texttt{click[next \&gt;]}. We also count the symmetric paging primitives \texttt{click[< Prev]} here.
    \item \textbf{click back}: return to the search input from a results or product page, e.g., \texttt{click[Back to Search]}, abandoning the current query.
    \item \textbf{click [Buy Now]}: commit to purchasing the currently configured product. Terminates the episode and yields the final reward in $[0,1]$ based on attribute match against the instruction.
\end{itemize}

\paragraph{The consolidated memory.} The full saved memory file contains 8 workflows, W1--W8, reproduced verbatim below. The original file delimits each block with XML-style \texttt{<think>} and \texttt{<action>} tags; we render them as italic ``Thought''/``Action sequence'' labels for readability and otherwise preserve the text. Highlighted spans in W8 key into the paired \texttt{annotations} callout that follows.

\begin{tcolorbox}[memorybox, title={Consolidated memory file: 8 workflows W1--W8 (verbatim)}, label={box:awm-memory}]
\textbf{Workflow 1: Search by attribute-rich query}\\
\textit{Thought.} Build a single search query that combines the headline category words plus the most distinguishing attributes, variant-like descriptors, and the price cap. Use the instruction's key terms so WebShop can match the product listing by keyword overlap.\\
\textit{Action sequence.} \texttt{search[<headline\_category> <key\_attribute\_1> <key\_attribute\_2> \ldots\ <required\_color> <required\_size> price < <price\_cap>]}

\smallskip
\textbf{Workflow 2: Open candidate item to inspect options}\\
\textit{Thought.} After a search, click the \anchor{best-matching ASIN-like result} to open the product detail page where color/size pickers or other variants are exposed.\\
\textit{Action sequence.} \texttt{click[<candidate\_item\_id>]}

\smallskip
\textbf{Workflow 3: Select required attributes before buying}\\
\textit{Thought.} On the product detail page, click each required variant option from the instruction, such as color and size, in any order before purchasing.\\
\textit{Action sequence.}
\begin{flushleft}
\texttt{click[<required\_color>]};\quad
\texttt{click[<required\_size>]};\quad
\texttt{click[Buy Now]}.
\end{flushleft}

\smallskip
\textbf{Workflow 4: Select required size, color, and other variant options before buying}\\
\textit{Thought.} Some products expose multiple selectable variants beyond color and size, such as fit type, style, pack count, flavor, or another categorical option embedded in the product page. Click every required option from the instruction before buying.\\
\textit{Action sequence.}
\begin{flushleft}
\texttt{click[<required\_variant\_1>]};\quad
\texttt{click[<required\_variant\_2>]};\quad
\texttt{click[<required\_variant\_3>]};\quad
\texttt{click[Buy Now]}.
\end{flushleft}

\smallskip
\textbf{Workflow 5: Search and select clothing variants with fit type}\\
\textit{Thought.} For apparel that specifies fit type along with color and size, search with the apparel category and all major descriptors, then \anchor{open the best result} and select fit type, color, size, and buy.\\
\textit{Action sequence.}
\begin{flushleft}
\texttt{search[<apparel\_category> <key\_descriptors> <required\_color> <required\_fit\_type> <required\_size> price < <price\_cap>]};\quad
\texttt{click[<candidate\_item\_id>]};\quad
\texttt{click[<required\_fit\_type>]};\quad
\texttt{click[<required\_color>]};\quad
\texttt{click[<required\_size>]};\quad
\texttt{click[Buy Now]}.
\end{flushleft}

\smallskip
\textbf{Workflow 6: Search and select home decor variants}\\
\textit{Thought.} For home decor items such as window coverings or lighting, search using the room/use-case keywords plus the product descriptors and price cap, then \anchor{open the best result} and select the requested size or color before buying.\\
\textit{Action sequence.}
\begin{flushleft}
\texttt{search[<home\_decor\_category> <key\_descriptors> <room\_or\_use\_case> <required\_color> <required\_size> price < <price\_cap>]};\quad
\texttt{click[<candidate\_item\_id>]};\quad
\texttt{click[<required\_size>]};\quad
\texttt{click[<required\_color>]};\quad
\texttt{click[Buy Now]}.
\end{flushleft}

\smallskip
\textbf{Workflow 7: Search and select multi-part apparel sizing variants}\\
\textit{Thought.} Some apparel sizes include compound labels like tall, big, or tag-based sizes. Search with the product descriptors and the base size text, \anchor{open the item}, then choose the exact compound size and other required options before buying.\\
\textit{Action sequence.}
\begin{flushleft}
\texttt{search[<apparel\_category> <key\_descriptors> <required\_color> <base\_size\_text> price < <price\_cap>]};\quad
\texttt{click[<candidate\_item\_id>]};\quad
\texttt{click[<required\_color>]};\quad
\texttt{click[<required\_compound\_size>]};\quad
\texttt{click[Buy Now]}.
\end{flushleft}

\smallskip
\textbf{Workflow 8: Search across pages when the first results do not match}\\
\textit{Thought.} If the initial result page does not contain a plausible match, \annot{1}{continue to the next page and inspect additional candidates until a suitable item appears,} then open it and proceed with variant selection and purchase.\\
\textit{Action sequence.}
\begin{flushleft}
\annot{3}{\texttt{click[Next >]}};\quad
\texttt{click[}\annot{2}{\texttt{<candidate\_item\_id>}}\texttt{]};\quad
\texttt{click[<required\_variant\_1>]};\quad
\texttt{click[<required\_variant\_2>]};\quad
\texttt{click[Buy Now]}.
\end{flushleft}
\end{tcolorbox}

\begin{annotations}
\annotitem{1}{No stop criterion}{The natural-language description tells the agent to continue paging until a ``suitable item appears,'' but suitability is not defined relative to any concrete feature of the agent's task. The agent therefore has no signal for when to halt the loop.}
\annotitem{2}{Abstract candidate placeholder}{In W2 and W5--W7 of \cref{box:awm-memory}, \texttt{click[<candidate\_item\_id>]} is preceded by an ``open the best-matching result'' step (highlighted in green in those workflows) that picks a concrete ASIN. W8 omits this selection step, so the placeholder is never bound to a real product.}
\annotitem{3}{Page-turn primitive promoted to a workflow}{\texttt{click[Next >]} is a low-level navigation action. It appears nowhere in the action sequences of W1--W7. W8 elevates it to a first-class workflow, signalling to the agent that turning pages without committing is a sanctioned strategy.}
\end{annotations}

\paragraph{Result.} \cref{tab:surgery-awm-w8} reports per-condition outcomes; \cref{tab:surgery-awm-w8-behavior} breaks down the action stream by category.

\begin{table}[h]
\centering
\caption{Removing W8 yields higher wins, higher mean reward, and lower mean steps in both rollouts; even so, gpt-5.4-mini reaches only $14/50$ wins after removal, so the bulk of failures under this rollout is not attributable to W8 alone. Under gpt-5.4-mini no episode reaches the 30-step cap in either condition; the cap reduction is concentrated in gpt-5-mini ($22\to17$). Per-condition outcomes for the two rollout policies, n=50 episodes per cell, single run. Columns: episodes solved out of 50; mean per-episode reward in $[0,1]$; mean trajectory length in steps; episodes that hit the max steps (which is 30).}
\label{tab:surgery-awm-w8}
\begin{tabular}{llcccc}
\toprule
Rollout & Memory & wins/50 & mean reward & mean steps & hit step cap \\
\midrule
gpt-5.4-mini & full (8 workflows)     & 7/50  & 0.23 & 12.4 & 0/50  \\
gpt-5.4-mini & minus W8 (7 workflows) & 14/50 & 0.37 & 11.4 & 0/50  \\
\midrule
gpt-5-mini   & full (8 workflows)     & 18/50 & 0.49 & 18.2 & 22/50 \\
gpt-5-mini   & minus W8 (7 workflows) & 23/50 & 0.59 & 15.8 & 17/50 \\
\bottomrule
\end{tabular}
\end{table}

\paragraph{Behavioral evidence.} With W8 in context, \texttt{click[Next >]} dominates the action stream relative to product-page interactions and \texttt{click[Buy Now]}. Removing W8 reduces this concentration in both rollouts (n=50 episodes per cell, single run; no significance test). Under gpt-5.4-mini, observed page-turn counts are $421$ (full) and $181$ (minus W8), a ratio of $2.3\times$; the count of distinct episodes that issue at least one \texttt{click[Buy Now]} is $14/50$ (full) and $21/50$ (minus W8). Under gpt-5-mini, page-turn counts are $559$ and $246$ (also $2.3\times$); the Buy-Now-episode count is $28/50$ and $33/50$.

\begin{table}[h]
\centering
\caption{With W8 in context, \texttt{click[Next >]} dominates the action stream relative to product-page interactions and \texttt{click[Buy Now]}; removing W8 reduces this concentration in both rollouts (page-turn ratio $2.3\times$ full/minus-W8 in both). Action-category counts over n=50 episodes per cell, single run. Action categories are defined in the bullet list of \cref{sec:surgery-awm-w8}. ``Buy Now (eps)'' is the number of distinct episodes that issued at least one \texttt{click[Buy Now]}.}
\label{tab:surgery-awm-w8-behavior}
\begin{tabular}{lcccc}
\toprule
 & \multicolumn{2}{c}{gpt-5.4-mini rollout} & \multicolumn{2}{c}{gpt-5-mini rollout} \\
\cmidrule(lr){2-3} \cmidrule(lr){4-5}
Action & Full memory & Minus W8 & Full memory & Minus W8 \\
\midrule
search             & 81    & 142   & 117  & 183 \\
click ASIN         & 42    & 72    & 70   & 97  \\
click variant      & 28    & 55    & 71   & 99  \\
click [Next >]     & 421   & 181   & 559  & 246 \\
click back         & 33    & 97    & 67   & 133 \\
click [Buy Now]    & 14    & 21    & 28   & 33  \\
Buy Now (eps)      & 14    & 21    & 28   & 33  \\
\bottomrule
\end{tabular}
\end{table}

\section{ScienceWorld}
\label{sec:scienceworld}

\subsection{Abnormal memories}
\label{sec:scienceworld-abnormal}

\subsubsection{Representative cases}

We first present three representative abnormal memory cases identified in the ScienceWorld environment, one from each abnormality category. These cases illustrate qualitatively different failure modes in memory consolidation: \texttt{over\_generalization}, where a memory abstracts from prior experience too broadly and may bias the agent toward inappropriate actions; \texttt{over\_specialization}, where a memory is tied to overly specific entities or procedures and therefore has limited transferability; and \texttt{useless\_garbage}, where a memory is vague, malformed, or insufficiently actionable. For each case, we annotate the problematic components of the memory and explain why retrieving it may hurt downstream decision-making.

\begin{tcolorbox}[memorybox, title={Representative \texttt{over\_generalization} memory from ScienceWorld}]
\annot{1}{Using a lighter, fire source, or oven MAY BE NECESSARY} to \annot{2}{change the state of a food or substance} in \annot{3}{state-change tasks}.
\end{tcolorbox}

\begin{annotations}
  \annotitem{1}{Overgeneralized tool prior}{The memory promotes a broad prior over heat sources, even though many state-change tasks may require cooling, freezing, melting, or task-specific apparatus rather than a lighter, fire source, or oven.}
  \annotitem{2}{Underspecified target transformation}{``Change the state'' collapses different physical transformations into one generic goal, ignoring whether the required transition is boiling, freezing, melting, or another state change.}
  \annotitem{3}{Cross-task overreach}{The rule is stated for all state-change tasks, so retrieval may bias the agent toward heat-related actions even in contexts where heating is irrelevant or harmful.}
\end{annotations}

\begin{tcolorbox}[memorybox, title={Representative \texttt{over\_specialization} memory from ScienceWorld}]
For conductivity-sorting tasks, \annot{1}{picking up and testing unknown substance B} in a circuit SHOULD BE NECCESSARY before \annot{2}{placing it in the correct box}.
\end{tcolorbox}

\begin{annotations}
  \annotitem{1}{Instance-specific binding}{The memory is tied to ``unknown substance B'' rather than the general class of unknown substances, so it may fail to transfer to otherwise identical tasks involving different item names.}
  \annotitem{2}{Narrow task schema}{The final action is framed around a specific sorting protocol, making the memory useful only for a narrow subset of conductivity tasks rather than a robust reusable strategy.}
\end{annotations}

\begin{tcolorbox}[memorybox, title={Representative \texttt{useless\_garbage} memory from ScienceWorld}]
\annot{1}{Using the correct container or setup} MAY BE CONTRIBUTE to \annot{2}{boiling the target substance}.
\end{tcolorbox}

\begin{annotations}
  \annotitem{1}{Vague object reference}{``Correct container or setup'' does not specify which observable object should be selected, moved, opened, filled, or heated, so it provides little actionable guidance.}
  \annotitem{2}{Generic task hint}{The memory merely restates that some appropriate setup may help with boiling, but does not identify a concrete precondition, action sequence, or failure mode.}
\end{annotations}

\subsubsection{Additional examples}

We further list additional abnormal memories selected by the LLM-as-judge from ScienceWorld memory banks. These examples are grouped into the same three categories: \texttt{over\_generalization}, \texttt{over\_specialization}, and \texttt{useless\_garbage}. 

\begin{tcblisting}{
  promptbox,
  title={Abnormal Memories: \texttt{over\_generalization}},
  listing only,
  listing options={
    basicstyle=\ttfamily\scriptsize,
    breaklines=true,
    columns=fullflexible,
    keepspaces=true,
    showstringspaces=false,
    escapeinside={(*@}{@*)}
  }
}
(*@\textbf{[over\_generalization \#1]}@*)
Memory: Observing all life stages in order MAY BE NECESSARY to complete a turtle-stage task.

(*@\textbf{[over\_generalization \#2]}@*)
Memory: Observing all life stages in order MAY BE NECESSARY to complete a moth-stage task.

(*@\textbf{[over\_generalization \#3]}@*)
Memory: Using a lighter or fire source MAY BE NECESSARY to change the state of matter of water in outdoor state-change tasks.

(*@\textbf{[over\_generalization \#4]}@*)
Memory: Using a lighter, fire source, or oven MAY BE NECESSARY to change the state of a food or substance in state-change tasks.

(*@\textbf{[over\_generalization \#5]}@*)
Memory: Focusing on the target substance SHOULD BE NECCESSARY to change its state.
\end{tcblisting}

\begin{tcblisting}{
  promptbox,
  title={Abnormal Memories: \texttt{over\_specialization}},
  listing only,
  listing options={
    basicstyle=\ttfamily\scriptsize,
    breaklines=true,
    columns=fullflexible,
    keepspaces=true,
    showstringspaces=false,
    escapeinside={(*@}{@*)}
  }
}
(*@\textbf{[over\_specialization \#1]}@*)
Memory: For conductivity-sorting tasks, picking up and testing unknown substance B in a circuit SHOULD BE NECCESSARY before placing it in the correct box.

(*@\textbf{[over\_specialization \#2]}@*)
Memory: MAY CONTRIBUTE to anchoring progress checks to observable context (current room, visible objects) before initiating life-stage sequencing in melting-point tasks, where initial observations (thermometer, boxes) must be secured before progress.

(*@\textbf{[over\_specialization \#3]}@*)
Memory: For salt water recipes, placing salt and a cup/container in the sink MAY BE NECESSARY before pouring water.

(*@\textbf{[over\_specialization \#4]}@*)
Memory: For sodium acetate recipes, moving the required ingredients into a glass cup MAY BE NECESSARY before focusing on the product.

(*@\textbf{[over\_specialization \#5]}@*)
Memory: SEED_ACQUISITION_BARRIER: Access to viable apple seeds often hinges on physically locating seeds in connected environments (e.g., kitchen) and then transporting them to an appropriate growing area.
\end{tcblisting}

\begin{tcblisting}{
  promptbox,
  title={Abnormal Memories: \texttt{useless\_garbage}},
  listing only,
  listing options={
    basicstyle=\ttfamily\scriptsize,
    breaklines=true,
    columns=fullflexible,
    keepspaces=true,
    showstringspaces=false,
    escapeinside={(*@}{@*)}
  }
}
(*@\textbf{[useless\_garbage \#1]}@*)
Memory: Assembling an electrical circuit DOES NOT CONTRIBUTE to completing a smores synthesis task.

(*@\textbf{[useless\_garbage \#2]}@*)
Memory: Mixing the correct ingredients MAY BE NECESSARY to create a target solution or mixture.

(*@\textbf{[useless\_garbage \#3]}@*)
Memory: Mixing ingredients in a glass cup MAY BE NECESSARY to create smores after reading the recipe.

(*@\textbf{[useless\_garbage \#4]}@*)
Memory: Applying the correct cultivation action after planting a seed MAY BE NECESSARY to advance plant growth tasks.

(*@\textbf{[useless\_garbage \#5]}@*)
Memory: Using the correct container or setup MAY BE CONTRIBUTE to boiling the target substance.

(*@\textbf{[useless\_garbage \#6]}@*)
Memory: Moving the target container to a working oven MAY BE CONTRIBUTE to boiling the target substance.

(*@\textbf{[useless\_garbage \#7]}@*)
Memory: Repeated heating actions MAY BE CONTRIBUTE to gradually raising the temperature of the target substance.

(*@\textbf{[useless\_garbage \#8]}@*)
Memory: Using a lighter MAY BE CONTRIBUTE to boiling a substance when combustion is allowed.

(*@\textbf{[useless\_garbage \#9]}@*)
Memory: Opening and using an oven MAY BE CONTRIBUTE to boiling or heating a target substance.

(*@\textbf{[useless\_garbage \#10]}@*)
Memory: Waiting after activating a heat source MAY BE CONTRIBUTE to changing the state of a target substance.

(*@\textbf{[useless\_garbage \#11]}@*)
Memory: Opening an oven, placing the target substance inside, and waiting MAY BE CONTRIBUTE to changing the state of matter of the target substance.

(*@\textbf{[useless\_garbage \#12]}@*)
Memory: Focusing on the target substance MAY BE NECCESSARY to change its state.

(*@\textbf{[useless\_garbage \#13]}@*)
Memory: Focusing on the wrong substance DOES NOT CONTRIBUTE to changing the state of the target substance.
\end{tcblisting}

\subsubsection{Memory banks across consolidation rounds}

We further examine the evolution of memory banks produced by GPT-5.4-mini in the ScienceWorld environment. The model performs 20 rounds of memory consolidation over 485 rollouts. We show three representative memory banks: \textbf{the first round}, which reflects the initial memory bank; \textbf{the tenth round}, which reflects an intermediate memory bank; and \textbf{the twentieth round}, which reflects the final memory bank. For each memory bank, we also report the corresponding average test score.

\begin{tcblisting}{
  promptbox,
  title={Initial memory bank: round 1, Memory Bank Size = 16, average test score = 7.45},
  listing only,
  listing options={
  basicstyle=\ttfamily\scriptsize,
  breaklines=true,
  breakautoindent=false,
  breakindent=0pt,
  breakatwhitespace=true,
  columns=fullflexible,
  keepspaces=true,
  showstringspaces=false,
  escapeinside={(*@}{@*)}
}
}

(*@\textbf{[memory \#1]}@*)
Life-stage identification tasks MAY BE NECESSARY to search the relevant rooms for the specified organism stages in order; wandering through unrelated rooms and inspecting irrelevant objects DOES NOT CONTRIBUTE to success.

(*@\textbf{[memory \#2]}@*)
Sandwich-making tasks MAY BE NECESSARY to read the recipe, gather the listed ingredients, and mix them in the correct container; opening empty drawers or collecting unrelated items DOES NOT CONTRIBUTE to success.

(*@\textbf{[memory \#3]}@*)
Conductivity tasks MAY BE NECESSARY to build the circuit, observe the switch result, and then place the substance in the box matching conductivity; repeated reconnect attempts on already-connected parts DOES NOT CONTRIBUTE to success.

(*@\textbf{[memory \#4]}@*)
Plant-growth tasks MAY BE NECESSARY to place the seed in soil or a pot, water it if required, and then wait for growth; pouring seeds into the agent or moving them between containers DOES NOT CONTRIBUTE to success.

(*@\textbf{[memory \#5]}@*)
State-change tasks MAY BE NECESSARY to use the task-specific tool or heat/cool source on the target substance and wait; interacting with unrelated objects or empty containers DOES NOT CONTRIBUTE to success.

(*@\textbf{[memory \#6]}@*)
Melting-point tasks MAY BE NECESSARY to measure the substance with a thermometer before choosing the box; focusing on the wrong box after a valid reading DOES NOT CONTRIBUTE to success.

(*@\textbf{[memory \#7]}@*)
Freezing tasks MAY BE NECESSARY to put the target substance in a freezer or otherwise cool it, then wait for the state change; wandering without locating the substance DOES NOT CONTRIBUTE to success.

(*@\textbf{[memory \#8]}@*)
Paint-mixing tasks MAY BE NECESSARY to combine the required colored paints in the correct container; repeatedly pouring a single color or empty containers DOES NOT CONTRIBUTE to success.

(*@\textbf{[memory \#9]}@*)
Life-stage identification tasks MAY BE NECESSARY to search the relevant rooms for the specified organism stages in order; wandering through unrelated rooms and inspecting irrelevant objects DOES NOT CONTRIBUTE to success.

(*@\textbf{[memory \#10]}@*)
Sandwich-making tasks MAY BE NECESSARY to read the recipe, gather the listed ingredients, and mix them in the correct container; opening empty drawers or collecting unrelated items DOES NOT CONTRIBUTE to success.

(*@\textbf{[memory \#11]}@*)
Some conductivity tasks MAY BE NECESSARY to use the correct circuit components and then sort the substance into the box matching the observed conductivity; testing with unrelated objects or looping on already-connected terminals DOES NOT CONTRIBUTE to success.

(*@\textbf{[memory \#12]}@*)
Melting-point tasks MAY BE NECESSARY to measure the target substance before choosing the box; focusing on the wrong box after a valid reading DOES NOT CONTRIBUTE to success.

(*@\textbf{[memory \#13]}@*)
Boiling/combustion tasks MAY BE NECESSARY to use the stove if it works, but if the stove is broken a lighter MAY BE NECESSARY to ignite the target substance.

(*@\textbf{[memory \#14]}@*)
Motor-power tasks MAY BE NECESSARY to connect the generator, wires, and motor correctly before activating the power source.

(*@\textbf{[memory \#15]}@*)
Ice-cream heating tasks MAY BE NECESSARY to move the target into an oven when the stove is broken, then wait for completion.

(*@\textbf{[memory \#16]}@*)
Dough-making tasks MAY BE NECESSARY to combine flour and water; pouring unrelated items or empty containers DOES NOT CONTRIBUTE to success.
\end{tcblisting}

\begin{tcblisting}{
  promptbox,
  title={Intermediate memory bank: round 10, Memory Bank Size = 10, average test score = -7.68},
  listing only,
  listing options={
  basicstyle=\ttfamily\scriptsize,
  breaklines=true,
  breakautoindent=false,
  breakindent=0pt,
  breakatwhitespace=true,
  columns=fullflexible,
  keepspaces=true,
  showstringspaces=false,
  escapeinside={(*@}{@*)}
}
}

(*@\textbf{[memory \#1]}@*)
Recipe-based mixing tasks MAY BE NECESSARY to read the recipe first, gather the exact listed ingredients, and mix them in the intended container; unrelated pouring, extra items, or mixing in the wrong place DOES NOT CONTRIBUTE to success.

(*@\textbf{[memory \#2]}@*)
Plant-growth tasks MAY BE NECESSARY to keep the seed in the greenhouse or other growth area with soil or a pot and water while waiting; moving seeds through cups, jars, toilets, or the agent DOES NOT CONTRIBUTE to growth.

(*@\textbf{[memory \#3]}@*)
Temperature-sorting tasks MAY BE NECESSARY to locate both the thermometer and the target substance, measure first, and only then choose the correct box; selecting a box before obtaining a reading or wandering without measuring the target DOES NOT CONTRIBUTE to success.

(*@\textbf{[memory \#4]}@*)
State-change tasks MAY BE NECESSARY to place the target substance in the correct heating or freezing setup, then wait long enough and verify the transformed state; skipping the wait or focusing on unrelated objects DOES NOT CONTRIBUTE to completion.

(*@\textbf{[memory \#5]}@*)
Conductivity tasks MAY BE NECESSARY to assemble the circuit with the target object, power source, wires, switch, and indicator before sorting it into the box matching the test result; repeated attempts to connect already-occupied terminals or other connection loops DOES NOT CONTRIBUTE to success.

(*@\textbf{[memory \#6]}@*)
Recipe-based mixing tasks MAY BE NECESSARY to read the recipe first, gather the exact listed ingredients, and mix them in the intended container; unrelated pouring, extra items, or mixing in the wrong place DOES NOT CONTRIBUTE to success.

(*@\textbf{[memory \#7]}@*)
Plant-growth tasks MAY BE NECESSARY to keep the seed in the greenhouse or other growth area with soil or a pot and water while waiting; moving seeds through cups, jars, toilets, or the agent DOES NOT CONTRIBUTE to growth.

(*@\textbf{[memory \#8]}@*)
Conductivity tasks MAY BE NECESSARY to assemble the circuit with the target object, battery, wires, switch, and light bulb, activate the switch, and then place the object into the box matching the observed result; the green box can be the correct destination for a conductive result.

(*@\textbf{[memory \#9]}@*)
Temperature-sorting tasks MAY BE NECESSARY to retrieve the thermometer from one room, measure the target substance in another room, and then choose the box based on the threshold; the bathroom can be the final placement room after measuring in the living room.

(*@\textbf{[memory \#10]}@*)
State-change tasks MAY BE NECESSARY to heat the target in a stove or pot setup, wait through multiple turns, and only then decide the box; starting the heating setup without completing the final reading leaves the task incomplete.
\end{tcblisting}

\begin{tcblisting}{
  promptbox,
  title={Final memory bank: round 20, Memory Bank Size = 17, average test score = -8.66},
  listing only,
  listing options={
  basicstyle=\ttfamily\scriptsize,
  breaklines=true,
  breakautoindent=false,
  breakindent=0pt,
  breakatwhitespace=true,
  columns=fullflexible,
  keepspaces=true,
  showstringspaces=false,
  escapeinside={(*@}{@*)}
}
}

(*@\textbf{[memory \#1]}@*)
Some life-stage tasks SHOULD BE NECESSARY to retrieve the seed, plant it in soil or a pot, then wait or water appropriately until the target stage appears; repeated self-pouring or unrelated interactions DOES NOT CONTRIBUTE to success.

(*@\textbf{[memory \#2]}@*)
Some recipe tasks SHOULD BE NECESSARY to read the recipe, gather the exact listed ingredients, combine them in the required container, and then focus on the created product to finish.

(*@\textbf{[memory \#3]}@*)
Some conductivity tasks SHOULD BE NECESSARY to assemble a clean circuit with battery, wires, switch, and bulb, observe the result, and place the substance in the matching box; repeated reconnecting or terminal-occupied errors DOES NOT CONTRIBUTE to success.

(*@\textbf{[memory \#4]}@*)
Some paint-mixing tasks MAY BE NECESSARY to use the exact color ingredients and avoid pouring empty cups or unrelated items into the agent.

(*@\textbf{[memory \#5]}@*)
Some conductivity tasks MAY BE NECESSARY to avoid repeated connection loops and terminal-occupied errors when building the circuit.

(*@\textbf{[memory \#6]}@*)
Some plant-growth tasks MAY BE NECESSARY to plant the seed first and then wait or water appropriately; repeatedly pouring the seed into the agent or moving it between inventory and hand DOES NOT CONTRIBUTE to success.

(*@\textbf{[memory \#7]}@*)
Some conductivity tasks MAY BE NECESSARY to avoid sorting the substance before the circuit test is complete; premature box placement DOES NOT CONTRIBUTE to success.

(*@\textbf{[memory \#8]}@*)
Some life-stage tasks MAY BE NECESSARY to ignore distractor animals or pictures and keep searching for the required species stages.

(*@\textbf{[memory \#9]}@*)
Some paint-mixing tasks MAY BE NECESSARY to combine the specified colors directly rather than repeatedly pouring a single color or empty containers.

(*@\textbf{[memory \#10]}@*)
Some state-change tasks MAY BE NECESSARY to use the correct heat or cold source on the target, wait, verify the final temperature or state, and then choose the matching box; skipping measurement or choosing the wrong box DOES NOT CONTRIBUTE to success.

(*@\textbf{[memory \#11]}@*)
Some state-change tasks MAY BE NECESSARY to use a lighter or oven when the stove is broken, especially for marshmallow, ice cream, or similar items.

(*@\textbf{[memory \#12]}@*)
Some freezing tasks MAY BE NECESSARY to place the target in an ultra-low temperature freezer, wait, and then retrieve it; removing it too early MAY BE CONTRIBUTE to incomplete progress.

(*@\textbf{[memory \#13]}@*)
Some temperature-box tasks MAY BE NECESSARY to measure the target first and then place it in the box dictated by the reading; failing to do the final placement leaves the task incomplete.

(*@\textbf{[memory \#14]}@*)
Some state-change tasks MAY BE NECESSARY to move the target onto a stove or pot, wait, then use the thermometer in the heated container before choosing the box.

(*@\textbf{[memory \#15]}@*)
Some state-change tasks MAY BE NECESSARY to heat ice cream in a cup on the stove when the task requires changing its state.

(*@\textbf{[memory \#16]}@*)
Some temperature-box tasks MAY BE NECESSARY to measure the target in the kitchen or living room, then carry it to the correct colored box in another room.

(*@\textbf{[memory \#17]}@*)
Some combustion tasks MAY BE NECESSARY to use a lighter directly on a marshmallow when boiling is not possible.
\end{tcblisting}

\section{Additional Analyses}
\label{app:additional}

\subsection{Auto-Mode Component Ablation}
\label{app:auto-components}
\begin{figure}[t]
\begin{minipage}{\linewidth}
\centering
\includegraphics[width=0.55\linewidth]{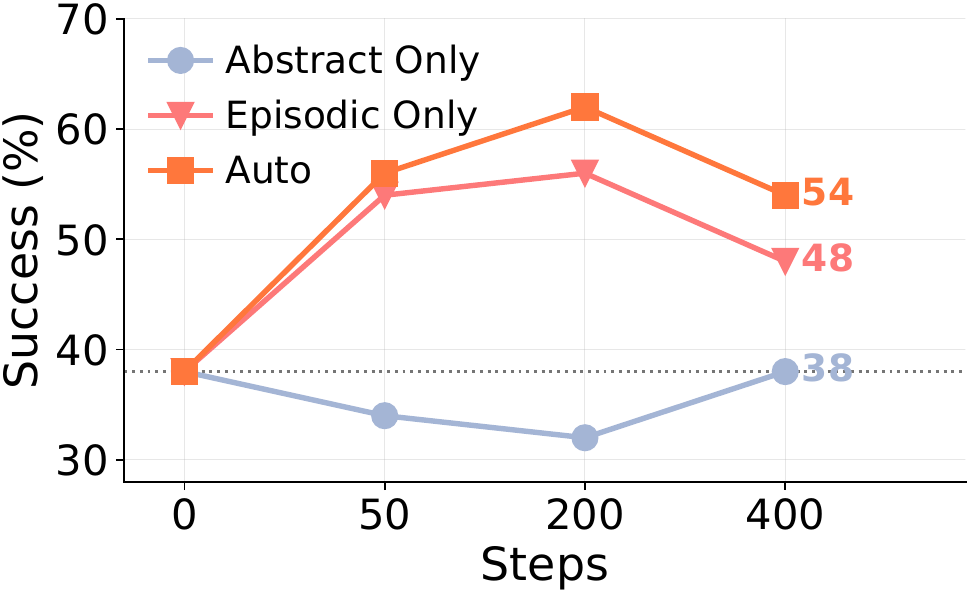}
\caption{\textbf{What carries the Auto-mode gain?} On the GT-400 stream, we
ablate which memory source conditions the solver at four checkpoints: \textit{Abstract Only}
uses only
distilled abstract entries, \textit{Episodic Only}
uses only the raw
episodic store, and \textit{Auto}
uses both. Episodic Only accounts for
nearly all of the gain; Abstract Only stays at or below the zero-shot baseline
(dotted) throughout; combining the two adds a few extra points at every
checkpoint but the late-step decline at $400$ tracks the abstract store's
gradual erosion.}
\label{fig:gt_400_components}
\end{minipage}
\end{figure}

The main-text Auto + Episodic curve in Fig.~\ref{fig:gt_400} couples two
memory sources. To isolate their contributions we re-run evaluation at four
checkpoints with each source restricted in turn: \textit{Abstract Only} reads
only the distilled abstract memory, \textit{Episodic Only} reads only the raw
episodic store, and \textit{Auto} reads both. The Episodic Only condition
recovers nearly all of the Auto gain; Abstract Only never improves on
zero-shot (Fig.~\ref{fig:gt_400_components}).

\subsection{Memory Composition under Task Switching}
\label{app:judge-cumcount}
\begin{figure}[t]
\centering
\includegraphics[width=0.75\linewidth]{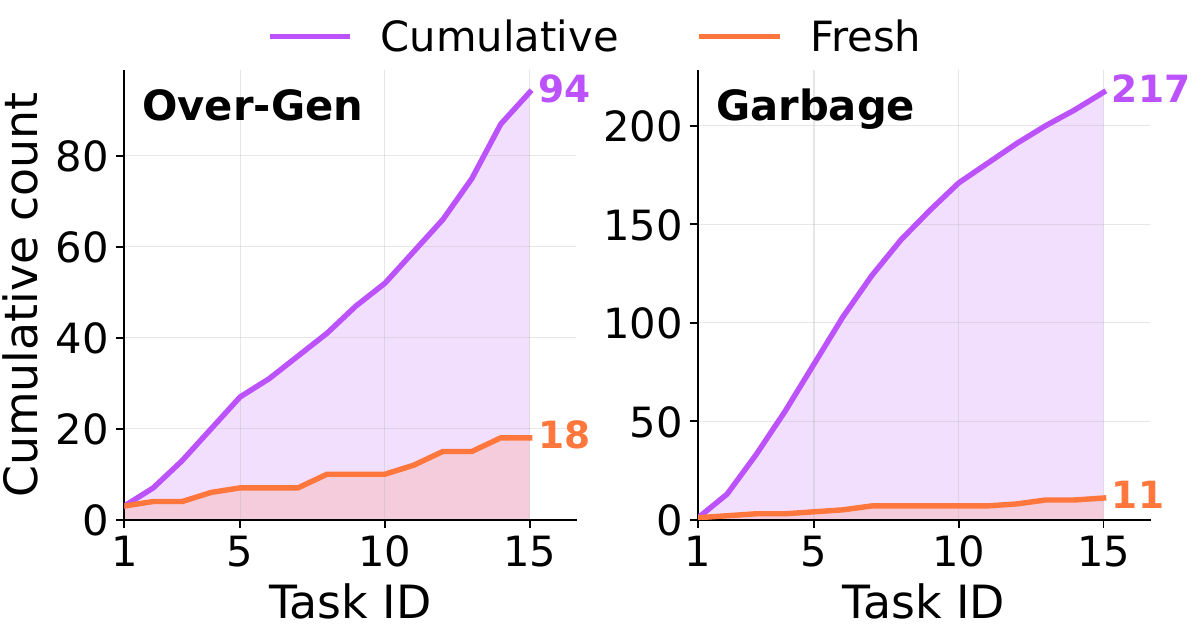}
\caption{\textbf{What accumulates in memory under task switching (GPT5.4-mini).} Running totals of LLM-judge--flagged over-generalized and garbage entries across the 15-task ScienceWorld switch sequence, comparing \emph{Cumulative} (consolidate over tasks $1\!\dots\!t$) against \emph{Fresh} (consolidate only on task $t$). Cumulative buries the store in over-gen and garbage memories that Fresh never accrues, with the gap widening monotonically over the sequence. Referenced from §\ref{sec:catastrophic-interference}.}
\label{fig:judge_cumcount}
\end{figure}

Supporting evidence for the Fresh-vs-Cumulative comparison in
\S\ref{sec:catastrophic-interference}. After each of the 15 ScienceWorld
switch tasks, we ask an LLM judge to label every entry currently in the
memory store as \emph{normal}, \emph{over-generalized}, \emph{over-specific},
or \emph{garbage}. Figure~\ref{fig:judge_cumcount} plots the running totals
of the two failure-mode categories (over-gen and garbage) across the
sequence for GPT5.4-mini. Cumulative's totals diverge from Fresh's almost
immediately and never recover, supplying a direct mechanistic account of the
$+203$-point score gap reported in the main text.

\section{Extended Experiments: Robustness, Attribution, and Generality}
\label{app:extended}

This appendix reports additional experiments that (i) attach variance estimates
to the headline results through repeated runs, (ii) isolate \emph{which} part of
the streamed-update procedure causes degradation, and (iii) test whether the
phenomenon survives stronger models, more memory frameworks, longer contexts,
and prompt optimization. Together they rule out the three most natural
alternative explanations for our central finding---bounded-memory forgetting,
context-window rot, and a single weak implementation---and leave the
consolidation step itself as the source of degradation.

\begin{tcolorbox}[takeawaysbox]
Across ten repeated end-to-end runs, seven published memory frameworks, four
frontier backbones, contexts up to ${\sim}790$K tokens, and an optimized-prompt
control, repeated textual consolidation still corrupts useful procedural
knowledge. The effect is not an artifact of sampling noise, memory capacity,
context length, or one abstraction pipeline.
\end{tcolorbox}

\subsection{Statistical Robustness under Repeated Runs}
\label{app:robustness}

We re-ran the main ARC-AGI Stream experiment (GPT-5.4) ten times end-to-end and
report mean $\pm$ standard deviation. The gap between raw-trajectory memory and
no-memory, and the gap between raw trajectories and streamed consolidation, both
exceed run-to-run variance (Table~\ref{tab:resp-main-ci}). The batch-size $\times$
update-mode grid (Table~\ref{tab:resp-bsz-mode}) shows the same non-monotonic
trajectory the main text reports, now with error bars.

\begin{table}[htbp]
\centering
\small
\begin{tabular}{lc}
\toprule
\textbf{Condition} & \textbf{Mean accuracy} $\pm$ \textbf{std} \\
\midrule
No memory & $38.6 \pm 3.6$ \\
Raw ground-truth-code trajectories & $76.6 \pm 3.1$ \\
Raw natural-language trajectories & $70.4 \pm 3.2$ \\
\bottomrule
\end{tabular}
\caption{Main conditions on ARC-AGI Stream (GPT-5.4), ten end-to-end runs.
Raw-trajectory memory is far above no-memory; neither raw condition degrades
over the stream.}
\label{tab:resp-main-ci}
\end{table}

\begin{table}[htbp]
\centering
\small
\begin{tabular}{llccc}
\toprule
\textbf{Batch size} & \textbf{Update mode} & \textbf{Step 100} & \textbf{Step 200} & \textbf{Step 400} \\
\midrule
$1$  & Force & $51.8 \pm 3.0$ & $41.0 \pm 4.7$ & $43.2 \pm 6.8$ \\
$1$  & Auto  & $51.4 \pm 3.1$ & $40.4 \pm 4.0$ & $49.0 \pm 3.4$ \\
$8$  & Force & $38.4 \pm 6.0$ & $52.6 \pm 3.7$ & $47.8 \pm 4.2$ \\
$8$  & Auto  & $35.6 \pm 2.3$ & $46.0 \pm 4.5$ & $39.8 \pm 4.2$ \\
$32$ & Force & $56.2 \pm 2.6$ & $50.4 \pm 3.7$ & $59.0 \pm 4.9$ \\
$32$ & Auto  & $58.6 \pm 6.0$ & $56.2 \pm 3.5$ & $55.4 \pm 4.4$ \\
\bottomrule
\end{tabular}
\caption{Accuracy by batch size and update mode on ARC-AGI Stream (GPT-5.4),
mean $\pm$ std over ten runs. Larger batches ($32$) consolidate more reliably
than small ones (see \Cref{app:delayed}).}
\label{tab:resp-bsz-mode}
\end{table}

\paragraph{The in-distribution regression is robust.}
Repeating the $19$-task regression (problems GPT-5.4 solves at $100\%$ without
memory; \S\ref{sec:phenomenon}) ten times reproduces the batched-versus-streamed
gap (Table~\ref{tab:resp-19task}). To check that the effect is not specific to
that $19$-task slice, we expanded to every task with at least $50\%$ no-memory
solvability across the ten runs, yielding $33$ tasks and excluding tasks already
at floor. With consolidated memory a third of these tasks are never solved at
epoch~1, rising to $40\%$ by epoch~10: degradation shows up as lost coverage
(pass@10), not only lower mean accuracy (Table~\ref{tab:resp-33task}).

\begin{table}[htbp]
\centering
\small
\begin{minipage}[t]{0.42\textwidth}
\centering
\begin{tabular}{lc}
\toprule
\textbf{Setting} & \textbf{Mean acc.\ $\pm$ std} \\
\midrule
Batched  & $70.0 \pm 7.8$ \\
Streamed & $55.8 \pm 7.1$ \\
\bottomrule
\end{tabular}
\caption{$19$-task regression ($100\%\!\to\!54\%$), ten runs, GPT-5.4.}
\label{tab:resp-19task}
\end{minipage}\hfill
\begin{minipage}[t]{0.55\textwidth}
\centering
\begin{tabular}{lcc}
\toprule
\textbf{Setting} & \textbf{pass@1 ($\pm$ std)} & \textbf{pass@10} \\
\midrule
No memory & $58.5 \pm 5.4$ & $100$ ($33/33$) \\
Epoch 1   & $55.1 \pm 4.9$ & $66.7$ ($22/33$) \\
Epoch 10  & $38.5 \pm 5.3$ & $60.6$ ($20/33$) \\
\bottomrule
\end{tabular}
\caption{Expanded $33$-task set (GPT-5.4). Coverage (pass@10) falls as
consolidation proceeds.}
\label{tab:resp-33task}
\end{minipage}
\end{table}

\paragraph{Benchmark tables with variance.}
Table~\ref{tab:resp-alfworld-ci} repeats the ALFWorld comparison of
Table~\ref{tab:strong_baseline_appworld} four times and reports mean $\pm$ std.
The ordering is unchanged: raw trajectory logs (and bounded subsets of them)
remain competitive with or ahead of the lesson-style consolidators.

\begin{table}[htbp]
\centering
\footnotesize
\setlength{\tabcolsep}{3pt}
\resizebox{\textwidth}{!}{%
\begin{tabular}{@{}lcccccccccc@{}}
\toprule
& & \multicolumn{6}{c}{\textbf{Trajectory Logs (Summarized)}} & & & \\
\cmidrule(lr){3-8}
\textbf{Backbone} & \textbf{NoMem} & \textbf{All} & \textbf{FS=64} & \textbf{FS=32} & \textbf{FS=16}
& \textbf{Incr.\ Keep} & \textbf{One-Pass Keep} & \textbf{ACE} & \textbf{AWM} & \textbf{DC} \\
\midrule
GPT-5.4-Mini & $51.56{\pm}6.67$ & $91.07{\pm}1.02$ & $88.99{\pm}3.12$ & $82.74{\pm}3.12$ & $76.79{\pm}4.07$ & $81.85{\pm}5.86$ & $82.74{\pm}2.32$ & $75.30{\pm}2.80$ & $69.05{\pm}5.30$ & $57.14{\pm}6.12$ \\
GPT-5-Mini   & $50.52{\pm}3.56$ & $81.85{\pm}2.61$ & $79.76{\pm}1.02$ & $74.70{\pm}2.23$ & $69.05{\pm}3.05$ & $75.89{\pm}3.78$ & $62.50{\pm}2.95$ & $60.42{\pm}6.25$ & $46.43{\pm}5.20$ & $48.21{\pm}1.87$ \\
\bottomrule
\end{tabular}%
}
\caption{ALFWorld task success (\%) with variance, four repeats. Columns match
Table~\ref{tab:strong_baseline_appworld}; ``All'' and the fixed-subset
(FS$=k$) trajectory-log baselines are the raw episodes that ACE / AWM / Dynamic
Cheatsheet (DC) compress.}
\label{tab:resp-alfworld-ci}
\end{table}

\subsection{Attribution: Which Part of the Streamed Update Causes Degradation}
\label{app:attribution}

Streamed consolidation could degrade for three separable reasons:
\textbf{(1) initial compression error}---information lost the first time a batch
is abstracted, independent of any memory history;
\textbf{(2) cross-update conditioning}---earlier imperfect abstractions in
context biasing how the next batch is consolidated; and
\textbf{(3) destructive rewriting}---repeatedly editing entries in place erodes
them cumulatively. We separate these with three streamed conditions that see the
same batches in the same order and differ only in what the consolidation step
may see or do (Table~\ref{tab:resp-attrib-design}); results in
Table~\ref{tab:resp-attrib-results}.

\begin{table}[htbp]
\centering
\small
\renewcommand{\arraystretch}{1.15}
\begin{tabular}{@{}>{\raggedright\arraybackslash}p{0.20\textwidth} c c >{\raggedright\arraybackslash}p{0.34\textwidth}@{}}
\toprule
\textbf{Condition} & \textbf{Prior store visible?} & \textbf{Entries modifiable?} & \textbf{What it isolates} \\
\midrule
A --- Append-only, visible   & Yes & No (append only)          & Baseline streamed update without rewriting \\
B --- Append-only, invisible & No  & No (append only)          & Removes mechanism~2; residual is mechanism~1 \\
C --- Add $+$ delete         & Yes & Delete only (no in-place) & Adds pruning to A; separates pruning from rewriting \\
\bottomrule
\end{tabular}
\caption{Attribution design. Neither A nor C edits entries in place, so both are
compared against the rewrite-enabled setting (the default streamed consolidator)
to isolate destructive rewriting.}
\label{tab:resp-attrib-design}
\end{table}

\begin{table}[htbp]
\centering
\small
\begin{tabular}{lllccc}
\toprule
\textbf{Batch} & \textbf{Mode} & \textbf{Condition} & \textbf{Step 100} & \textbf{Step 200} & \textbf{Step 400} \\
\midrule
$1$ & Force & A (visible)      & $48.0$ & $54.0$ & $50.0$ \\
$1$ & Force & B (invisible)    & $72.0$ & $70.0$ & $70.0$ \\
$1$ & Force & C (add+delete)   & $62.0$ & $62.0$ & $64.0$ \\
$1$ & Auto  & A (visible)      & $56.0$ & $60.0$ & $56.0$ \\
$1$ & Auto  & B (invisible)    & $60.0$ & $66.0$ & $66.0$ \\
$1$ & Auto  & C (add+delete)   & $66.0$ & $60.0$ & $60.0$ \\
$8$ & Force & A (visible)      & $60.0$ & $58.0$ & $56.0$ \\
$8$ & Force & B (invisible)    & $52.0$ & $56.0$ & --- \\
$8$ & Force & C (add+delete)   & $58.0$ & $58.0$ & $66.0$ \\
$8$ & Auto  & A (visible)      & $58.0$ & $68.0$ & $58.0$ \\
$8$ & Auto  & B (invisible)    & $56.0$ & $54.0$ & --- \\
$8$ & Auto  & C (add+delete)   & $62.0$ & $64.0$ & $64.0$ \\
\bottomrule
\end{tabular}
\caption{Attribution results (ARC-AGI Stream, GPT-5.4). Missing cells:
condition~B at batch size $8$ overruns the context window (see below).}
\label{tab:resp-attrib-results}
\end{table}

\paragraph{Reading the contrasts.}
\emph{Cross-update conditioning is real, but its sign depends on update
frequency.} At batch size $1$, hiding the prior store helps sharply (Force
$50.0\!\to\!70.0$ at step 400; Auto $56.0\!\to\!66.0$); at batch size $8$ the
effect reverses. This is expected if conditioning carries both a cost---each
imperfect abstraction biases the next update, compounding over $400$ updates at
batch $1$ versus $50$ at batch $8$---and a benefit (seeing where new material
belongs). Conditioning therefore hurts specifically in the high-frequency regime
where our main result is observed.
\emph{Initial compression error persists even with conditioning removed.}
Condition B at batch $1$ is the strongest streamed setting, yet $70.0$ still
trails raw trajectories ($76.6$ ground-truth-code, $70.4$ natural-language;
Table~\ref{tab:resp-main-ci}), neither of which degrades over the stream. With no
prior memory and no rewriting, this residual gap can only be initial compression
error.
\emph{Removing conditioning is diagnostic, not deployable.} Unable to
deduplicate, condition B accumulates $4.83\times$ more entries and $3.81\times$
more text than its visible counterpart at batch $8$, eventually overrunning the
context window (the missing step-400 cells).
\emph{Rewriting damages memory; deletion does not.} Condition C beats A at both
batch sizes ($66.0$ vs.\ $56.0$ at batch $8$, Force, step 400) with a smaller
store, and both beat the rewrite-enabled setting at matched batch size---C
differs from A only by permitting removal, and from rewrite-enabled only by
forbidding in-place edits. Pruning helps; the harm is specific to recursive
in-place rewriting.

All three mechanisms contribute and are separable: initial compression error is
present even in the cleanest condition (B, batch $1$), cross-update conditioning
adds a penalty that scales with the number of updates, and destructive rewriting
adds a third, independent penalty. This predicts the next intervention: fewer,
larger updates should reduce both the conditioning cost (fewer update events) and
the compression error (more reliable task-type segmentation), which
\Cref{app:delayed} confirms.

\subsection{Delayed Compression Mitigates but Does Not Eliminate the Gap}
\label{app:delayed}

Giving the consolidator larger batches (it sees $32$ examples at once rather than
$8$) raises terminal success (Table~\ref{tab:resp-bsz-mode}) and roughly doubles
the fraction of examples the agent chooses to consolidate in autonomous mode
(Table~\ref{tab:resp-delayed}). Both follow from broader within-batch evidence
enabling more accurate task-type segmentation---consistent with the finding
(\S\ref{sec:chunking-failure}) that whole-pool and task-grouped construction
outperform incremental streaming, and that heterogeneous batches merge
incompatible procedures. Delaying consolidation materially mitigates degradation
but does not close the gap to retaining raw experience.

\begin{table}[htbp]
\centering
\small
\begin{tabular}{lcccc}
\toprule
\textbf{Batch size} & \textbf{Up to 100} & \textbf{Up to 200} & \textbf{Up to 400} & \textbf{Ratio} \\
\midrule
$8$  & $27$ & $37$ & $74$  & $74/400 = 18.5\%$ \\
$32$ & $45$ & $76$ & $183$ & $183/400 = 46\%$ \\
\bottomrule
\end{tabular}
\caption{Cumulative consolidation events (autonomous mode) on ARC-AGI Stream.
Larger batches consolidate more often once segmentation is reliable.}
\label{tab:resp-delayed}
\end{table}

\subsection{Relation to Catastrophic Forgetting}
\label{app:cf}

Our failure is related to but distinct from catastrophic forgetting (CF)
\citep{mccloskey1989catastrophic,french1999catastrophic}, the loss of performance
on earlier tasks when a system is sequentially trained on new ones. CF is induced
by domain change and is standardly mitigated by rehearsal/replay or
regularization against drift. Our degradation instead arises from the cumulative
effect of imperfect metacognitive chunking and lossy rewrites, on the same set of
problems. Four observations separate the two:
\begin{itemize}[leftmargin=1.4em,itemsep=2pt,topsep=2pt]
  \item \textbf{Degradation occurs with the task distribution held fixed.} No new
  domain is introduced; repeated consolidation renders ${\sim}46\%$ of
  previously solved ARC-AGI Stream problems unsolvable---the agent loses ground
  on problems it has already mastered, which CF does not predict.
  \item \textbf{Mixing task types within a batch is worse than separating them}
  (\S\ref{sec:phenomenon}, Fig.~\ref{fig:whole-batch-vs-stream}). Under CF,
  interleaving should \emph{reduce} forgetting; we observe the opposite,
  consistent with the consolidator chunking task types incorrectly over a mixed
  batch.
  \item \textbf{The same evidence produces different outcomes by arrival order.}
  Consolidating a stream incrementally is worse than consolidating the identical
  trajectories in one pass, so the damage comes from repeated rewriting rather
  than from what the evidence contains.
  \item \textbf{The corruption is distortion, not loss.} Inspecting the stores
  (\Cref{app:wrong-examples}) we find merged task families, procedures
  over-generalized past their scope, confident-but-incorrect procedures, and
  deleted exceptions/conditionals. CF predicts old knowledge fading; here memory
  instead accumulates assertions that were never true.
\end{itemize}
The degradation is therefore a distinct failure mode that calls for different
mitigations (gated, delayed abstraction with preserved episodes;
\S\ref{sec:lessons-underperform}) than rehearsal or drift regularization.

\subsection{Context Length Is Not the Cause}
\label{app:contextrot}

Because consolidation is a procedural-memory mechanism rather than
context-window compression, we test directly whether extreme context length,
rather than faulty abstraction, drives the collapse. We pad the ARC-AGI Stream
prompt with randomly sampled training problems---first \emph{without} solutions
(length but no usable signal; Table~\ref{tab:resp-e1}), then \emph{with}
solutions (length and information; Table~\ref{tab:resp-e2}). Even at
${\sim}790$K tokens the decline is small and statistically insignificant---far
smaller than the collapse we attribute to consolidation---so context rot cannot
account for our findings.

\begin{table}[htbp]
\centering
\small
\begin{tabular}{lccc ccc cc}
\toprule
& \multicolumn{3}{c}{\textbf{E1-400}} & \multicolumn{3}{c}{\textbf{E1-3.2K}} & & \\
\cmidrule(lr){2-4}\cmidrule(lr){5-7}
\textbf{Model} & \textbf{acc} & \textbf{sd} & \textbf{tokens} & \textbf{acc} & \textbf{sd} & \textbf{tokens} & \textbf{Decline} & $t$ \\
\midrule
GPT-5.5 & $44.2$ & $2.7$ & $98{,}844$ & $42.0$ & $3.7$ & $790{,}535$ & $-2.2$ & $-1.52$ \\
GPT-5.4 & $28.2$ & $4.4$ & $98{,}844$ & $24.6$ & $4.5$ & $790{,}535$ & $-3.6$ & $-1.81$ \\
\bottomrule
\end{tabular}
\caption{Distractor problems shown \emph{without} solutions. E1-400 and E1-3.2K
denote $400$ and $3{,}200$ padding problems; ``tokens'' is the mean context
length over $50$ test problems; ten runs per condition. Decline is in accuracy
points.}
\label{tab:resp-e1}
\end{table}

\begin{table}[htbp]
\centering
\small
\begin{tabular}{lccc ccc ccc}
\toprule
& \multicolumn{3}{c}{\textbf{E2-400}} & \multicolumn{3}{c}{\textbf{E2-1.2K}} & & & \\
\cmidrule(lr){2-4}\cmidrule(lr){5-7}
\textbf{Model} & \textbf{acc} & \textbf{sd} & \textbf{tokens} & \textbf{acc} & \textbf{sd} & \textbf{tokens} & \textbf{Decline} & $t$ & $p$ \\
\midrule
GPT-5.5       & $61.0$ & $4.5$ & $250{,}625$ & $59.2$ & $5.5$ & $733{,}137$ & $-1.8$ & $-0.80$ & $0.437$ \\
GPT-5.4       & $43.4$ & $3.9$ & $250{,}625$ & $41.2$ & $2.1$ & $733{,}137$ & $-2.2$ & $-1.56$ & $0.140$ \\
Claude Opus 5 & $87.6$ & $3.1$ & $250{,}625$ & $86.0$ & $3.3$ & $733{,}137$ & $-1.6$ & $-1.12$ & $0.276$ \\
\bottomrule
\end{tabular}
\caption{Padding problems shown \emph{with} solutions (context is both long and
informative). Declines remain small even at ${\sim}733$K tokens; frontier models
are resilient to extreme context.}
\label{tab:resp-e2}
\end{table}

\subsection{Generality Across Models and Memory Frameworks}
\label{app:generality}

The degradation is not specific to the four consolidation methods or the
backbones in the main text. We extend to seven published memory frameworks with
an additional backbone (Claude Sonnet 5) and to a general-purpose coding agent
(OpenCode) given full autonomy over memory management at maximum reasoning
budget.

\paragraph{Seven frameworks, two backbones.}
Table~\ref{tab:resp-frameworks} reports the peak-minus-step-400 decline for
LangMem \citep{langmem2024}, Mem0 \citep{chhikara2025mem0}, ProPlay, ACE
\citep{zhang2025ace}, AWM \citep{wang2024awm}, Letta \citep{letta2024}, and
Dynamic Cheatsheet \citep{suzgun2025dynamiccheatsheet}. Every framework degrades
in at least one setting; the corresponding ARC-AGI drawdowns (GPT-5.4) are in
Table~\ref{tab:resp-arc-drawdown}. We report mixed cases transparently (e.g.,
LangMem has no WebShop decline with Sonnet 5): the point is that the phenomenon
is not specific to one implementation or design family.

\begin{table}[htbp]
\centering
\footnotesize
\setlength{\tabcolsep}{5pt}
\begin{tabular}{@{}lccccc@{}}
\toprule
& \multicolumn{2}{c}{\textbf{AppWorld}} & \multicolumn{2}{c}{\textbf{WebShop}} & \textbf{SciWorld} \\
\cmidrule(lr){2-3}\cmidrule(lr){4-5}\cmidrule(lr){6-6}
\textbf{Method} & GPT-5.4-mini & Sonnet 5 & GPT-5.4-mini & Sonnet 5 & GPT-5.4-mini (succ.\ / rew.) \\
\midrule
LangMem            & $14$ & $10$ & $21.1$ & $0$   & $20$ / $28.9$ \\
Mem0               & $6$  & $20$ & $9.0$  & $2.6$ & $10$ / $19.6$ \\
ProPlay            & $24$ & $6$  & $9.6$  & $0.6$ & $0$ / $0^{*}$ \\
ACE                & $50$ & $10$ & $17.6$ & $7.6$ & $6$ / $7.9$ \\
AWM                & $14$ & $4$  & $6.4$  & $3.3$ & $6$ / $13.6$ \\
Letta              & $12$ & $12$ & $8.5$  & $0.9$ & $14$ / $20.1$ \\
Dynamic Cheatsheet & $20$ & $14$ & $16.0$ & $5.0$ & $\dagger$ \\
\bottomrule
\end{tabular}
\caption{Peak-minus-step-400 decline across seven frameworks and two backbones.
SciWorld reports success / reward for GPT-5.4-mini.
$^{*}$ProPlay's $0/0$ is a floor effect (it never establishes a useful peak).
$\dagger$Dynamic Cheatsheet overflows the max token budget before finishing on
SciWorld.}
\label{tab:resp-frameworks}
\end{table}

\begin{table}[htbp]
\centering
\small
\begin{minipage}[t]{0.40\textwidth}
\centering
\begin{tabular}{lc}
\toprule
\textbf{Method} & \textbf{Drawdown} \\
\midrule
LangMem  & $8$ \\
Letta    & $8$ \\
ProPlay  & $22$ \\
Mem0 v3  & $6$ \\
\bottomrule
\end{tabular}
\caption{ARC-AGI drawdowns (GPT-5.4).}
\label{tab:resp-arc-drawdown}
\end{minipage}\hfill
\begin{minipage}[t]{0.56\textwidth}
\centering
\begin{tabular}{lc}
\toprule
\textbf{Condition} & \textbf{Accuracy} \\
\midrule
Streamed, epoch 10          & $47.9 \pm 5.0$ \\
Streamed, epoch 10 $+$ GEPA & $66.3 \pm 4.4$ \\
\bottomrule
\end{tabular}
\caption{Prompt optimization (GEPA) on the collapsed epoch-10 state
(\Cref{app:gepa}).}
\label{tab:resp-gepa}
\end{minipage}
\end{table}

\paragraph{Autonomous agentic memory management (OpenCode).}
Using the OpenCode harness with each backbone at its maximum reasoning budget and
broad latitude to inspect, revise, and manage memory, all four models follow the
same trajectory: accuracy rises above zero-shot, peaks, then declines
(Table~\ref{tab:resp-opencode}). By step $400$, GLM 5.2 is back at its zero-shot
level and DeepSeek V4 Pro is below it---the accumulated memory is worse than no
memory at all. Capability shifts where the peak sits (and every model peaks well
before step $400$) but does not prevent the decline, indicating the failure is a
property of repeated rewriting rather than of any particular summarization
pipeline.

\begin{table}[htbp]
\centering
\small
\begin{tabular}{lccccc}
\toprule
\textbf{Model} & \textbf{Zero-shot} & \textbf{Max} & \textbf{ckpt 400} & \textbf{Max $-$ ckpt 400} & \textbf{Peak ckpt} \\
\midrule
GPT-5.5         & $39$ ($78.0\%$) & $40$ ($80.0\%$) & $34$ ($68.0\%$) & $6$ ($12.0$ pp)  & $32$  \\
Claude Sonnet 5 & $20$ ($40.0\%$) & $29$ ($58.0\%$) & $24$ ($48.0\%$) & $5$ ($10.0$ pp)  & $64$  \\
GLM 5.2         & $11$ ($22.0\%$) & $22$ ($44.0\%$) & $11$ ($22.0\%$) & $11$ ($22.0$ pp) & $64$  \\
DeepSeek V4 Pro & $14$ ($28.0\%$) & $20$ ($40.0\%$) & $9$ ($18.0\%$)  & $11$ ($22.0$ pp) & $128$ \\
\bottomrule
\end{tabular}
\caption{OpenCode agentic memory management on ARC-AGI Stream (out of $50$
problems). Zero-shot is no-memory accuracy; Max is the best accuracy reached at
any point; ckpt~400 is accuracy after $400$ consolidation steps; the final
column is the peak checkpoint.}
\label{tab:resp-opencode}
\end{table}

\subsection{Prompt Optimization Does Not Overcome It}
\label{app:gepa}

The consolidation comparison is already prompt-controlled: Static and Stream use
the same solver and consolidation prompt, yet whole-pool consolidation stays at
$18/19$ while incremental consolidation falls to $10/19$ on the regression slice.
To test whether prompt engineering can \emph{recover} a collapsed memory, we
applied GEPA (reflective prompt evolution) to the streamed epoch-10 configuration.
GEPA recovers part of the loss ($47.9\!\to\!66.3$; Table~\ref{tab:resp-gepa}) but
does not overcome the effect. Tellingly, the selected prompt does so by cautioning
the model \emph{against} relying on the memory:

\begin{tcolorbox}[promptbox, title={GEPA-selected prompt (excerpt)}]
You are an expert Python programmer. Treat any provided strategy only as a
tentative hint, not a correct rule. Infer the transformation from the training
examples first; if the strategy conflicts with your reasoning, ignore it. [\dots]
avoid forcing the memorized recipe onto every task, they may be inaccurate or
noisy.
\end{tcolorbox}

\noindent Across diverse frameworks, frontier models, unbounded memory, extreme
context, and prompt optimization, repeated consolidation can erase useful
knowledge---so the failure is not attributable to one weak abstraction method, a
fixed memory budget, or an ineffective prompt.

\end{document}